\documentclass[11pt,a4paper,oneside,final]{memoir}

\usepackage[numbers,square,sort&compress]{natbib}

\usepackage[american]{babel}

\usepackage[utf8]{inputenc} 
\usepackage[T1]{fontenc}

\usepackage{mathptmx}

\usepackage[super]{nth}

\usepackage{microtype}

\usepackage{amsmath}
\usepackage{amssymb}
\usepackage{bbm}
\usepackage{stmaryrd} 
\newcommand{\mA}{\mathbf{A}}
\newcommand{\mD}{\mathbf{D}}
\newcommand{\mQ}{\mathbf{Q}}
\newcommand{\mK}{\mathbf{K}}
\newcommand{\mV}{\mathbf{V}}
\newcommand{\mW}{\mathbf{W}}
\newcommand{\mM}{\mathbf{M}}
\newcommand{\mX}{\mathbf{X}}
\newcommand{\va}{\mathbf{a}}
\newcommand{\vq}{\mathbf{q}}
\newcommand{\vk}{\mathbf{k}}
\newcommand{\vv}{\mathbf{v}}
\newcommand{\vm}{\mathbf{m}}
\newcommand{\vx}{\mathbf{x}}
\newcommand{\vy}{\mathbf{y}}
\newcommand{\vz}{\mathbf{z}}
\newcommand{\vp}{\mathbf{p}}
\newcommand{\vd}{\mathbf{d}}
\newcommand{\vqn}{\vq_\text{new}}
\newcommand{\vkn}{\vk_\text{new}}
\newcommand{\vko}{\vk_\text{old}}

\newcommand{\vvn}{\vv_\text{new}}
\newcommand{\vvo}{\vv_\text{old}}
\newcommand{\sR}{\mathbb{R}}

\DeclareMathOperator*{\Concat}{\mathbin\Vert}

\DeclareMathAlphabet{\mathcal}{OMS}{cmsy}{m}{n}

\newcommand\sbullet[0]{\mathbin{\vcenter{\hbox{\scalebox{1.4}{$\bullet$}}}}}
\newcommand\ssquare[0]{\mathbin{\vcenter{\hbox{\scalebox{0.75}{$\blacksquare$}}}}}
\newcommand\sdiamond[0]{\mathbin{\vcenter{\hbox{\rotatebox{45}{\scalebox{0.70}{$\blacksquare$}}}}}}
\newcommand\striangle[0]{\mathbin{\vcenter{\hbox{\scalebox{0.75}{$\blacktriangle$}}}}}

\usepackage{graphicx}
\usepackage{multirow}

\DeclareSymbolFontAlphabet{\amsmathbb}{AMSb}%
\usepackage[inkscapeformat=eps]{svg}
\usepackage{epsfig}
\usepackage{color}
\usepackage{epstopdf} 
\usepackage[final]{pdfpages}
\usepackage{caption}
\usepackage{subcaption}

\usepackage{soul}
\usepackage{xcolor}
\usepackage{pifont}

\makeatletter
\newcommand\footnoteref[1]{\protected@xdef\@thefnmark{\ref{#1}}\@footnotemark}
\makeatother

\usepackage{thmtools,thm-restate}

\newtheorem{principle}{Principle}

\usepackage{xcolor}
\usepackage{colortbl}
\definecolor{green}{RGB}{7,70,80}
\definecolor{lightgreen}{RGB}{0,146,146}
\definecolor{pink}{RGB}{254,109,182}
\definecolor{lightpink}{RGB}{254,181,218}
\definecolor{purple}{RGB}{72,0,145}

\definecolor{lgreen}{HTML}{5fcf5f} 
\definecolor{lred}{HTML}{db6063} 

\definecolor{ruby}{HTML}{E02020} 
\definecolor{amber}{HTML}{FA6400} 
\definecolor{citrine}{HTML}{F7B500} 
\definecolor{emerald}{HTML}{6DD400} 
\definecolor{amazonite}{HTML}{44D7B6}
\definecolor{apatite}{HTML}{32C5FF}
\definecolor{sapphire}{HTML}{0091FF} 
\definecolor{ioite}{HTML}{6236FF} 
\definecolor{amethyst}{HTML}{B620E0}

\newcommand{\best}[1]{\textbf{#1}}
\newcommand{\nextbest}[1]{\textbf{\textit{#1}}}

\makeatletter
\def\ifdraft{\ifdim\overfullrule>\z@
  \expandafter\@firstoftwo\else\expandafter\@secondoftwo\fi}
\makeatother

\usepackage{xspace}

\makeatletter
\DeclareRobustCommand\onedot{\futurelet\@let@token\@onedot}
\def\@onedot{\ifx\@let@token.\else.\null\fi\xspace}

\def\eg{\emph{e.g}\onedot} \def\Eg{\emph{E.g}\onedot}
\def\ie{\emph{i.e}\onedot}

\makeatother

\usepackage{tikz}

\newcommand{\mycbox}[3]{
\begin{tikzpicture}[remember picture, overlay]
 \draw[#1,fill=#1] (current page.north west) rectangle ([yshift=-0.5cm]current page.north east);
 \node[text=#2] at ([yshift=-0.25cm]current page.north) {#3};
\end{tikzpicture}
}

\usepackage[hyphens]{url}
\usepackage[bookmarks,pagebackref,breaklinks]{hyperref}
\usepackage[capitalise]{cleveref}

\newcommand\contribution[1]{\hspace{0.5em}\hyperref[#1]{P\ref{#1}}}

\newcounter{papercounter}

\newcommand*{\paperref}[1]{Paper~\hyperref[#1]{\ref{#1}}}

\hypersetup{
    linktocpage=true,
    colorlinks=true,
    linkcolor=ioite,
    urlcolor=ioite,
    citecolor=amazonite,
}

\title{Efficient Online Processing with Deep Neural Networks}
\author{Lukas Hedegaard}
\date{June 2, 2023}

\begin{document}

\pagenumbering{roman} 

\hypersetup{
    pdfauthor={\theauthor},
    pdftitle={\thetitle},
}

\thispagestyle{empty}
\setcounter{secnumdepth}{-1}
\begin{center}
\includegraphics[width=4.5cm]{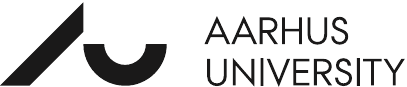}\\[2em]
\vspace*{\fill}
{\Large\scshape PhD Dissertation}\\[1em]
\noindent\rule{\linewidth}{0.5mm}\\[2em]
{\noindent\Huge\bfseries\thetitle}\\[2em]
\noindent\rule{\linewidth}{0.5mm}\\[2em]
{\noindent \Large \textit{by} \ \theauthor}\\[2em]
\vspace*{\fill}
\includegraphics[width=8cm]{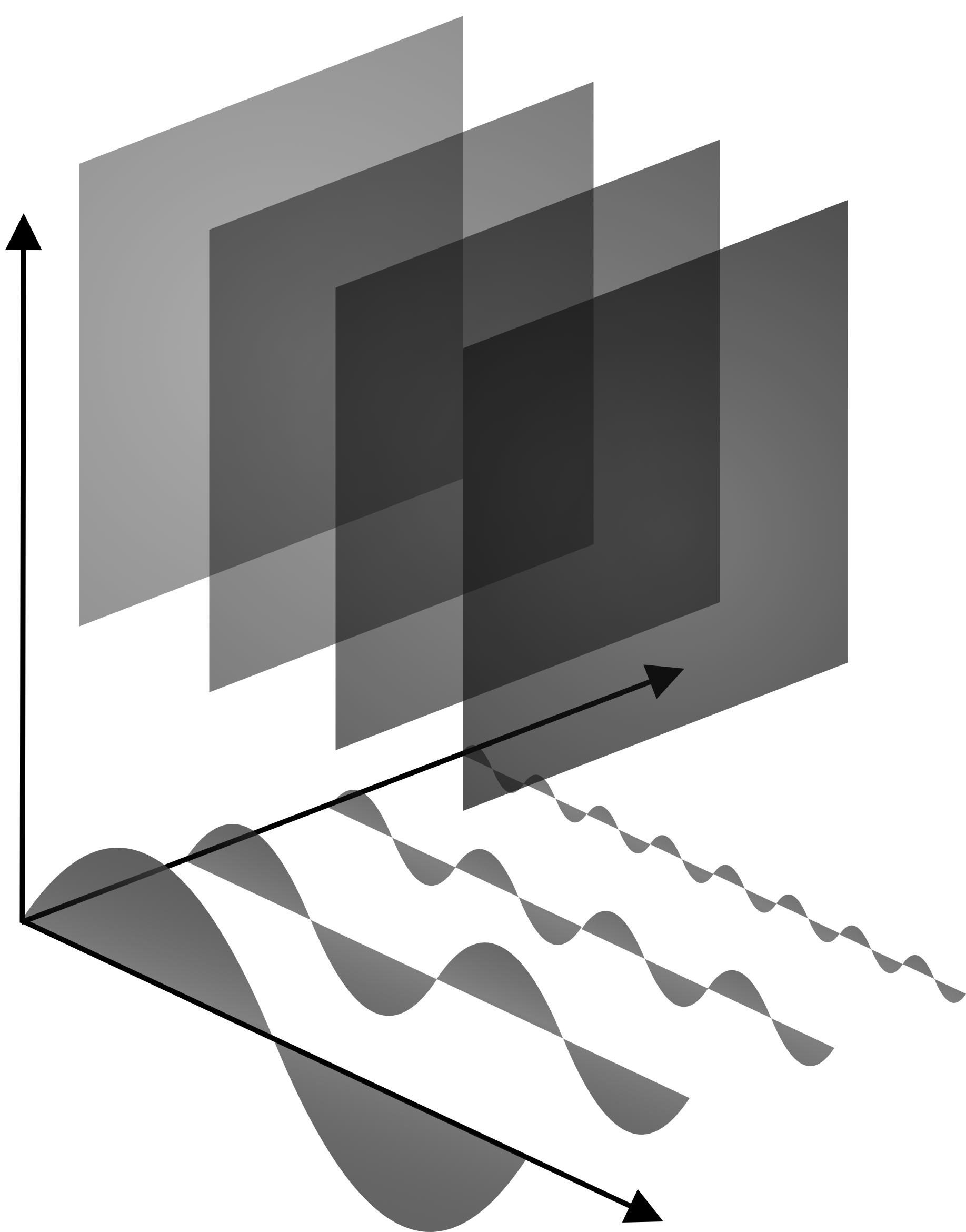}

\end{center}

\cleardoublepage

\thispagestyle{empty} 
\vspace*{\fill}
\begin{center}
{\Large\scshape PhD Dissertation}\\[1em]
\noindent\rule{\linewidth}{0.5mm}\\[2em]
{\Huge\bfseries\thetitle}\\[2em]
\noindent\rule{\linewidth}{0.5mm}\\[1em]
{\itshape Author: \hfill Supervisor:}\\
\theauthor \hfill Alexandros Iosifidis
\vfill
\includegraphics[width=4cm]{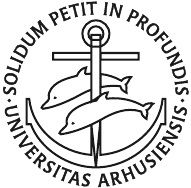}\\[3em]
{ \itshape
    A thesis submitted in fulfillment of the requirements\\
    for the degree of Doctor of Philosophy\\
    \vspace{1em}
    in the faculty of \\
}
\vspace{1em}
Technical Sciences\\
Signal Processing and Machine Learning\\
Department of Electrical and Computer Engineering\\
Aarhus University, Denmark
\vspace{1em}
  
\vfill
\thedate
\end{center}
\vfill

\frontmatter
\cleardoublepage
\chapter*{{\Huge Abstract}}
\addcontentsline{toc}{chapter}{Abstract}
The capabilities and adoption of deep neural networks (DNNs) grow at an exhilarating pace: Vision models accurately classify human actions in videos and identify cancerous tissue in medical scans as precisely than human experts; generative models create beautiful artistic renderings based on simple textual prompts; large language models answer wide-ranging questions, generate code, and write prose, becoming the topic of everyday dinner-table conversations.
Even though their uses are exhilarating, the continually increasing model sizes and computational complexities have a dark side. The economic cost and negative environmental externalities of training and serving models is in evident disharmony with financial viability and climate action goals. 

Instead of pursuing yet another increase in predictive or generative performance, this dissertation is dedicated to the improvement of neural network efficiency. Specifically, a core contribution addresses the efficiency aspects during online inference, where the system must make an updated prediction swiftly after new input information is received, without waiting for subsequent data.
Here, the concept of \textit{Continual Inference Networks} (CINs) is proposed and explored across four publications. CINs extend prior state-of-the-art methods developed for offline processing of spatio-temporal data and reuse their pre-trained weights, improving their online processing efficiency by an order of magnitude. These advances are attained through a bottom-up reorganization of computational components and judicious architectural modifications. The benefit to online inference is demonstrated by reformulating several widely used network architectures into CINs, including 3D Convolutional Nerual Networks, Spatio-temporal Graph Convolutional Networks, and Transformer Encoders.
An orthogonal contribution tackles the concurrent adaptation and computational acceleration of a large source model into multiple lightweight derived models. Drawing on fusible adapter networks and structured pruning, \textit{Structured Pruning Adapters} achieve superior predictive accuracy under aggressive pruning using significantly fewer learned weights compared to fine-tuning with pruning.

This dissertation and the contributions outlined herein constitute necessary steps towards more efficient use of DNNs in scenarios of task specialization as well as online operation. All research-code and supporting Python libraries have been open-sourced at \url{https://github.com/lukashedegaard} in the hope that they will be put to good use and spurt additional research into the efficiency aspects of DNNs.

\cleardoublepage
\chapter*{{\Huge Resum\'e}}
\addcontentsline{toc}{chapter}{Resum\'e}
Dybe neurale netværksmodellers formåen og udbredelse gror med hastigige skridt: Visuelle modeller kan klassificere menneskelige handlinger på videoer og identificerer kraftramt væv på medicinske scanninger lige så præcist som menneskelige eksperter; generative modeller skaber kunstfærdige visuelle renderinger ud fra simple tekstuelle beskrivelser; store sprogmodeller besvarer vidtfavnende spørgsmål, genererer kode, skriver prosa, og er blevet et alment samtaleemne omkring middagsbordet.
Selvom anvendelsesmulighederne er utroligt spændende, har de stødt stigende modelstørrelser og -beregningskompleksiteter en skyggeside. De voksende økonomiske udgifter og negative klimaeksternaliteter ved træning og anvendelse af store neurale modeller ligger i tydelig disharmoni til finansiel rentabilitet og globale klimamål.
 
I stedet for at forfølge endnu en forbedring i prædiktiv eller generativ ydeevne, er denne afhandling dedikeret til effektiviseringen af neurale netværk. 
Et kernebidrag i afhandlingen omhandler effektivitetsaspekter i online anvendelsesscenarier, hvor et system forventes af give opdaterede prædiktioner umiddelbart efter ny input information er modtaget, uden at vente på yderligere data.
Konceptet \textit{Continual Inference Network} (CIN) blev udviklet til denne kontekst på tværs af fire publikationer. CIN’er genbruger netværksarkitekturer og -vægte, der originalt blev trænede til offline \mbox{bearbejdning} af flerdimensionelle tidsvarierende data, og accelererer deres online effektivitet mangefold via reorganiseringer af grundlæggende beregningssekvenser samt minutiøse netværksarkitektoniske justeringer.
Fordelene under online processering blev demonstrerede ved konvertering af adskillige vidt anvendte netværksarkitekturer til CIN’er, herunder 3D CNN’er, ST-GCN’er og Transformer Encodere. 

Et yderlige bidrag takler kombinationen af netværksacceleration og -tilpasning af store basismodeller til flere små specialiserede netværk, der udfører afledte prædiktive opgaver.
Ved at trække på adapternetværk, hvis vægte kan fusioneres med hovednetværket, samt strukturerede pruning metoder, opnår \textit{Structured Pruning Adapters} overlegen prædiktiv performance med væsentligt færre lærte vægte end fine-tuning under aggressiv pruning.

Denne afhandling og dens associerede videnskabelige udgivelser udgør nødvendige skridt i retning af mere effektive dybe neurale netværk i scenarier med opgavespecialisering og online processering. Al forskningskode og supporterende Python biblioteker er frit tilgængelige på \url{https://github.com/lukashedegaard} i håbet om, at de vil finde god anvendelse og inspirere til fremtidig forskning inden for dybe neurale netværks effektivitetsaspekter.

\cleardoublepage
\chapter*{{\Huge Acknowledgments}}
\addcontentsline{toc}{chapter}{Acknowledgments}

First and foremost, I would like to thank my supervisor, Professor Alexandros Iosifidis, for his mentorship and trust throughout my studies, as well as for suggesting that I pursue a PhD degree in the first place. 
I would also like to thank my co-authors and colleagues at Aarhus University and the Machine Learning and Computational Intelligence group for shaping the positive, informal, and inclusive atmosphere that made the work so enjoyable.

We and our work are the product of our circumstances.
I have always had the unwavering support of my father, Birger, and late mother, Wioletta, with whom I spent countless hours doing homework at the kitchen table while my father tended to the farm.
Any words I can muster are insufficient in describing the gratitude I fell for their unconditional love and belief in me.
Thanks also to my brother, Adam, the extended family on my wife's side, as well as my close friends, Morten, Kristian, Omar, and Alexander, for their support, interest, and the talks we have had these past years.
To the love of my life, Ninna, and our son, Arthur, I give thanks for constantly reminding me that life has more to offer than work and the pursuit of abstract goals.
As this treatise builds on prior work, so the people listed above have built me up. Thank you all.


\vspace{2ex}
\begin{flushright}
  \makeatletter\emph{\theauthor,}\makeatother\\
  \emph{Aarhus, \today.}
\end{flushright}

\cleardoublepage
\chapter*{{\Huge Preface}}

This PhD dissertation is written in accordance with the rules and regulations of the Graduate School of Technical Sciences~\cite{au-regulations}. 
Following \S 11.1, this dissertation is a composition of publications and manuscripts relating to the PhD project alongside sections introducing these publications.
Accordingly, the introduction are given in Part~\ref{part:overview} and the publications are re-printed in Part~\ref{part:publications}.
Except for custom color identification on the header, the publications are identical to those published in respective journals, conferences, books, and as pre-prints.

The works comprised within the dissertation received funding from the European Union’s Horizon 2020 research and innovation programme under grant agreement No 871449 (OpenDR).

The dissertation was written without the use of large language models and typeset in \LaTeX.




\newpage
\section{List of publications}

\renewcommand*{\thefootnote}{\fnsymbol{footnote}}

The dissertation is based on the following publications:
\begin{description}
    
    \item[[\ref{pap:co3d}]\hspace{-0.5em}] \textbf{Lukas Hedegaard} and Alexandros Iosifidis. ``Continual 3D Convolutional Neural Networks for Real-time Processing of Videos''. In: Avidan, S., Brostow, G., Cissé, M., Farinella, G.M., Hassner, T. (eds) Computer Vision – ECCV 2022. ECCV 2022. Lecture Notes in Computer Science, vol 13664. Springer, Cham. https://doi.org/10.1007/978-3-031-19772-7\_22 \cite{hedegaard2022co3d}.

    \item[[\ref{pap:costgcn}]\hspace{-0.5em}] \textbf{Lukas Hedegaard},
    Negar Heidari, and Alexandros Iosifidis. ``Continual Spatio-Temporal Graph Convolutional Networks''. Pattern Recognition, Volume 140, 2023, 109528, ISSN 0031-3203 \cite{hedegaard2022costgcn}.
    
    \item[[\ref{pap:cotrans}]\hspace{-0.5em}] \textbf{Lukas Hedegaard}, Arian Bakhtiarnia, and Alexandros Iosifidis. ``Continual Transformers: Redundancy-Free Attention for Online Inference''. In International Conference on Learning Representations (ICLR), 2023
    \cite{hedegaard2022cotrans}.
    
    \item[[\ref{pap:colib}]\hspace{-0.5em}] \textbf{Lukas Hedegaard} and Alexandros Iosifidis. ``Continual Inference: A Library for Efficient Online Inference with Deep Neural Networks in PyTorch''. In Computer Vision – ECCV 2022 Workshops. Lecture Notes in Computer Science, vol 13803. Springer. Isbn: 978-3-031-25066-8 \cite{hedegaard2022colib}.
    
    \item[[\ref{pap:chap14}]\hspace{-0.5em}] \textbf{Lukas Hedegaard}, Negar Heidari, and Alexandros Iosifidis. ``Human Activity Recognition'' in Deep Learning for Robot Perception and Cognition, A. Iosifidis and A.  Tefas, Eds., 1st ed., Academic Press, Jan.  2022, ch. 14, isbn: 9780323857871 \cite{hedegaard2022human}.
    
    \item[[\ref{pap:chap4}]\hspace{-0.5em}] Negar Heidari, \textbf{Lukas Hedegaard}, and Alexandros Iosifidis. ``Graph Convolutional Networks''. In Deep Learning for Robot  Perception and Cognition, A. Iosifidis and A. Tefas, Eds., 1st ed., Academic Press,  Jan. 2022, ch. 4, isbn: 9780323857871 \cite{heidari2022graph}.

    \item[[\ref{pap:spa}]\hspace{-0.5em}] \textbf{Lukas Hedegaard}, Aman Alok, Juby Jose, and Alexandros Iosifidis ``Structured Pruning Adapters''. ArXiv preprint arXiv:2211.10155, 2022 \cite{hedegaard2022structured}. 
    
\end{description}
The following publication was published during the Ph.D. but is not part of the dissertation:
\begin{description}
    \item[\phantom{[}$\bullet$\phantom{]} \hspace{-0.5em}] \textbf{Lukas Hedegaard}, Omar Ali Sheikh-Omar, and Alexandros Iosifidis. "Supervised Domain Adaptation: A Graph Embedding Perspective and a Rectified Experimental Protocol," in IEEE Transactions on Image Processing, vol. 30, pp. 8619-8631, 2021, doi: 10.1109/TIP.2021.3118978 \cite{hedegaard2021sda}.
\end{description}
\newpage
\noindent
Likewise, the following software libraries were developed during the Ph.D. but will not be covered as part of the dissertation:
\begin{description}

    \item[\phantom{[}$\bullet$\phantom{]} \hspace{-0.5em}]  \texttt{Ride}: Training wheels, side rails, and helicopter parent for your Deep Learning projects in PyTorch~\cite{hedegaard2021ride}. \newline\url{https://github.com/lukashedegaard/ride}

    \item[\phantom{[}$\bullet$\phantom{]} \hspace{-0.5em}]  \texttt{Co-Rider}: Tiny configuration library tailored for the Ride ecosystem~\cite{hedegaard2021corider}. \newline\url{https://github.com/lukashedegaard/co-rider}

    \item[\phantom{[}$\bullet$\phantom{]} \hspace{-0.5em}]  \texttt{OpenDR}: A modular, open and non-proprietary toolkit for core robotic functionalities by harnessing deep learning~\cite{passalis2022opendr}. \newline\url{https://github.com/opendr-eu/opendr}

    \item[\phantom{[}$\bullet$\phantom{]} \hspace{-0.5em}]  \texttt{DatasetOps}: Fluent dataset operations, compatible with your favorite libraries~\cite{hedegaard2022datasetops}. \newline\url{https://github.com/lukashedegaard/datasetops}

    \item[\phantom{[}$\bullet$\phantom{]} \hspace{-0.5em}]  \texttt{PyTorch Benchmark}: Easily benchmark PyTorch model FLOPs, latency, throughput, allocated gpu memory and energy consumption~\cite{hedegaard2022pytorchbenchmark}. \newline\url{https://github.com/lukashedegaard/pytorch-benchmark}

    \item[\phantom{[}$\bullet$\phantom{]} \hspace{-0.5em}]  \texttt{Supers}: Call a function in all superclasses using supers(self).foo(42)~\cite{hedegaard2021supers}. \newline\url{https://github.com/lukashedegaard/supers}

\end{description}

\renewcommand*{\thefootnote}{\arabic{footnote}}



\cleardoublepage

\tableofcontents
\cleardoublepage

\mainmatter 

\pagenumbering{arabic}
\setsecnumdepth{subsubsection}

\part{Overview}
\label{part:overview}

\chapter{Introduction} 
\section{The triumph of Deep Learning }\label{sec:triump}

Neural networks have come a long way since McCulloch and Pitts conceived the perceptron in 1943~\cite{mcculloch1943perceptron} and Rosenblatt subsequently implemented the first prototype in 1958~\cite{rosenblatt1957perceptron}.
While Rosenblatt had a clear vision that computers would one day see and understand language, the artificial intelligence (AI) field has undergone multiple upswings interspersed with troughs of disillusionment (\textit{AI winters})
~\cite{russel2003aima}.
Fast-forward to the 2010's and 2020's, deep neural networks (DNNs) are broadly studied and used in universities and companies worldwide, performing many cognitive tasks exclusively reserved for humans but a few years earlier.

In computer vision, a research group at the University of Toronto developed AlexNet~\cite{krizhevsky2012alexnet}, a convolutional neural network (CNN), that achieved a 15.3\% top-5 error rate in the challenging ILSVRC-2012 competition (also known as ImageNet-1k), wherein the algorithm must predict among $1{,}000$ object categories on a test set of $100{,}000$ yet unseen images.
Only four years later, ResNets~\cite{he2016resnet} achieved human-level performance of five percent top-5 error on the benchmark by scaling the network to hundreds of layers.
DNNs are not limited to visual perception but can also be trained to generate images of faces and natural-world objects as showcased by Generative Adversarial Nets in 2014~\cite{goodfellow2014gan}. 
Later, DALL-E~\cite{ramesh2021dalle}, Latent Diffusion Models~\cite{radford2021learning}, and Imagen~\cite{saharia2022photorealistic} brought textual prompt-based photo-realistic scene generation into the mainstream.


In 2017, Transformers~\cite{vaswani2017attention} ushered in a similar architectural revolution in the domain of natural language processing (NLP) as AlexNet did in computer vision. The BERT model~\cite{devlin2019bert}, which utilized the Transformer architecture, achieved large-margin improvements on multiple NLP tasks through the use of masked language modeling. 
Large language models (LLMs) were subsequently scaled to over one billion (GPT-2~\cite{radford2019language}) and one trillion (Switch Transformer~\cite{fedus2022switch}) parameters.
In 2022, ChatGPT~\cite{chatgpt} made LLMs part of the dinner table conversion by providing a chat experience with a seemingly all-knowing conversational partner who could also assist in editing, summarizing, and translating text. 
Despite occasionally fabricating answers and lacking proper disclosure of informational sources~\cite{llmfailures}, some speculate it a possible contender to the PageRank-based search engine \texttt{google.com}~\cite{page98pagerank}.

Deep learning also made headlines after AlphaGo~\cite{silver2016alphago}, a DNN trained with a mix of human supervision and reinforcement learning from games with self-play, beat Go champion Leo Sedoll in 2016. The next year, AlphaZero~\cite{silver2018alphazero} attained superhuman performance in the games of chess, shogi, and Go within only 24 hours of training using tabula rasa reinforcement learning with self-play.
DeepMind, the creators of AlphaGo and AlphaZero, also attracted attention when AlphaFold~\cite{jumper2021highly} achieved a quantum leap in the speed and precision of protein structure prediction in the CASP14 challenge in 2020. Touted as the possibly most important scientific achievement of AI in history~\cite{alphafoldachievement}, AlphaFold has generated a database of over 200 million protein structure predictions that will empower future research in medicine and bioengineering.


\section{The burden of Deep Neural Networks}

\begin{figure}[tb]
    \centering
    \includegraphics[width=0.7\linewidth]{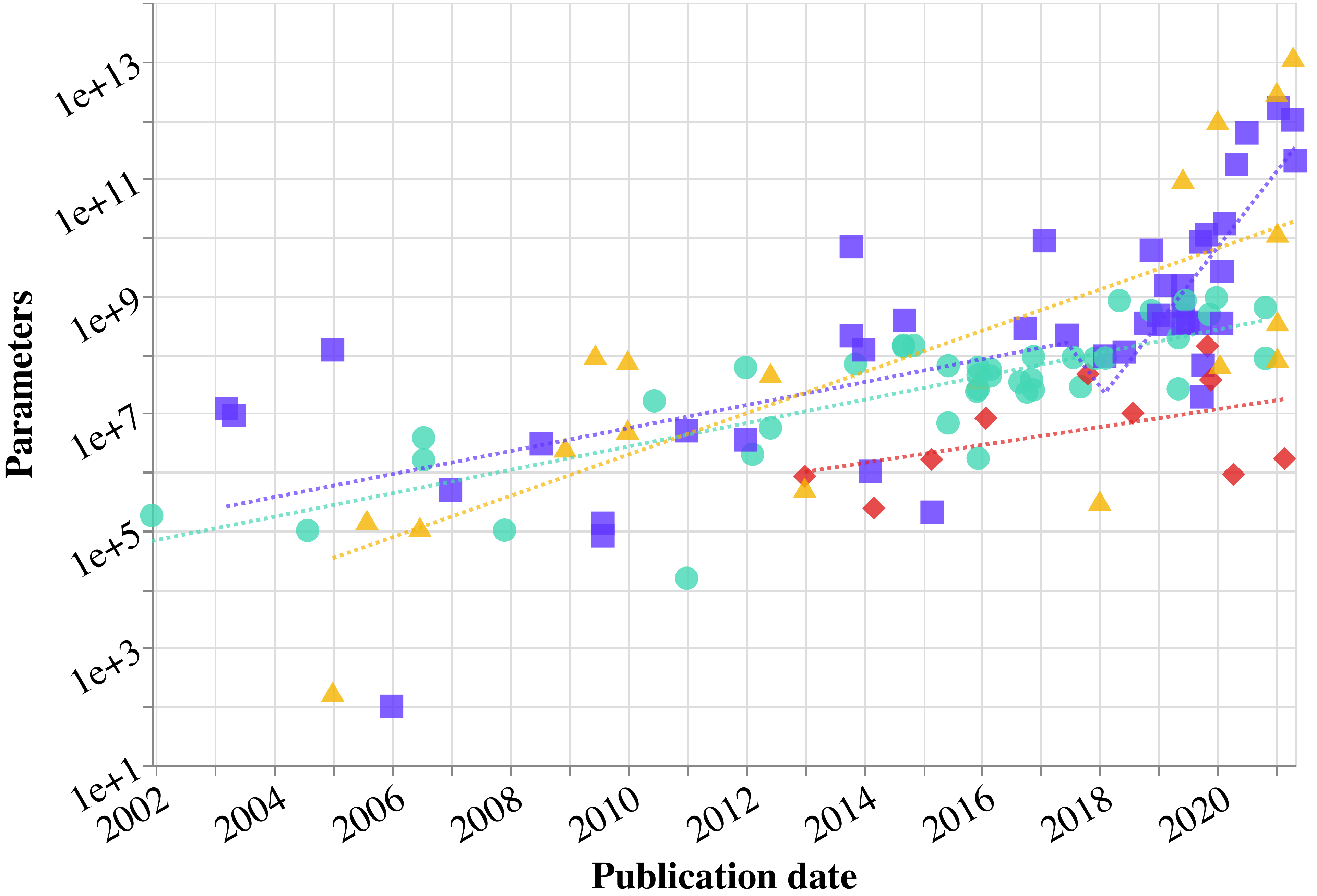}
    \caption{
        Parameter count of prominent deep learning DNNs from 2002 to 2022 for vision (\textcolor{amazonite}{$\sbullet$}), language (\textcolor{ioite}{$\ssquare$}), game (\textcolor{ruby}{$\sdiamond$}), and other (\textcolor{citrine}{$\striangle$}) models. Data source:~\cite{paramcounts}.
    }
    \label{fig:params-through-time}
\end{figure}

The larger the capacity of a Deep Neural Network (DNN), the more abstract and fine-grained distinctions and reasoning processes can be modeled.
\emph{Overparameterization} of DNNs, \ie, the use of significantly more network parameters than training data to fit, has been repeatedly demonstrated to possess favorable learning characteristics such as lower generalization error and robustness to over-training despite perfectly fitting the training data~\cite{soltanolkotabi2017theoretical, advani2020high, nakkiran2020Deep}.
By this logic, networks are continually scaled up in pursuit of higher accuracy and modeling capability. However, this often happens with a disregard to the expense and environmental impact of doing so. Moreover, the continual up-scaling provides diminishing returns with an evident limit in sight~\cite{thompson2020computational}.
%
Consider, for instance, an over-parameterized network optimized to reduce the root mean squared (RMS) prediction error of unseen data given $n$ data points for training. 
The training cost of a network scales multiplicatively with the data in the training set and the parameter count of the model. If we keep the over-parameterization factor $o$ constant\footnote{\Eg, $o=250\times$ as found for the 296M parameter VOLO-D5 model~\cite{yuan2022volo} compared to 1.2M data points in the ILSVRC dataset~\cite{russakovsky2015imagenet}.}, and assume the cost of processing a single data-point $c$ scales $c\propto o \cdot n$, we can provide a rough estimate of the cost of training a model:
\begin{equation}
    \text{cost} \propto \underbrace{o \cdot n}_{c} \cdot n \cdot e \cdot h,
\end{equation}
where $e$ denotes the number of training epochs and $h$ is the number of tested hyperparameter configurations.
Following statistical learning theory, the RMS prediction error cannot not drop more than $1/\sqrt{n}$~\cite{loh2017lower, thompson2020computational}. Equivalently, at least a quadratic increase in available data, $n \rightarrow n^2$, is required to reduce the error by half. 
Accordingly, the training cost increases from $n^2 \cdot o \cdot e \cdot h \rightarrow n^4 \cdot o \cdot e \cdot h$.
Note that this cost estimate is simplified and does not take into account learning strategies such as early stopping nor the dynamics of the double descent phenomenon and the complex interactions between training data volume, model capacity, and training time~\cite{nakkiran2020Deep, belkin2019reconciling, advani2020high, geiger2019jamming}. Still, the outlook to a such a scaling of training cost should be a concern. Judging by the parameter count of the largest DNNs from the year 2002 to 2022, visualized in \cref{fig:params-through-time}, we follow the up-scaling trend with a rapid, even increasing pace, especially in the language-domain. 

Continuing this trend would not only be economically infeasible for the organizations training and utilizing the models, but additionally comes with the externality of increasing load on our collective computational infrastructure and energy supply. By extension, the trend comes at a cost to our environment and climate. For instance, the training of BLOOM was reported to use more than one million NVIDIA A100 GPU hours, and GPT-3 incurred an estimated 502 tons of CO$_2$ equivalent emissions~\cite{scao2022bloom}, the equivalent of driving three million kilometers in a diesel car.
Furthermore, this cost creates a democratic divide between those who cannot afford the expense and the largest commercial actors who are increasingly choosing not to open-source their work.

Are these immense trained models worth the computational cost?
As described in \cref{sec:triump}, DNNs can achieve astounding performance on tasks that were previously reserved for humans and even go beyond human capability.
Once trained, the model can be widely deployed to create value for the users. 
For instance, ChatGPT reached 100 million users within two months of its launch~\cite{chatgptusage}.
However, using approximately $190{,}000$ GPUs hours on NVIDIA A100 GPUs to serve users each day~\cite{chatgptcarbon}, even training cost becomes less important than increased inference efficiency.
Moreover, AlphaGo may have bested Lee Sedol in Go, but while AlphaGo required an estimated one megawatt of power, Lee Sedol's human brain uses only 20 watts - that is $50.000\times$ fewer watts~\cite{alpha-go-lee-sedol}.
Clearly, nature shows that there is room for improvement.

\section{A pursuit of efficiency}



The encompassing theme of \emph{efficient} deep neural networks contains a plethora of potential research directions and improvement opportunities.
These include efficient training methods~\cite{micikevicius2018mixed, loshchilov2017sgdr}, hyperparameter selection approaches~\cite{bergstra2011algorithms}, transfer learning~\cite{hedegaard2021sda, howard2018universal}, and efficient hardware~\cite{chen2019eyeris, yao2020protonic}. 
Within this PhD dissertation, the work towards efficiency is anchored in the algorithmic and architectural aspects of DNNs with the goal of improving the computational and memory efficiency during online processing.

Fully autonomous cars and robots interacting with humans have been frequently imagined in popular culture movies for many years. An ongoing challenge in realizing these visions is the required speed and computational capacity of the computing devices empowering such entities.
To get there, it takes significant research efforts in making visual perception algorithms more computationally efficient and capable of real-time processing under recourse-constrained settings.
%
%
As an effort in this direction, \cref{chap:cin} presents a line of works on online processing of continual input streams. It is based on the observation that many prior state-of-the-art architectures for time-series processing, such as 3D Convolutional Neural Networks~\cite{carreira2017quo, feichtenhofer2020x3d}, Spatio-temporal Graph Neural Networks~\cite{yan2018spatial, shi2019agcn}, and Transformers~\cite{vaswani2017attention}, exhibit significant computational redundancy during online inference.
The associated research question is:
\begin{itemize}
    \item[RQ1]\emph{How can we accelerate state-of-the-art deep neural networks, which were tailored for offline batch-processing of time-series data, to perform online stream processing efficiently?}
\end{itemize}

The concept of \textit{Continual Inference Networks}, proposed and refined in papers \ref{pap:co3d} to \ref{pap:colib}, is introduced as a solution to this. These works feature bottom-up reformulations of the computational sequence of multiple DNN building blocks, alongside guidelines and best practices for implementing architectures that operate efficiently in the stream-processing setting. An overview of these works is given in \cref{chap:cin}.

A second line of work lies in the intersection of computational and storage efficiency.
The disk-space associated with the execution and training of DNNs is often overlooked in favor of ever-more exhilarating network capabilities and predictive performance.
Yet, a possible scenario for the future is a convergence to a smaller set of large \textit{foundation models} forming the base for a large number of derived networks.
If each derived model has the same number of parameters as its base model, it poses both an issue during long-term storage as well as during model transfer and serving: The electricity cost of transferring data is easily overlooked despite posing a non-trivial burden on our infrastructure and environment.
Arguably, derived tasks should not impose the same run-time cost as the foundation model, either. Since the solved task is most likely simpler than that of the base model, should it not be able to solve it using less computational resources? In particular, the studied research question is:

\begin{itemize}
    \item[RQ2]\emph{How can we make parameter-efficient adaptations to pre-trained models while improving their computational efficiency on derived tasks?}
\end{itemize}

\textit{Structured Pruning Adapters (SPAs)} were proposed in paper~\ref{pap:spa} to improve the attainable trade-offs between storage and computational efficiency of derived models. Specifically, SPAs combine parallel fusible Adapters~\cite{rebuffi2018efficient} with structured pruning, which is described in \cref{sec:pruning}, to provide derived models with an order of magnitude fewer learned parameters than fine-tuning with pruning while achieving significantly accelerated inference compared to the base models. The work on Structured Pruning Adapters is described in \cref{chap:spa}.

The computational efficiency aspects of DNNs could be studied alongside any task and task performance metric of interest. 
In this dissertation, the tasks of \textit{human activity recognition} and \textit{image classification} are used to gauge algorithmic efficiency improvements in Chapters \ref{chap:cin} and \ref{chap:spa}, respectively. 
Publication \ref{pap:chap14} covers human activity recognition and reviews prior works on both video-based and skeleton-based recognition. 
Publication \ref{pap:chap4} covers graph theory and graph convolutional networks, which form the basis of skeleton-based recognition methods.


Before delving into the scientific contributions in Chapter~\ref{chap:cin} and \ref{chap:spa}, an introduction to the metrics that quantify efficiency is provided in \cref{chap:eff-defs}, followed by a brief overview of prior approaches for enhancing the inference efficiency of DNNs in \cref{chap:related}.

\chapter{Quantifying efficiency}\label{chap:eff-defs}

\section{Task efficiency}
The majority of prior research on DNNs focuses near unidirectionally on improving predictive performance on tasks of interest, be it the classification accuracy on images~\cite{krizhevsky2012alexnet, liu2015vgg, he2016resnet, dosovitskiy2021vit, yuan2022volo} or of human action recognition on video-clips~\cite{tran2015learning, carreira2017quo, tran2018closer, feichtenhofer2019slowfast} or skeleton-sequences~\cite{yan2018stgcn, shi2019agcn, plizzari2021str}; the mean average precision (mAP) of object detectors~\cite{Girshick2015FastR, redmon2016yolo, liu2016ssd}; the CLEAR and IDF1 scores of trackers~\cite{bernardin2008clear, ristani2016idf1}; the mean intersection over union (mIoU) of semantic segmentation networks~\cite{long2015fcn, he2017maskrcnn, mohan2021efficientps}; the Inception score or Fréchet inception distance (FID) of visual generative models~\cite{heusel2017fid, ho2020denoising}; or the F1, exact match (EM)~\cite{rajpurkar2016squad}, perplexity (PPL), Spearman correlation~\cite{wang2018glue}, BLEU-scores~\cite{papineni2002bleu} of language models~\cite{peters2018elmo, vaswani2017attention, devlin2019bert, raffel2020t5, brown2020gpt3}.

Since overparameterization is known to improve the generalization capabilities of a network~\cite{soltanolkotabi2017theoretical, advani2020high, nakkiran2020Deep}, the simple act of increasing network parameters improves task-related performance metrics in most cases. The trend of growing model size is illustrated in \cref{fig:params-through-time}.
However, the financial realities and associated environmental externalities of training larger and larger models necessitate a multi-faceted view of a model's favorability. 
Moreover, some use-cases require faster model inference to avoid outdated information corrupting the decision processes, for instance real-time monitoring and perception with application in robotics and autonomous vehicles. 

\section{Storage efficiency}
There are multiple metrics, which can aid us in quantifying the storage efficiency of a DNN. 
The \textit{learned parameter count} of a model is reported in most works. 
The numeric format of parameters (\eg, \texttt{float64} or \texttt{int8}) has a significant impact on both predictive performance and on actual \textit{disk space consumption}. 
The model size in bytes, however, is seldom reported if numeric quantization is not of primary focus. 
Instead, most works implicitly assume a \texttt{float32} format, which is the default datatype for popular deep learning frameworks such as PyTorch~\cite{paszke2019pytorch} and TensorFlow~\cite{tensorflow2015whitepaper}.
While the \textit{in-memory size} of a model during execution is proportional to the learned parameter count, the model architecture and shape of input data have an equally important effect that should be considered in unison.
Although model parameters can serve as a proxy for computational efficiency as well, they can be misleading. This is for instance the case in sparsely activated models~\cite{fedus2022switch, nan2022glam}, where only a fraction of model weights are active during inference, or autoregressive models~\cite{devlin2019bert, ho2020denoising}, where parameters are iterative applied many times. 

\section{Computational efficiency}
DNNs produce predictions via complex structures of multiplications and additions between the input data and learned model parameters.
Therefore, the metric \textit{floating-point operations} (FLOPs\footnote{FLOPs in this context should not be confused with floating-point operations per second, which is a common metric of computational performance for hardware such as GPUs.}), \ie, the total number of multiplications and additions during the forward pass of a model, is commonly used as a measure of computational efficiency.
Since most modern hardware uses a fused multiply-add (FMA) instruction set, where an input is multiplied with- and added to a floating-point number in one operation, the \textit{multiply-accumulate operation} (MAC)\footnote{Occastionally referred to as multiply-adds (mult-adds).} metric is commonly used as well. FLOPs and MACs are tightly related with approximately two FLOPs per MAC. 

FLOPs and MACs are hardware-agnostic\footnote{Disregarding the fact, that the MACs definition is based on a hardware-implementation detail.} measures of computational complexity and can be derived by counting the operations of a network during inference.
However, they are not always representative of actual on-hardware performance. For instance, a modified ResNet was trained~\cite{bello2021revisiting} to achieve half the \textit{latency} on both a graphical processing unit (GPU) and tensor processing unit (TPU) despite meausuring double the FLOPs of an EfficientNet~\cite{tan2019efficientnet} with similar accuracy. 
This discrepancy can be explained by the heavy use of depth-wise convolutions in EfficientNets, which have a higher latency per floating-point operation on GPUs and TPUs compared to regular convolutions.

While FLOPs can give a good indication of theoretical computational efficiency, on-hardware tests are necessary for true use-case specific comparisons. 
Commonly, either the \textit{speed} (\ie, inverse latency) or the \textit{throughput} (\ie, number of predictions per second) during inference is noted. Speed measurements use a batch size of one and throughput measurement may use higher batch sizes. This makes a considerable difference on massively parallel hardware such as GPUs or TPUs but has limited effect on CPUs that predominantly conduct sequential processing. 
Similarly, the \textit{energy} used (in joules) to process an input or the \textit{power} consumption (in watts) are interesting metrics for battery-powered devices.

A final note on bench-marking is that GPUs and TPUs do not perform computations in isolation but require another processor (usually a CPU) to transfer data.
The measurement of computational latency should thus include the data transmission to and from GPU/TPU in addition to the latency of computation\footnote{Data transfer times, on-GPU processing time, maximum allocated GPU memory and energy consumption on NVIDIA Jetson devices can be easily measured using the Python library \textit{PyTorch-Benchmark}, \url{https://github.com/lukashedegaard/pytorch-benchmark}.}.

\begin{figure}[tb]
    \centering
    \includegraphics[width=0.5\linewidth]{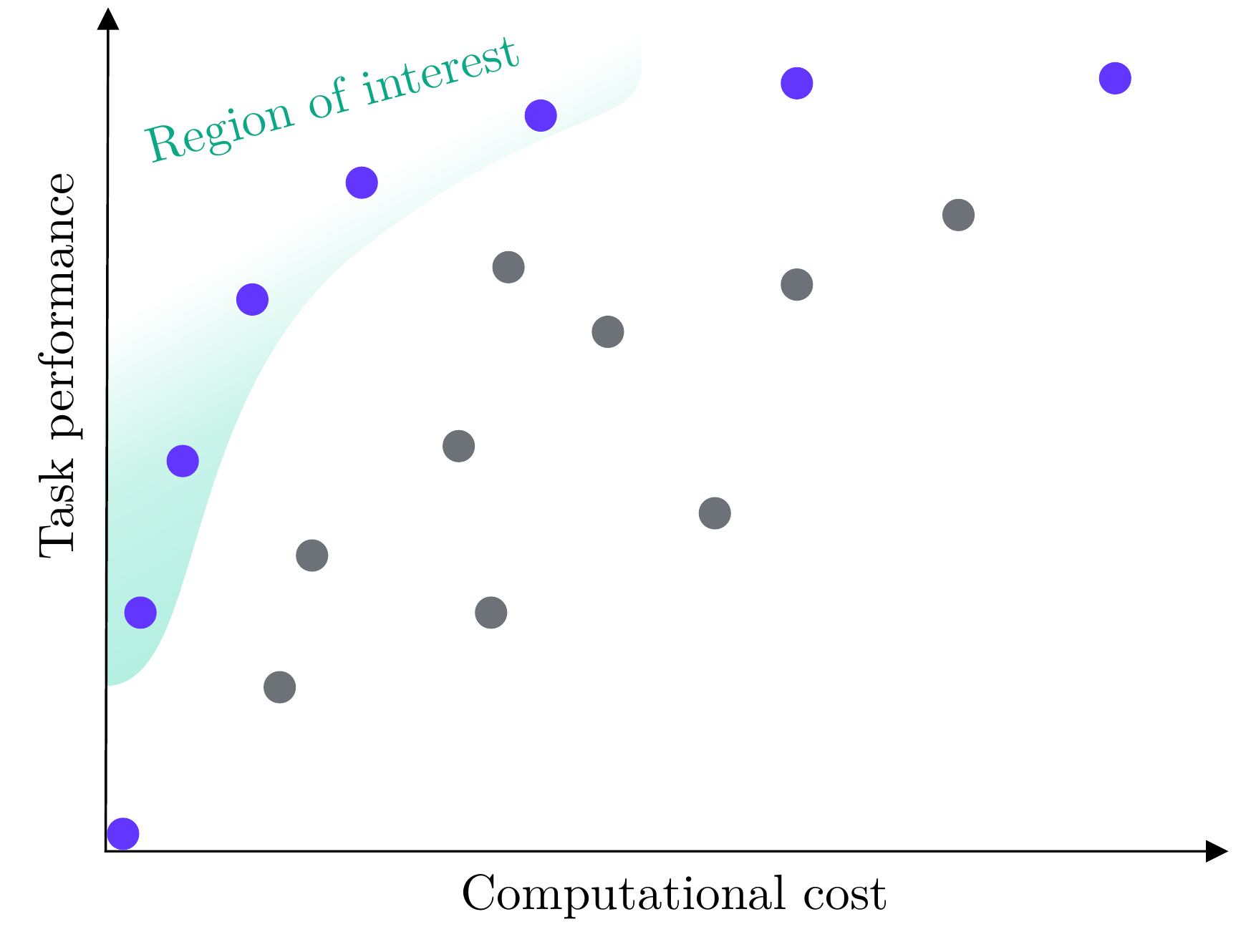}
    \caption{
        Pareto frontier of models \textcolor{ioite}{$\bullet$}. Models that strike a useful balance between task performance and computational cost, \ie, those withing the green region, are of primary interest.
    }
    \label{fig:region-of-interest}
\end{figure}

\section{Efficiency is a trade-off}
The story so far has been restrictive: It makes little sense to judge a model solely on its parameter count, FLOPs, MACs, inference latency, or throughput. A one-parameter algorithm, which always produces a constant-valued output, could be considered as ideal by these measures, even though its usability on any metric of predictive performance is naught.
Hence, any practical measure of efficiency requires trading off predictive and computational capability. \textit{Pareto efficiency}~\cite{pareto} captures this notion: A solution is optimal if there is no other solution where one metric improves without deterioration in another. In that sense, both a single parameter ``constant producing'' algorithm and a 1.6 trillion parameter colossal Switch Transformer~\cite{fedus2022switch}\footnote{Switch-C is among the world's largest language models at the time of writing (Feb, 2023).} are Pareto efficient, but in each their end of the \textit{Pareto frontier}. While either has limited practical utility due to either unusable predictions or prohibitive computational expense, the Pareto frontier covers a multitude of solutions which provide useful trade-offs. It is exactly these possible solutions that are the focus of this dissertation (see \cref{fig:region-of-interest}).

\chapter{Approaches for enhancing inference efficiency}\label{chap:related}

The pursuit of efficient deep neural networks has resulted in a multitude of techniques, which have produced models that exhibit greatly improved inference characteristics compared to non-optimized DNNs. 
This chapter provides an overview of available approaches, specifying how the contributions outlined in this dissertation complements and extends prior literature.


\section{Pruning methods}\label{sec:pruning}

\begin{figure}[bt]
     \centering
     \begin{subfigure}[b]{0.2\textwidth}
         \centering
         \includegraphics[width=\textwidth]{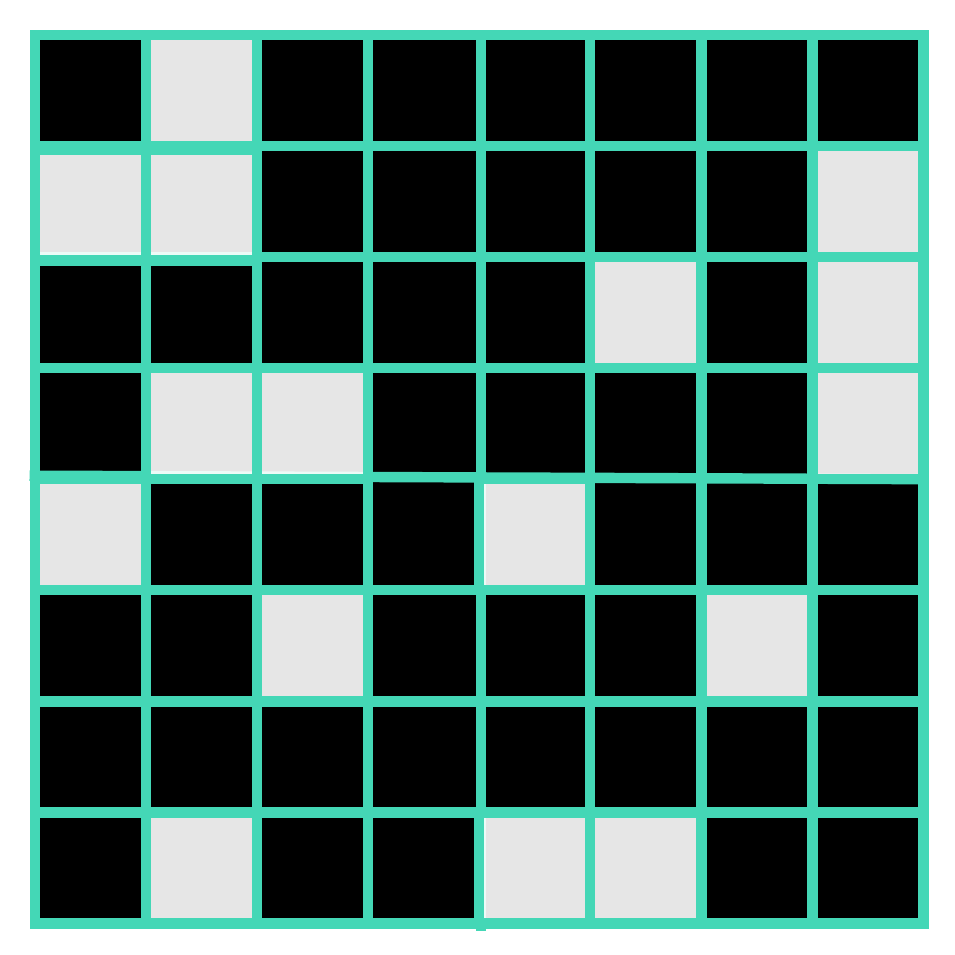}
         \caption{Unstructured}
         \label{fig:unstructured-pruning}
     \end{subfigure}
     \hfill
     \begin{subfigure}[b]{0.2\textwidth}
         \centering
         \includegraphics[width=\textwidth]{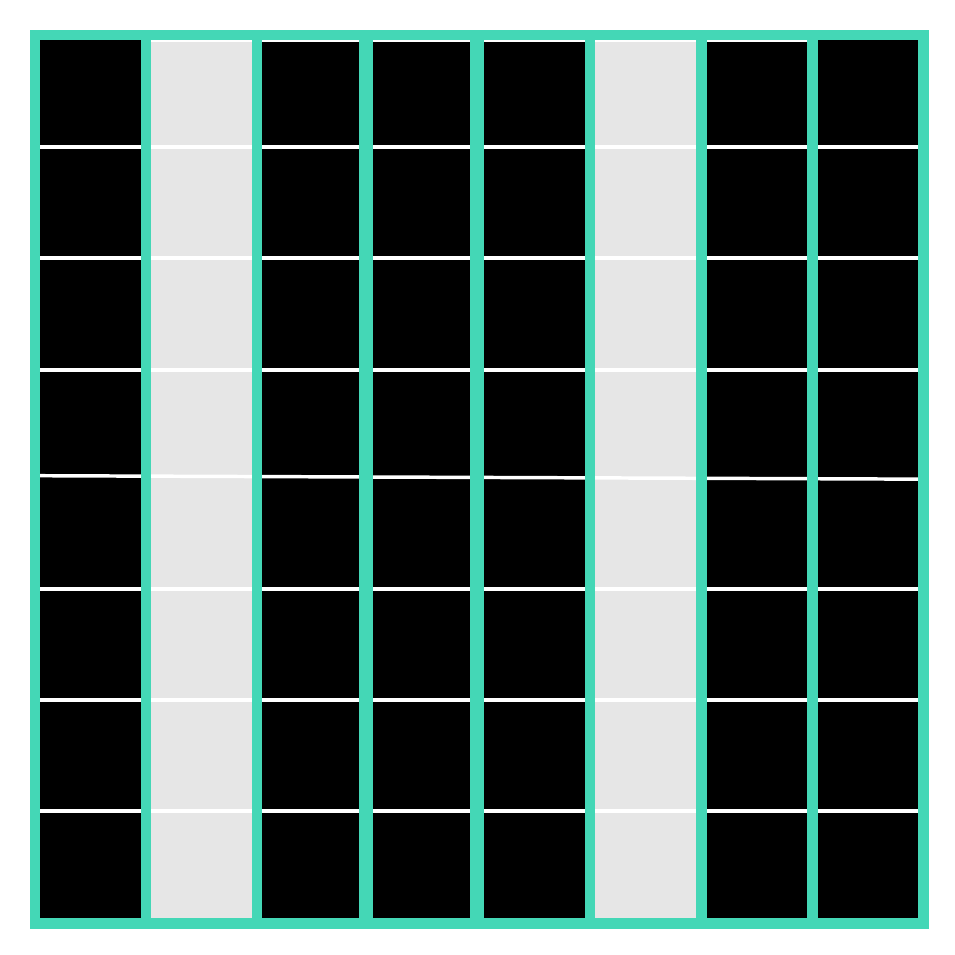}
         \caption{Channel}
         \label{fig:channel-pruning}
     \end{subfigure}
     \hfill
     \begin{subfigure}[b]{0.2\textwidth}
         \centering
         \includegraphics[width=\textwidth]{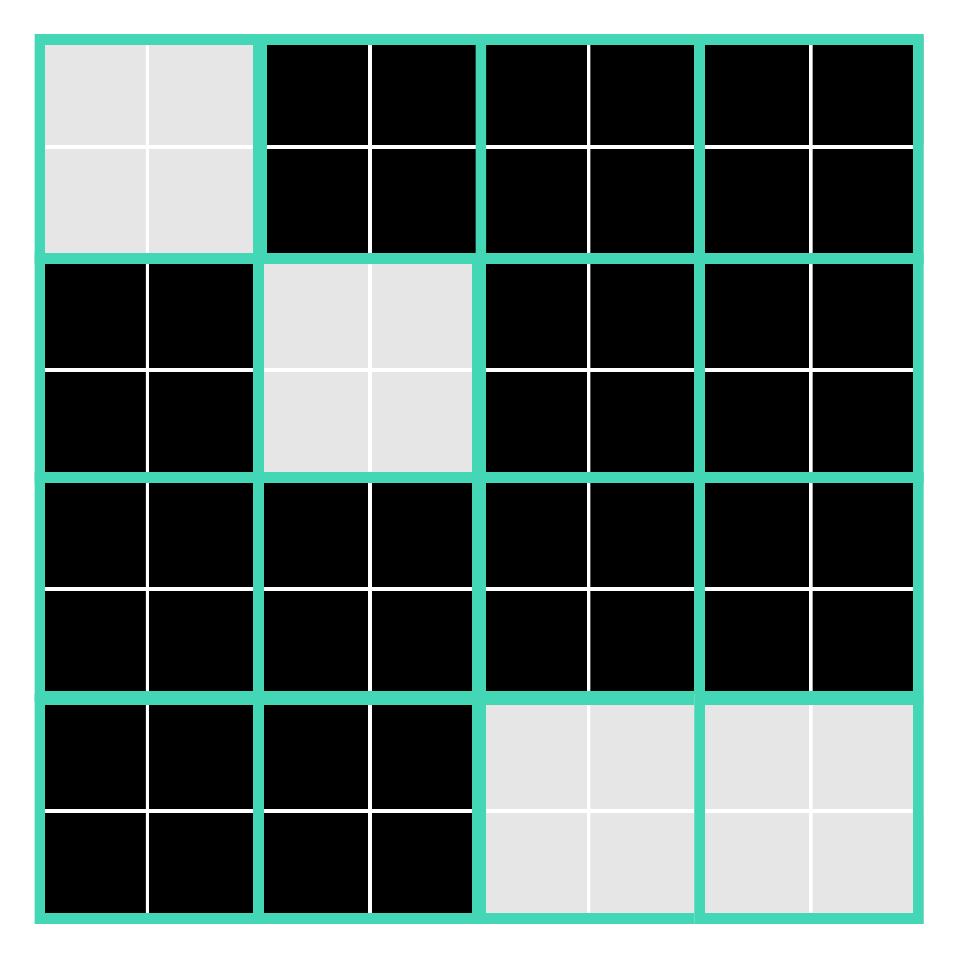}
         \caption{Block}
         \label{fig:block-pruning}
     \end{subfigure}
     \hfill
     \begin{subfigure}[b]{0.2\textwidth}
         \centering
         \includegraphics[width=\textwidth]{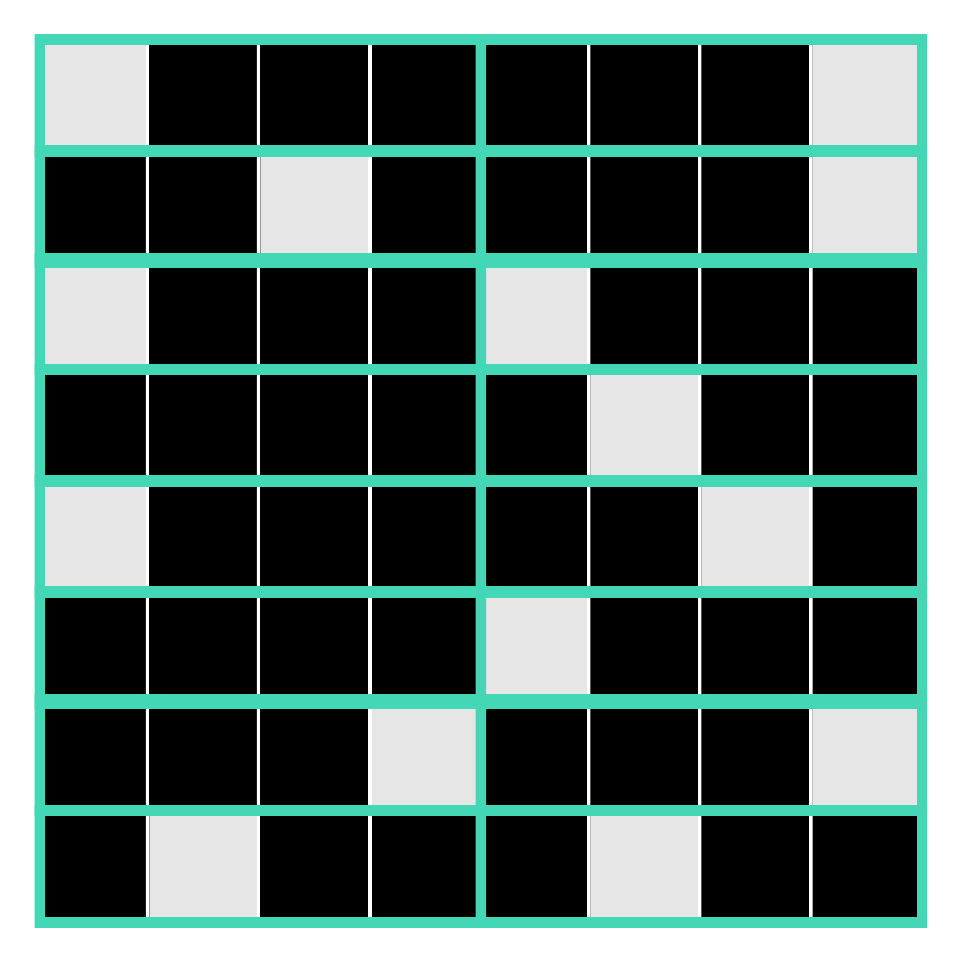}
         \caption{N:M}
         \label{fig:nm-pruning}
     \end{subfigure}
        \caption{Different pruning approaches at 25\% pruning. White blocks are pruned and green contours indicate pruning structure.}
        \label{fig:pruning-approaches}
\end{figure}

An intuitive and effective approach to creating models with good efficiency trade-offs is to take an existing DNN that performs well on the task-metric of interest and \textit{prune} (\ie, remove) its least important weights. 
In the simplest version, every parameter in the network is considered in its own right and ranked either \textit{globally} (for the whole network) or \textit{locally} (within a network layer) to determine which weights to discard. This is referred to as \textit{unstructured} pruning (\cref{fig:unstructured-pruning}). 
Other methods prune multiple structurally connected weights in unison. This is called \textit{structured} pruning.
Owing to the flexibility in weight selection, unstructured pruning generally entails less deterioration of predictive performance when the model is pruned than structured pruning. However, unstructured pruning results in sparse weight matrices at a computational disadvantage on GPUs and TPUs, but it has the potential to greatly reduce the computational run-time on CPUs~\cite{neuralmagic}.

Structured pruning methods are generally capable of speeding up models across CPUs as well as GPUs and TPUs, though some restrictions may apply.
Pruning of entire network \textit{channels} is a simple structured approach, which always comes with computational benefits. For fully connected layers, pruning of a unit corresponds to removing an entire column/row of weights (\cref{fig:channel-pruning}) before/after the feature-map whose channel is pruned. For convolutional layers, a channel plane in the convolutional kernel is removed during channel-pruning.

\textit{Block} pruning~\cite{lagunas2021block} structures a matrix into a grid of $p\times q$ weights where each weight block is bundled as one structural unit (\cref{fig:block-pruning}). This corresponds to the manner in which GPUs divide and process large matrix multiplications. GPU acceleration can subsequently be attained using a block-sparse GPU kernel~\cite{gray2017gpu}. 

In \textit{N:M} structured pruning~\cite{pool2021nmpruning, zhou2021learning}, weights are removed in such a way that N out of M consecutive elements in a matrix row become zero (\cref{fig:nm-pruning}). By permuting the weight matrix channels and surrounding layers appropriately, findings suggest that a 2:4 sparsity pattern could achieve a 2$\times$ speedups on NVIDIA Ampere GPUs without sacrificing accuracy~\cite{mishra2021accelerating}.

The literature on pruning has many other interesting directions. 
These include the design on ranking criteria (\eg, based on weight magnitude~\cite{li2017pruning}, \nth{1} and \nth{2} order gradients~\cite{sun17meprop, molchanov2017pruning}, or explainability~\cite{yeom2021pruning}), and when to prune (\eg, iteratively~\cite{molchanov2017pruning} and with the option to recover from pruning~\cite{zhu2018toprune}). Finally, \textit{knowledge distillation}, which we will describe in the next section, is a popular technique for retaining model performance during pruning~\cite{Sanh2019DistilBERTAD, tang2019distilling, sanh2020movement}.

Paper \ref{pap:spa}, which is summarized in \cref{chap:spa}, proposes the use of ``fusible'' adapters \cite{rebuffi2017learning} alongside channel-based structured pruning to create light-weight add-on networks that specialize a base-network with extremely few new learned parameters.

\section{Knowledge distillation}
Originally proposed for compressing the knowledge of an ensemble of models into a single model~\cite{hinton15distilling}, \textit{knowledge distillation} (KD) is a widely applied model compression technique~\cite{bucilua2006model} for transferring the knowledge of a cumbersome teacher model to an efficient student model.
Specifically, a student model is supervised by the soft target distributions produced by a teacher model using a high temperature coefficient for the softmax operation. If ground truth labels (hard targets) are available, a weighted sum of objectives for the soft and hard targets is employed.

There are multiple variants of KD, including \textit{deep mutual learning}~\cite{zhang2018mutual}, where an ensemble of students teach each other concurrently during the training process; \textit{assistant teaching}~\cite{mirzadeh2019improvedkd}, where models of intermediary size (teacher assistants) are employed to bridge the supervision from teacher models with too large capacity to effectively teach low-capacity student networks; \textit{lifelong learning}~\cite{zhai2019lifelong} uses KD to avoid catastrophic forgetting while learning new tasks by employing the prior model as teacher alongside an auxiliary set of data from previous tasks; \textit{teacher-free KD}~\cite{yuan2020revisiting} uses a pre-trained student model to regularize itself.

KD has also been successfully used in combination with other efficiency approaches such as pruning~\cite{jiao2020tinybert, Sanh2019DistilBERTAD, sanh2020movement}, efficient architectures~\cite{touvron2021training, xie2020noisystudent}, and quantization~\cite{polino2018model, mishra2018apprentice}. Many more methods have employed and expanded on KD than can be covered here. A comprehensive survey by Guo et al.~\cite{gou2021kd} explores the field in more detail.

\section{Quantization}
Quantization methods are concerned with the mapping of continuous numbers to discrete numerical representations.
Here, the opposing goals of high representation accuracy (for a use-case of interest) and of low representation size (few storage bits) set the stage for a plethora of methods~\cite{gholami2022quantization}. 
Although quantization is a long-studied problem~\cite{riemann1867ueber, shannon1948mathematical}, the over-parameterization and computational expense of DNNs yields novel opportunities.
Particularly, their over-parameterization makes them robust to intense quantization, which can reduce computational expenses significantly.
Also, their high degrees of freedom makes it possible to attain good generalization performance on the forward error metric of interest (\eg, accuracy) despite having high error between original and quantized weights.

Quantization formats are either \textit{uniform} or \textit{non-uniform}, \ie, with or without equal spacing between values.
A uniform quantization $Q$ for a real valued input $r$ is given by
\begin{equation}
    Q(r) = \text{Int}(r / S) - Z,
\end{equation}
where $S\in \mathbb{R}$ is a scaling factor and $Z \in \mathbb{Z}$ denotes the zero point.
If $Z=0$, the quantization is \textit{symmetric}, \ie, zero-centered, which can reduce computational costs~\cite{wu2020integerqf}. $Z\neq0$ yields an \textit{asymmetric} quantization, which better represents imbalanced weights and activations such as ReLU. 
In either case, an important consideration for the quantization is the \textit{calibration} of its \textit{clipping range} $\left[\alpha, \beta\right]$, \ie, the minimum and maximum represented values. For a bit width $b$, the scaling factor is
\begin{equation}
    S = \frac{\beta - \alpha}{2^b - 1}.
\end{equation}
Here, $-\alpha = \beta$ is used for symmetric quantization.
Commonly, the minimum and maximum observed values are used to define $\alpha$ and $\beta$. Percentiles~\cite{mckinstry2019discovering} or minization or the KL divergence between real and quantized values~\cite{migacz2017nvidia} are also usable.
While it is straight-forward to determine the clipping range for model weights, which are readily available, the calibration of activations can either be \textit{static} or \textit{dynamic}. 
In dynamic quantization, signal statistics are determined on the fly. Although this can provide higher predictive performance, it entails a considerable computational overhead. Static quantization uses a set of input data to calculate typical activation ranges \textit{a priori}~\cite{jacob2018quantization}.

Quantization of DNNs is rich field with many research directions. These include quantization granularity (considering layers~\cite{krishnamoorthi2018quantizing}, groups~\cite{shen2020qbert}, or channels~\cite{jacob2018quantization}), quantization aware training (QAT)~\cite{courbariaux2015binaryconnect, jung2018learning, esser2020learned}, post-training quantization (PTQ)~\cite{banner2019post, fang2020post, nagel2019datafree}, mixed-precision quantization~\cite{dong2019hawq, habi2020hmq}, and extreme quantization~\cite{courbariaux2015binaryconnect, hubara2016bnn}.

\section{Low-rank factorization of weights}
Another way to compress model weights is through low-rank tensor decomposition.
Multiple decomposition schemes have been explored, including Singular Value Decomposition~\cite{denton2014exploiting}, CP decomposition~\cite{kolda2009tensor, tran2018improving}, and a fast SVD-free ``greedy bilateral'' scheme called GreBsmo~\cite{zhou2013greedy, yu2017compressing}. 
In addition to the low-rank factorization, which captures global weight trends, some methods~\cite{yu2017compressing, guo2019compressing} add sparse weights to model local variations better. This may, however, prohibit reduction in computational latency.
Finally, some methods treat the decomposed weights as efficient architectural structures and train them from random initialization~\cite{tran2018closer, tran2018improving}.

\section{Efficient architectures}
Fully connected layers, convolutions and scaled dot-product attention~\cite{vaswani2017attention} operations can be computationally expensive if feature and weight dimensions are not carefully managed. 

Depth-wise separable convolutions, \ie, a depth-wise convolution followed by a point-wise convolution, are widely used to reduce the computational complexity of both 2D convolutions~\cite{he2016resnet, howard2017mobilenet, zhang2018shufflenet, tan2019efficientnet} and 3D convolutions~\cite{kopuklu2019resource, feichtenhofer2020x3d} on a micro-level. In practice, however, their speed can be memory constrained on some computational devices~\cite{zhang2018shufflenet}. Multi-linear convolutional filters are explored in~\cite{tran2018improving}.

Shift modules, which shift features along a feature-dimension, have been utilized by multiple works~\cite{bichen2028shift, yan2018shiftnet, jeon2018constructing, lin2019tsm, yang2019synetgy, cheng2020skeleton} as an alternative to spatial or temporal convolutions, with greatly reduced number of floating-point operations. While feature shifts do not add FLOPs, it should be noted that they do impose a computational overhead in practice. 

The scaled dot-product attention (SDA) in Transformers~\cite{vaswani2017attention} has itself received considerable attention in attempts to alleviate its computational complexity scaling quadratically with the sequence length. Particularly, some works~\cite{parmar2018image, dosovitskiy2021vit} group entities into fewer blocks that attend to one another. Other works approximate the self-attention matrix~\cite{wang2020linformer, xiong2021nystromformer, choromanski2021rethinking}. Paper \ref{pap:cotrans}~\cite{hedegaard2022cotrans} exploits temporal redundancies observed in the SDA during stream processing to reduce complexity to scale linearly with the sequence length.

On an architectural level, one line of works is concerned with balancing the scaling of input resolutions, network depth, and network channels~\cite{tan2019efficientnet, feichtenhofer2020x3d, cai2020once}.
Early exit architectures dynamically reduce computational cost based the ``hardness'' of predictions or to meet computational deadlines~\cite{scardapane2020why, bakhtiarnia2022single}.
Finally, neural architecture search (NAS) methods automate the network design based on a predefined search space and performance metric (\eg, an accuracy-latency trade-off on mobile devices~\cite{tan2019mnasnet}). While the architecture search of early approaches was very computationally expensive~\cite{zoph2017neural, zoph2018learning}, newer works focus on reducing computational expense, for instance by means of weight sharing~\cite{pham2018efficient, cai2020once} of progressive search~\cite{tran2017progressive}.

\section{Caching and temporal redistribution of features}
Recent developments in time-series processing and specifically in human action recognition have favored convolutional- and Transformer-based models over Recurrent Neural Networks (RNNs)~\cite{carreira2017quo, feichtenhofer2019slowfast, arnab2021vivit, wang2021oadtr, xu2021long}. 
While this pushed the state-of-the-art in task performance as well as offline processing efficiency (\ie, inference over a spatio-temporal clip), Convolutional Neural Networks (CNNs) and Transformers~\cite{vaswani2017attention} could not incrementally process individual frames at a time to produce predictions in an online manner, except through the use of input caching and sliding window processing, where a shifted spatio-temporal clip is assembled and processed in each step. 

To alleviate this, \textit{Continual Inference Networks}, which are described in the papers \ref{pap:co3d} to \ref{pap:costgcn} and summarized in \cref{chap:cin}, propose a bottom-up restructuring of the CNN, Spatio-temporal Graph Convolutional Network (ST-GCN)~\cite{yan2018spatial} and Transformer building blocks, to augment prior models with the ability to incrementally process time-series input efficiently one time-step at a time. 
The proposed restructured networks can reuse the network weights of pre-trained CNNs, ST-GCNs, and Transformers to provide reductions in computational complexity proportional to the temporal receptive field of the source networks during online processing of a continually arriving stream of data.

Other approaches are not weight compatible with prior model weights and instead utilize specialty architectures that either segregate spatial and temporal convolutions~\cite{singh2019recurrent, kopuklu2022dissected} or cache spatial features through time~\cite{habibian2022delta, wu2022memvit}.

\section{Efficient run-times}
In addition to the efficiency approaches outlined in the prior sections, there exist a number of \textit{compilers}. These optimize the network structure for training and deployment, \eg, by fusing batch normalization weights with convolutional kernels. 
While most frameworks widely support PyTorch~\cite{paszke2019pytorch}, TensorFlow~\cite{tensorflow2015whitepaper} and ONNX~\cite{bai2019onnx} model formats, the support for target hardware varies.
Apache TVM~\cite{chen2018tvm}, ONNX Runtime~\cite{onnxruntime}, OpenXLA~\cite{openxla}, and BladeDISC~\cite{bladedisc} support a wide variety models formats and platforms;
OpenVINO~\cite{openvino} optimizes for Intel hardware; TensorRT~\cite{tensorrt} optimizes deployments on NVIDIA GPUs; DeepSparse~\cite{deepsparse} optimizes the runtime of sparse networks on CPUs; and TFLite~\cite{tflite} optimizes TensorFlow for mobile and edge devices.
Finally, Speedster~\cite{speedster} is an effort to provide a unified interface that automatically selects the appropriate compiler.

\chapter{Redistributing computations through time with Continual Inference Networks}\label{chap:cin}





\section{Introduction}

The chase to push predictive performance in tasks such as human activity recognition has led the field through a series of phases, where different architectural designs have been dominant.
For instance, Recurrent Neural Networks (RNNs), which can be used to temporally aggregate the spatial (image) features extracted by 2D Convolutional Neural Networks (CNNs), were considered state-of-the-art between 2014 and 2017~\cite{donahue2017long, ng2015beyond}. 

While CNNs with spatio-temporal 3D kernels were not uncommon during this time-period, they initially struggled to beat prior predictive performance, presumably due to the limited dataset-sizes, which were insufficient to effectively optimize the 3D kernels. With the introduction of the Kinetics-400 dataset~\cite{kay2017kinetics} as well as the pre-training technique of \textit{inflating} 2D kernels of models trained on ImageNet1k~\cite{russakovsky2015imagenet} to 3D, a new level of predictive performance was achieved with I3D~\cite{carreira2017quo}. 
In the following years, 3D CNN architectures were further refined with methods such as R(2+1)D~\cite{tran2018closer}, which partitioned the 3D convolution into a separate spatial 2D convolution and temporal 1D convolution; SlowFast~\cite{feichtenhofer2019slowfast}, which processed the input data in multiple branches with different capacities; X3D~\cite{feichtenhofer2020x3d}, which employed an EfficientNet-inspired~\cite{tan2019efficientnet} design based on depth-wise separable convolutions and proposed a progressive network scaling approach; and MoViNets~\cite{kondratyuk2021movinet}, which used neural architecture search and temporal ensembles to boost performance further.

Inspired by the non-local means method ~\cite{buades2005nonlocal} and Transformer self-attention~\cite{vaswani2017attention}, Non-local Neural Networks~\cite{wang2018nonlocal} were an early work to augment 3D CNNs such as I3D with spatio-temporal self-attention.
A few years later, the success of ViT~\cite{dosovitskiy2021vit} in the image domain, spurted multiple methods that adopted transformer architectures in the video-domain, \eg, ViT-B VTN~\cite{neimark2021vtn}, ViViT~\cite{arnab2021vivit}, TimeSFormer~\cite{bertasius2021timesformer}, MViT-v1~\cite{fan2021mvit} and -v2~\cite{li2022mvitv2}. 
At the time of writing, the dominant design is still based on Transformers.

\begin{figure}[bt]
    \centering
    \includegraphics[width=0.7\linewidth]{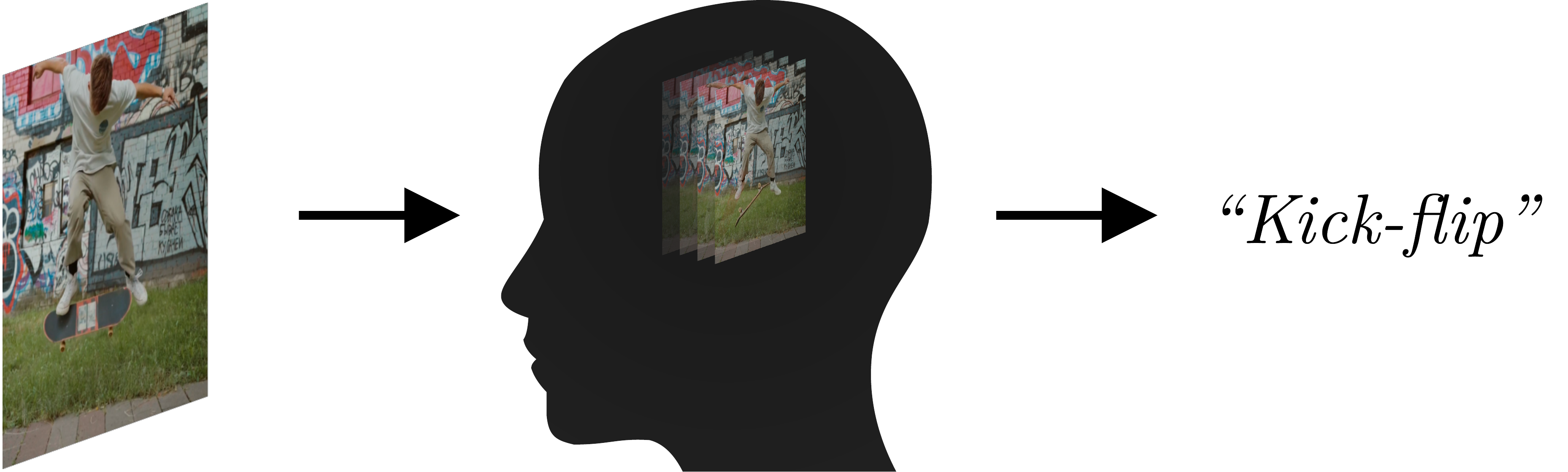}
    \caption{
        Perception in the natural world happens continually, one impression at a time, with temporal context modeled internally.
    }
    \label{fig:natural-world}
\end{figure}

The aforementioned methods were all driven by a competition to attain the best performance on \textit{offline} video-clip prediction problems, \ie, the assignment of class-labels to a fixed-size image sequence.
However, the natural world is not composed of fixed-size temporal blocks. 
We experience a continual stream of visual impressions without inherent beginning and end (see \cref{fig:natural-world}). 
In our interaction with the world, on-the-fly classification of events is crucial to our survival.
Similarly, online visual recognition is central in many important use-cases such as autonomous vehicles, collaborative robots, and live monitoring systems. 

\begin{figure}[bt]
    \centering
    \includegraphics[width=0.8\linewidth]{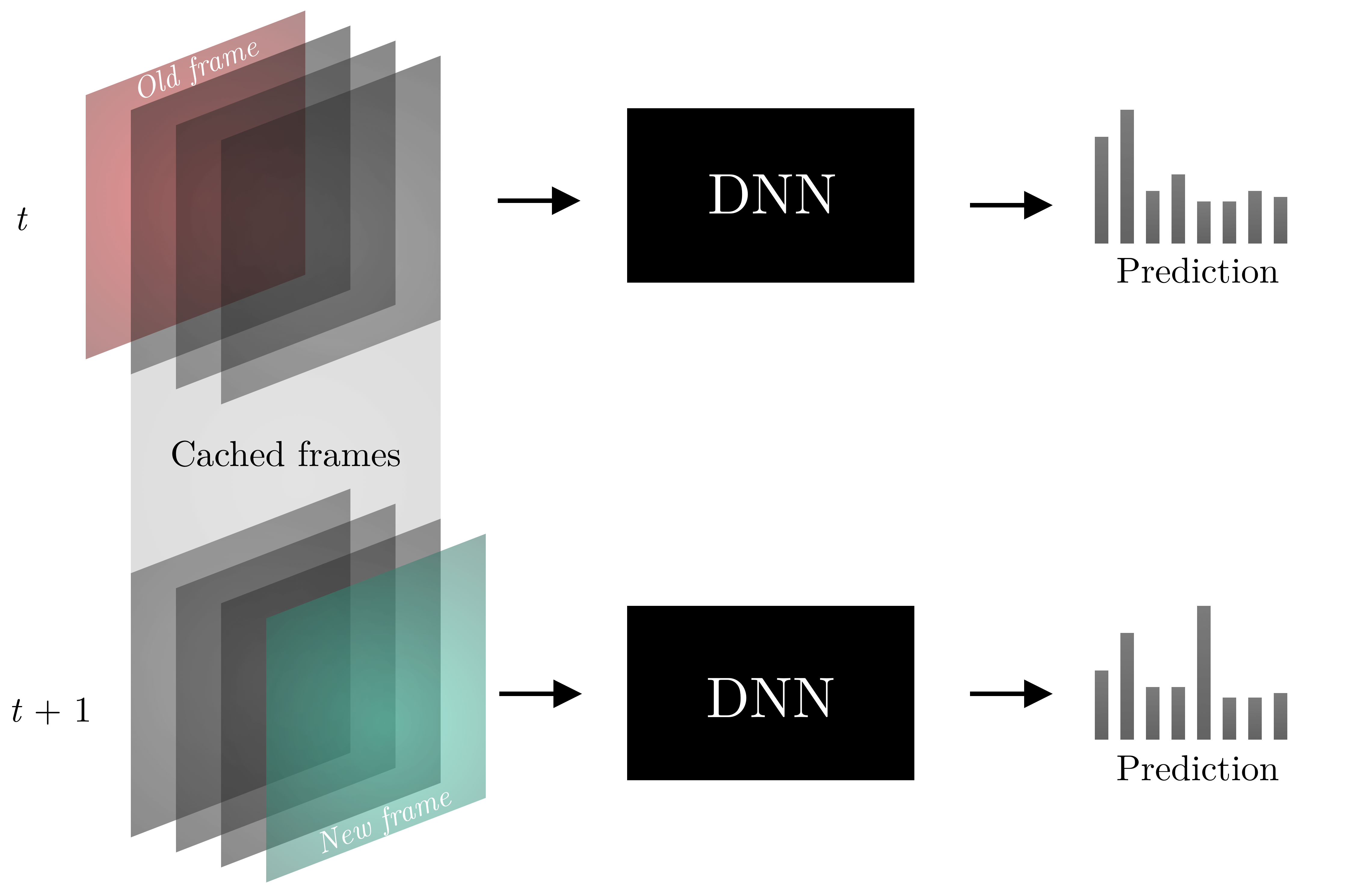}
    \caption{
        Sliding-window processing with DNNs. With each passing time-step, the \textcolor{ruby}{old frame} is discarded while intermediary frames are cached and prepended to the \textcolor{amazonite}{new frame}. To process each time-step, many DNNs (\eg, 3D CNN or Transformer-based networks) require the full frame-sequence as input. This entails considerable computational redundancy.
    }
    \label{fig:sliding-window}
\end{figure}

Yet, naive attempts to perform online recognition with the recent dominant designs, 3D CNNs and Transformers, are both inelegant and computationally inefficient. The issue relates to the input format required by 3D CNNs, which expect a spatio-temporal clip as input, and Transformers, which process sets of tokens. 
Consequently, they are not able to handle single frames but must aggregate data within a sliding window and pass full spatio-temporal clips or token sets to make predictions. This process is illustrated in \cref{fig:sliding-window}.

Besides the additional complexity of input caching, sliding-window processing entails considerable hidden computational redundancy, due to repeat computations of mutual interactions of weight and input data.
While RNN-based methods do not suffer this inefficiency, their predictive capabilities are often inferior, and 3D CNN and Transformer methods may still provide better computational/predictive trade-offs.
Even in online action detection~\cite{geest2016online}, \ie, per-frame action recognition where methods do not peek into the future, recent works have favored Transformer-based models~\cite{wang2021oadtr, xu2021long} despite their inability to process one frame at a time without the use of sliding-window processing.

This leads to the central research questions of this chapter:
\begin{itemize}
    \item[RQ1]\emph{How can we accelerate state-of-the-art deep neural networks, which were tailored for offline batch-processing of time-series data, to perform online stream processing efficiently?}
\end{itemize}

\section{Related works}
Multiple recent works have proposed specialty architectures, which can efficiently process input streams.
\textit{Recurrent Convolutional Units}~\cite{singh2019recurrent} replace 3D kernels with a 2D kernel to model the current image alongside a temporal 1D convolution to aggregate temporal context.
The \textit{Temporal Shift Module}~\cite{lin2019tsm} augments prior 2D CNNs by shifting a portion of channels along the temporal axis and fusing temporal information within residual connections.
\textit{MoviNets}~\cite{kondratyuk2021movinet} use stream buffers to cache activations from previous frames and aggregate them with causal 3D convolutions to achieve efficient stream processing. 
While the utilization of stream buffers alongside 3D convolution has similarities to the concurrently published \textit{Continual 3D Convolutions} proposed in paper \ref{pap:co3d}, the stream versions of MoviNets are not weight-compatible with regular 3D CNNs.
\textit{Dissected 3D CNNs}~\cite{kopuklu2022dissected} use residual connections to pass temporal information to the subsequent time-step and aggregate information over the temporal dimension via 3D convolutional kernels with temporal kernel size of two, fusing information from the prior time-step and the current one. 

The \textit{Streaming Transformer}~\cite{wang2021streaming} processes segments of an input stream (\ie, multiple time-steps at a once) and augments key and value matrices with the keys and values from prior time-steps without letting gradient flow back into prior processed segments. The tokens in the current segment thus attend mutually to one another, while tokens from prior steps are causally attended to without generating output tokens themselves. 
A similar approach is used for the training of Transformer-XL~\cite{dai2019transformer}, where causal scaled dot-product attention (SDA) is also processed in segments and gradients are restricted to flow in the current segment only. Since the employed self-attention is causal (\ie, only newer tokens attend to older ones), Transformer-XL can process tokens sequentially in a stream if appropriate caching is used between time-steps. Moreover, the SDA operation is modified with a relative positional encoding scheme.

The \textit{Continual Single-output SDA} proposed in paper~\ref{pap:cotrans}, which is a simplification of the full \textit{Continual Retroactive SDA}, operates in a similar manner to the Transformer-XL attention. However, it retains backward compatibility with regular SDA~\cite{vaswani2017attention} by utilizing \textit{Recycling Positional Encoding} as its positional encoding scheme.

None of the described related works provide backwards compatibility with basic 3D CNNs and Transformer architectures. 
In contrast, the \textit{Continual Inference Networks} proposed in papers \ref{pap:co3d}, \ref{pap:costgcn}, and \ref{pap:cotrans}, and are computational extensions of basic 3D CNNs, Transformers, and ST-GCNs~\cite{yan2018stgcn}, which retain weight-compatibility and enable efficient online stream processing in addition to the traditional offline processing.

\section{Continual Inference Networks}\label{sec:cin}

\begin{figure}[bt]
     \centering
     \begin{subfigure}[b]{\textwidth}
         \centering
         \includegraphics[width=0.7\textwidth]{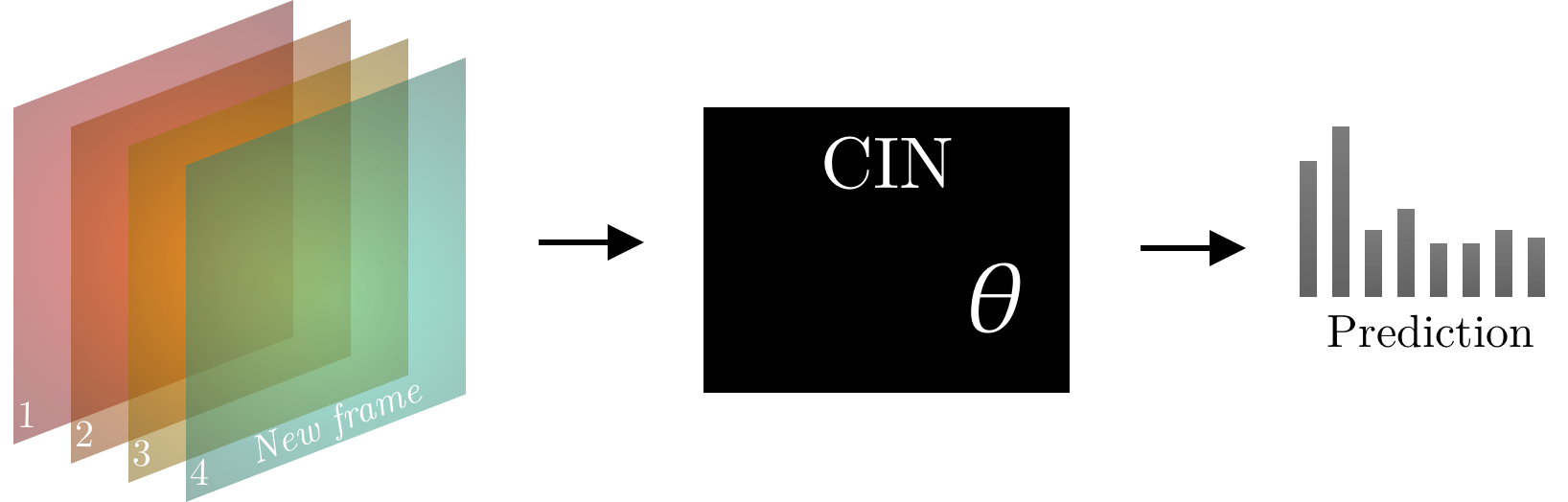}
         \caption{Offline processing}
         \label{fig:cin-offline}
     \end{subfigure}
     \\
     \hspace{1em}
     \begin{subfigure}[b]{\textwidth}
         \centering
         \includegraphics[width=0.7\textwidth]{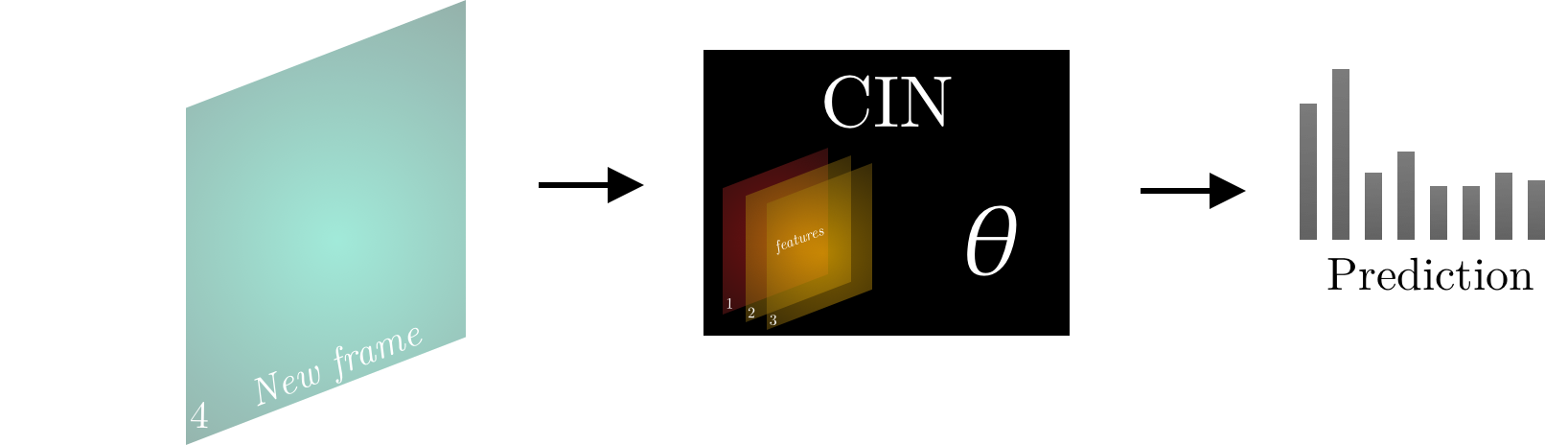}
         \caption{Online processing}
         \label{fig:cin-online}
     \end{subfigure}
        \caption{Continual Inference Network (CIN) used for (a) offline processing of a spatio-temporal batch and (b) online processing of an input stream, where intermediary frame features are cached internally. Given identical inputs, a CIN can use one set of learned weights $\theta$ in either situation and produces identical results.}
\end{figure}

\subsection{Definition}\label{sec:cin-def}
Continual Inference Networks (CINs) spring from a desire to continue the fruitful competition on offline prediction benchmarks while retaining the ability to efficiently process a continual input stream without predefined beginning and end. 
Formally, we can define CINs to encompass network architectures which uphold the ideal:
{
\begin{enumerate}
    \item[\textit{Online inference}:] A CIN can process a continual input stream time-step by time-step, producing a prediction for each step, without computational redundancy (\cref{fig:cin-online}).
    \item[\textit{Offline inference}:] A CIN can process a spatio-temporal input similarly to a regular DNN (\cref{fig:cin-offline}). 
    \item[\textit{Identical results}:] A CIN produces identical results during online and offline inference given identical inputs.
    \item[\textit{Identical weights}:] A CIN uses one set of trainable parameters for online and offline inference.
\end{enumerate}
}

\subsection{Relation to recurrent neural networks}
\begin{figure}[b]
     \centering
     \begin{subfigure}[b]{\textwidth}
         \centering
         \includegraphics[width=0.8\textwidth]{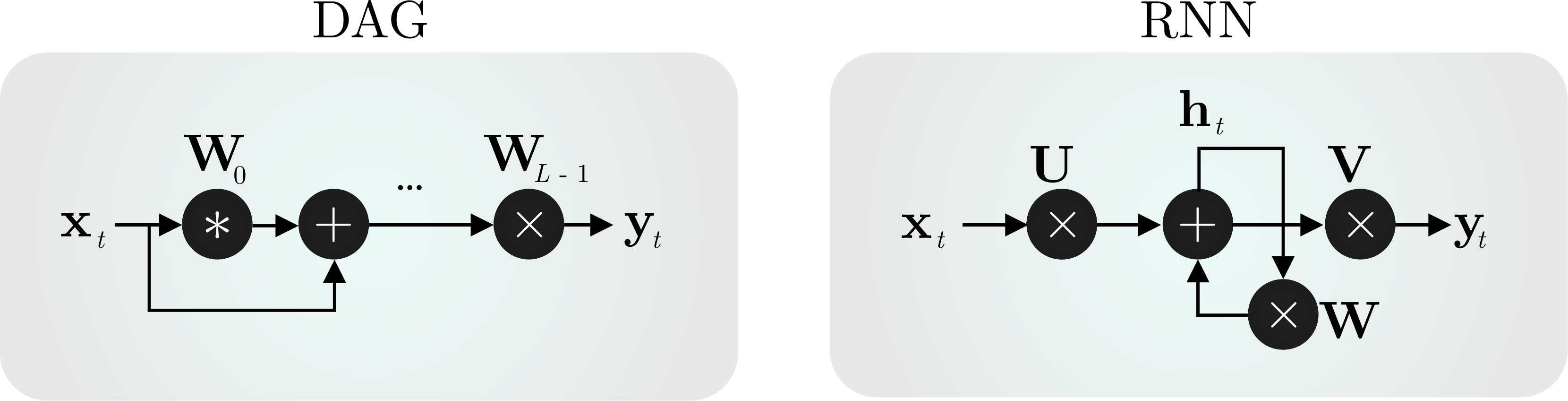}
         \caption{Continual Inference Networks}
         \label{fig:cin-dag-and-rnn}
     \end{subfigure}
     \\
     \hspace{1em}
     \begin{subfigure}[b]{\textwidth}
         \centering
         \includegraphics[width=0.8\textwidth]{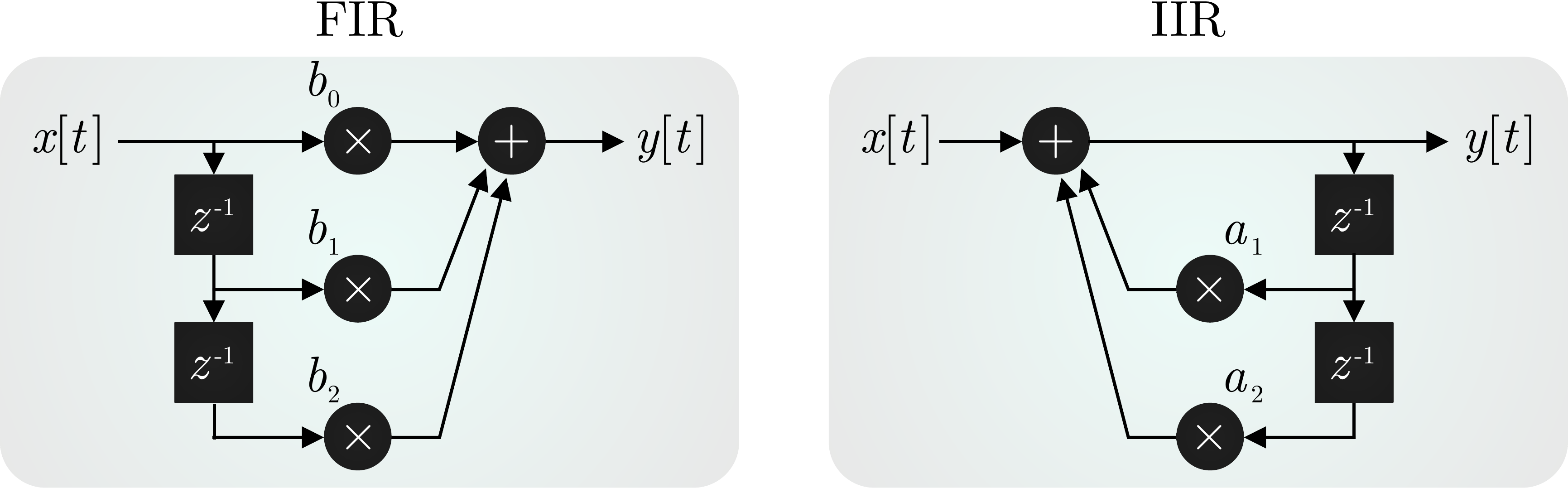}
         \caption{Discrete-time filters}
         \label{fig:signal-processing-filters}
     \end{subfigure}
        \caption{Continual Inference Networks are multi-modal discrete-time filters with learned weights. 
        Specifically, directed acyclic graph (DAG) structured networks are finite impulse response (FIR) filters and recurrent neural networks (RNNs) are infinite impulse response (IIR) filters. }
\end{figure}

The notion of a Continual Inference Network encompasses all networks, which can operate sequentially, producing outputs for each observed step without peeking into the future. This includes uni-directional recurrent neural networks (RNNs) as well as causal neural networks with directed acyclic graph (DAG) structure (\cref{fig:cin-dag-and-rnn}). DAG neural networks include feed-forward neural networks, CNNs, and Transformers, which operate locally within a time-step, as well as \textit{Continual} CNNs (see Sections \ref{sec:co3d} and \ref{sec:costgcn}) and \textit{Continual} Transformer Encoders (see \cref{sec:cotrans}) with spatio-temporal receptive fields.

\subsection{Relation to digital signal processing}
In signal processing, digital discrete-time filters are widely used to remove unwanted frequency components of an input signal. In particular, the signal processing literature~\cite{lyons2011understanding} distinguishes between two filter types, categorized by their response to a Kronecker delta: Finite impulse response (FIR) filters have an impulse response with finite duration, after which it settles at zero; infinite impulse response (IIR) filters have an impulse response, which continues indefinitely. 
Mechanistically, FIR filters consist of multiplicative forward coefficients ($b$) and tapped delays (noted as $z^{-1}$ using z-domain notation), which are arranged as a DAG. 
IIR filters, on the other hand, contain feedback cycles with multiplicative coefficients ($a$). These are illustrated in \cref{fig:signal-processing-filters}.

Continual Inference Networks can be viewed as large discrete-time filters with learned coefficients.
In particular, RNN-structured CINs are IIR filters while DAG-structured CINs are FIR filters.

\subsection{Architectural considerations}\label{sec:co-arch-considerations}

A wide range of existing network architectures either satisfy or can be modified to satisfy the CIN criteria outlined in \cref{sec:cin-def}. In particular, the temporal (re)distribution of computations plays a crucial role. Moreover, zero-padding must be used with care. 

\subsubsection{Add delays to satisfy causality}
CINs operate with a linear progression of time and satisfy \textit{causality}, \ie, they do not peek into the future.
While many DNNs designed for offline processing do not satisfy this, surgical placement of tapped delays can rectify the issue.
The \textit{Continual 3D CNNs} proposed in paper~\ref{pap:co3d} demonstrate this by delaying partially convolved frame-features before summing them at the appropriate time-step.
In general, a network module with temporal kernel size $K_T$, dilation $D_T$, and zero-padding $P_T$ has a delay of
\begin{equation}
    \underbrace{K_T + (K_T - 1)(D_T-1)}_{\text{Receptive field}} - P_T - 1.
    \label{eq:cin-delay}
\end{equation}

\subsubsection{Remove end-padding in the temporal dimension}
\begin{figure}[tb]
     \centering
     \includegraphics[width=0.5\textwidth]{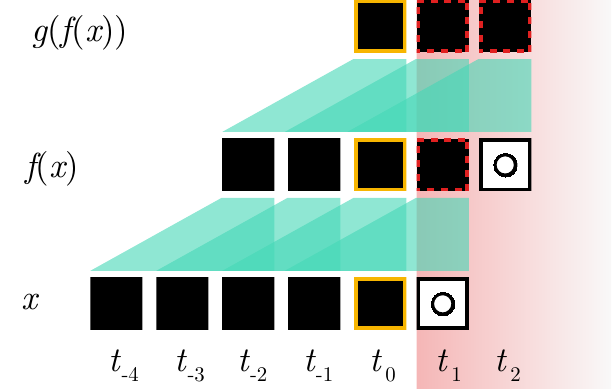}
     \caption{Temporal end-padding $\boxcircle$ in neural network layers with receptive field of size \textcolor{amazonite}{$\blacktriangle$} results in acausal outputs (red dotted boxes \textcolor{ruby}{$\square$}). The features corresponding to the current time-step $t_0$ are highlighted yellow \textcolor{citrine}{$\square$}.}
     \label{fig:cin-end-padding}
\end{figure}

Many CNNs use zero-padding to avoid shrinking features maps. 
However, zero padding of the temporal edge in the direction of most recent data corresponds to observing zeros in the future. Consider the situation depicted in \cref{fig:cin-end-padding} where a multi-layer neural network conducting online inference processes an input feature at time $t_0$. 
The $t_0$ input feature results in one intermediary feature-map after layer $f$ and one output feature after layer $g$ (yellow highlight). If temporal end-padding is used with each layer, additional \textit{acausal} outputs (red dotted boxes) will be computed. To make matters worse, the extra acausal computations accumulate with each zero-padding added throughout the network! When the subsequent time-step arrives, all the acausal features become obsolete and must be computed anew for the next padded zeros. Causal features, on the other hand, can be cached and reused. 
Accordingly, CINs should not use temporal end-padding. 

Empirically, it was observed in paper~\ref{pap:co3d} that weights from networks which were trained with temporal end-padding can be reused for efficient stream processing in corresponding CINs without end-padding. 
This, however, introduced a \textit{model shift}, \ie a deviation of the model operation, with lead to slight accuracy deterioration.

\subsubsection{Residual connections}
\begin{figure}[tb]
     \centering
     \begin{subfigure}[b]{0.46\linewidth}
         \centering
         \includegraphics[width=0.98\textwidth]{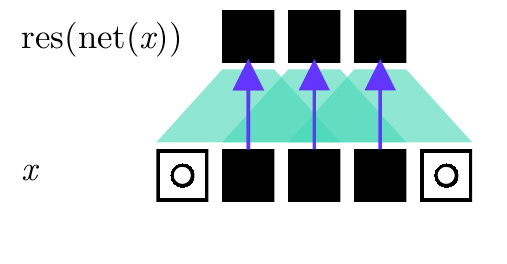}
         \caption{Offline inference}
         \label{fig:residual-forward}
     \end{subfigure}
     \hfill
     \begin{subfigure}[b]{0.46\linewidth}
         \centering
         \includegraphics[width=0.98\textwidth]{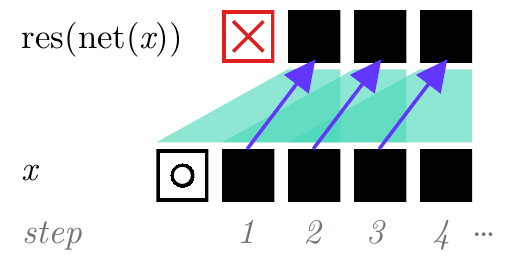}
         \caption{Online inference}
         \label{fig:residual-step}
     \end{subfigure}
        \caption{Residual connections \textcolor{ioite}{↑} over a module with receptive field of size \textcolor{amazonite}{$\blacktriangle$} and padding one (``equal padding'') $\boxcircle$. \textcolor{ruby}{$\boxtimes$} are empty outputs. This illustration first appeared in paper \ref{pap:colib}.}
        \label{fig:residual}
\end{figure}

The above-mentioned architectural consideration has particularly interesting implications for the residual connection of temporal CNNs. Commonly, residuals are employed in the context of an ``equal padding'' strategy, \ie, where padding is used to retain the temporal feature shape after the layer. This is illustrated in \cref{fig:residual-forward}.
If we remove the end-padded zero, the alignment of the residual shifts: The input of a time-step 1 must be added to the output in time-step 2 (see \cref{fig:residual-step}). This corresponds to delaying the residual connection by the same number of steps as removed zeros.
Residual connections should generally match the delay of the network module they wrap, as described by \cref{eq:cin-delay}.

\subsubsection{Temporal stride}
CINs are designed to perform online processing on continual streams, where one output is produced for each input step.
However, some network layers, \eg, convolution and pooling, are occasionally used with temporal stride larger than one. 
While this can be used to \textit{reduce} the computational complexity for full-sequence processing in a regular network due to the reduced temporal dimension of subsequent feature-maps, it \textit{increases} the per-prediction complexity of CINs.
In fact, each network layer with a stride larger than one corresponds to a temporal down-sampling operation. 

Generally, the stride of a neural network, $S_{NN}$, aggregates multiplicatively throughout the $L$ layers of a network:
\begin{equation}
    S_{NN} = \prod_{l=1}^{L} S_l.
\end{equation}
The network prediction rate, $R_{NN}$, is given by the inverse stride:
\begin{equation}
    R_{NN} = 1 / S_{NN}.
\end{equation}

\subsection{Training}
Since CINs can operate efficiently on both spatio-temporal inputs and on step sequences, either can be used for training. Most prior methods train on datasets with spatio-temporal inputs. Full sequence training is recommended for CINs as well.

Should step-wise training be desired, nonetheless, two modifications are recommended to match the full sequence training. First, the normalization layer momentum parameters should be adjusted to accommodate the smaller statistical sample provided by a step compared to a full clip. To match the running statistics, the step momentum $M_\text{step}$ should be updated to
\begin{equation}
    M_\text{step} = \frac{2}{\left(L * \left(2 / M_\text{seq} - 1\right) + 1\right)},
\end{equation}
where $M_\text{seq}$ is the momentum during full sequence processing, and $L$ is the length of the sequence observed by the layer.
Second, the drop-out mask should be fixed during the steps originating from one sequence to avoid within-sequence changes in network configuration.

\section{Continual 3D Convolutional Neural Networks}\label{sec:co3d}
\subsection{Related works}
3D CNNs are a widely used architecture for clip-based spatio-temporal perception tasks such as human activity recognition. 
Specifically, networks such as I3D~\cite{carreira2017quo}, R(2+1)D~\cite{tran2018closer}, SlowFast~\cite{feichtenhofer2019slowfast}, and X3D~\cite{feichtenhofer2020x3d} each pushed the state-of-the-art on the large-scale video-classification dataset Kinetics-400~\cite{kay2017kinetics}.

Despite efficiency-focused innovations introduced in recent works~\cite{tran2018closer, feichtenhofer2019slowfast, feichtenhofer2020x3d}, 3D CNNs remained fundamentally hamstrung in the context of online-recognition, where a prediction is desired for each input frame. This stems from the convolution implementation in deep learning tool-kits such as PyTorch~\cite{paszke2019pytorch} and TensorFlow~\cite{tensorflow2015whitepaper}, where the full input sequence is processed at once during the forward pass. 
On the other hand, convolutions are routinely implemented as FIR filters for digital signal processing of continual, uni-dimensional input streams (see \cref{fig:signal-processing-filters} left).
A similar idea can be used for multi-dimensional 3D convolutions in the context of DNNs.

\subsection{Convolutions for processing of continual input streams}
Consider an input $\mathbf{X}_t \in \mathbb{R}^{H\times W}$ sampled from a time-series $\mathbf{X} = \left[\mathbf{X}_t : t \in 0..T-1\right] \in \mathbb{R}^{T \times H \times W}$ and a 3D convolutional kernel $\mathbf{W} \in \mathbb{R}^{K_T \times K_H \times K_W}$ with time, height, width, and kernel dimensions $K$.
Without loss of generality, we will omit the channel dimensions of input and convolutional kernel. 
A three-dimensional convolution between the sequence $\mathbf{X}$ and kernel $\mathbf{W}$ is defined as:
\begin{equation}
    \left({W} * {X}\right)^{(t, h, w)} \overset{\operatorname{def}}{=} \sum_{k_t = 0}^{K_T-1} \sum_{k_h=0}^{K_H-1} \sum_{k_w=0}^{K_W-1}
        {W}^{(k_t, k_h, k_w)} {X}^{(t - k_t, h - k_h, w - k_w)}.
\end{equation}
The feature-map $\mathbf{Y}^{(t)}$ at time-step $t$ can be computed by summation of the spatial convolutions:
\begin{equation}
    \mathbf{Y}^{(t)} = \left(\mathbf{W} * \mathbf{X}\right)^{(t)} = \sum_{k_t = 0}^{K_T-1} 
        \mathbf{W}^{(k_t)} * \mathbf{X}^{(t - k_t)}.
\end{equation}

\begin{figure}[tb]
     \centering
     \begin{subfigure}[b]{0.46\linewidth}
         \centering
         \includegraphics[width=0.98\textwidth]{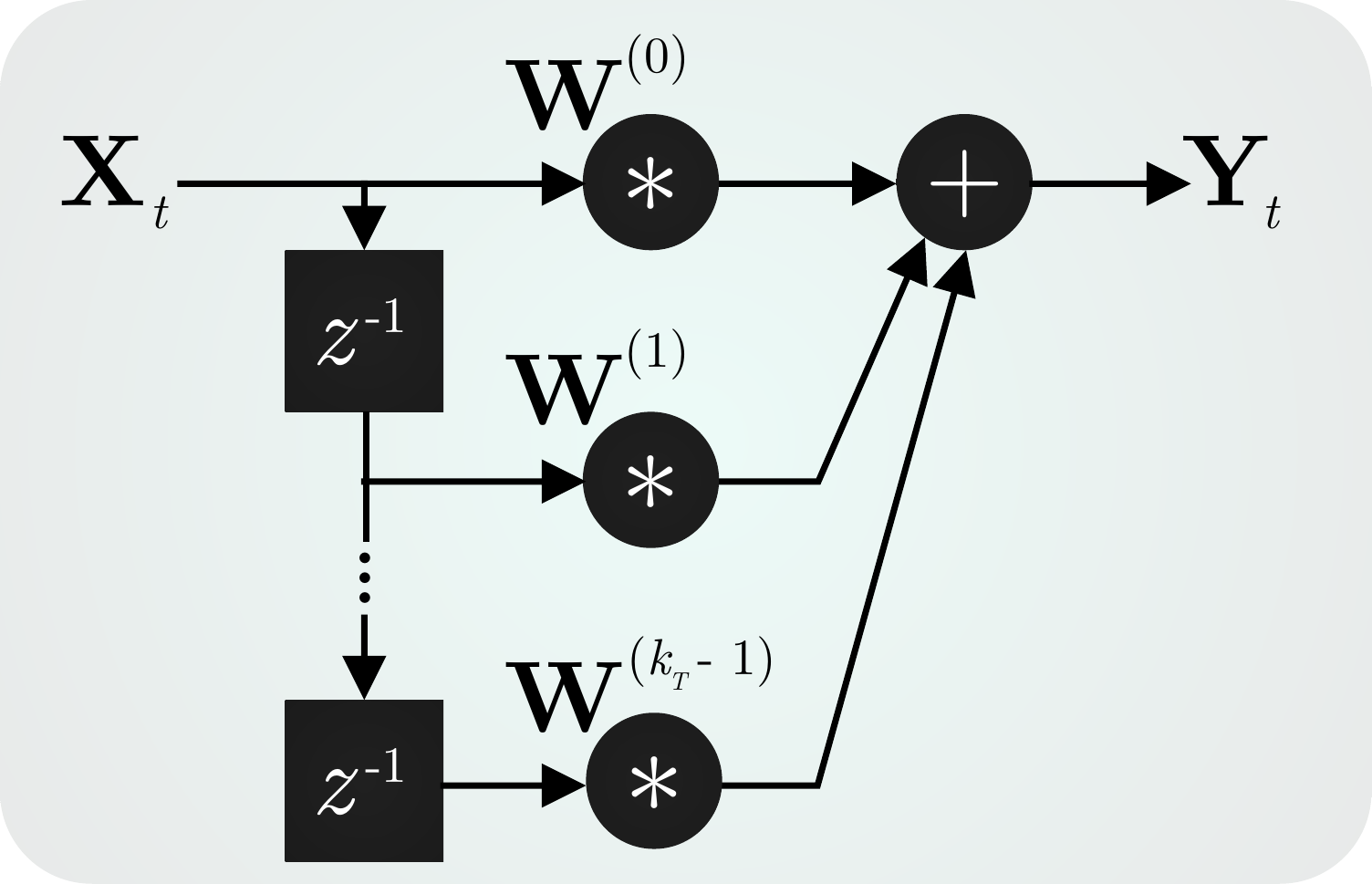}
         \caption{Pre-cached}
         \label{fig:coconv-pre}
     \end{subfigure}
     \hfill
     \begin{subfigure}[b]{0.46\linewidth}
         \centering
         \includegraphics[width=0.98\textwidth]{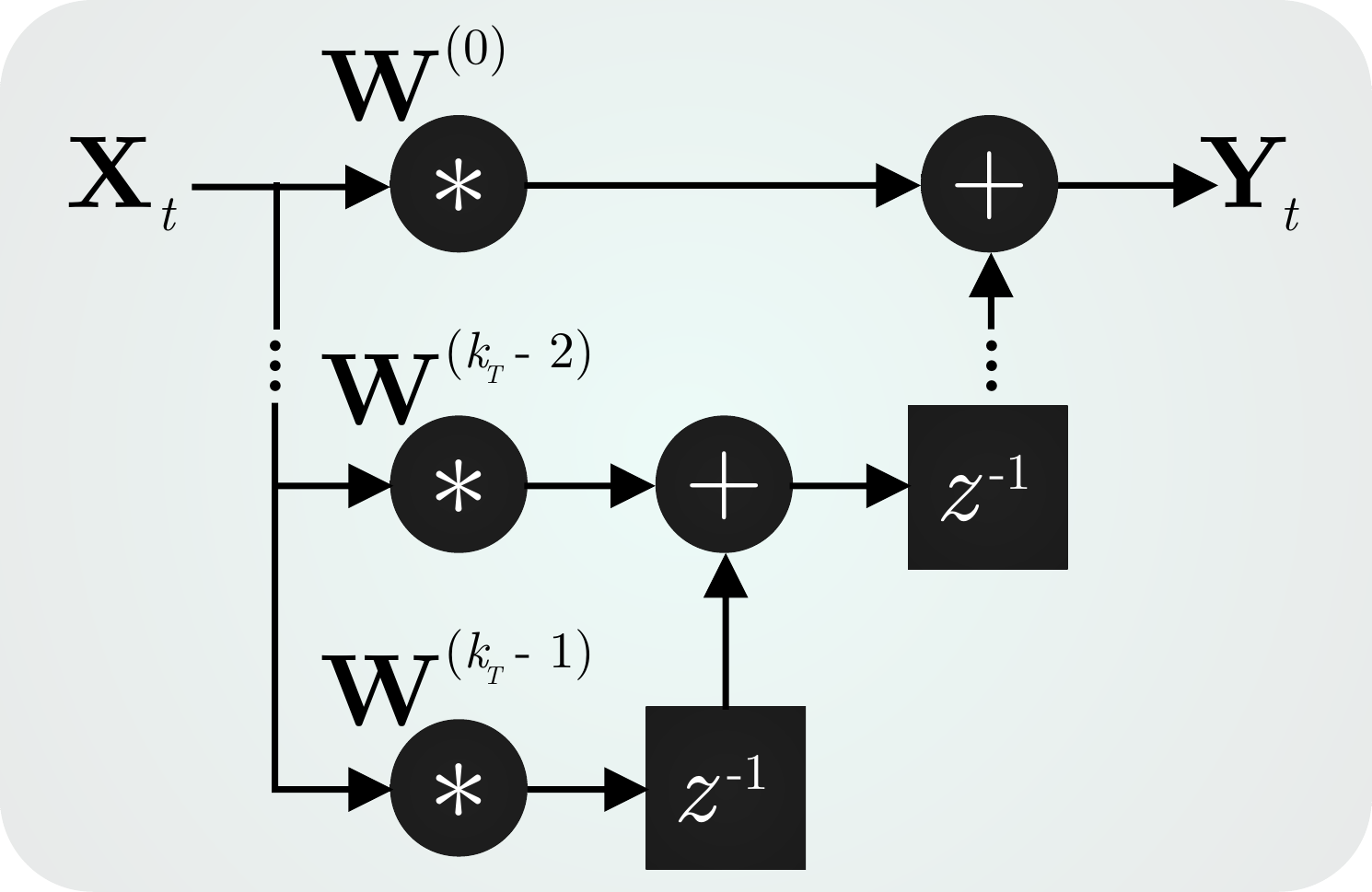}
         \caption{Post-cached}
         \label{fig:coconv-post}
     \end{subfigure}
        \caption{Continual convolutions can be modeled as digital finite impulse response filters with temporal caching (a) pre or (b) post spatial convolutions of a temporally sliced kernel $\mathbf{W}^{(i)} \in \mathbb{R}^{K_H \times K_W}, i \in 0..K_T - 1$.}
        \label{fig:co-conv}
\end{figure}

The above can be extended to any spatial convolution, including dilated, strided, padded, and grouped convolutions, and holds for continual input streams where $T\rightarrow \infty$.
\textit{Continual convolutions} for DNNs, \ie, convolutions for processing of a continual input stream, can be efficiently implemented through the use of appropriate caching, similar to a FIR filter for digital signal processing, but with multi-dimensional signals and spatial convolutions instead of multiplication by simple filter coefficients.
There are two implementation options for this.
\begin{enumerate}[(a)]
    \item \textit{Pre-cached}: Prior input values can be cached first and spatially convolved with the kernel later when all required time-steps have been observed (see \cref{fig:coconv-pre}). This corresponds to a \textit{direct form} FIR filter implementation.
    \item \textit{Post-cached}: Individual inputs steps are eagerly convolved with the kernel, cached, and progressively summed up (see \cref{fig:coconv-post}). This corresponds to a \textit{transposed form} FIR filter implementation.
\end{enumerate}

The floating-point operations are identical in both implementation options, but the memory requirements can vary.
If the convolutional kernel $\mathbf{W}\in \mathbb{R}^{C_{in} \times C_{out} \times K_T \times \cdots}$ performs an up-projection (\ie, $C_{in} > C_{out}$), the pre-cached structure gives the smallest cache size. Similarly, post-cached structures are more memory-efficient for down-projections.

To enable prior state-of-the-art 3D CNNs to operate efficiently during stream processing, the complete architecture requires a modified implementation. \textit{Cf.} \cref{sec:co-arch-considerations}, this includes delaying residuals appropriately, removing end-padding, as well as modifying average pooling operations. The latter is easily achieved by pooling each frame, caching the appropriate receptive field, and updating a running average wherein the newest frame is added and oldest the frame is subtracted in each time-step.
For additional details, please see paper~\ref{pap:co3d}.

\subsection{Experiments}

To validate the efficacy of reusing prior 3D CNNs for efficient stream processing by adapting the principles of Continual Inference Networks (\cref{sec:cin}), we implemented \textit{continual} versions of I3D, Slow, and X3D-\{S, L, M\} (denoted \textit{Co}I3D, etc.) and benchmarked them on Kinetics-400 using pre-trained weights from corresponding regular 3D CNNs. 
The regular 3D CNNs, which served as our baselines, were evaluated with sliding-window processing, \ie by processing full spatio-temporal clips in the usual manner. 
The \textit{Continual} 3D CNNs were initialized with frames corresponding to their receptive field minus one frame, and evaluated on the next frame.
In addition to accuracy, we report parameter count, GPU memory consumption, FLOPs, as well as throughput on the CPU of a MacBook Pro 16" 2019 2.6 GHz i7 processor, NVIDIA TX2 and Xavier embedded GPUs, and an NVIDIA RTX 2080Ti GPU. 

\begin{figure}[b]
    \centering
    \includegraphics[width=0.75\linewidth]{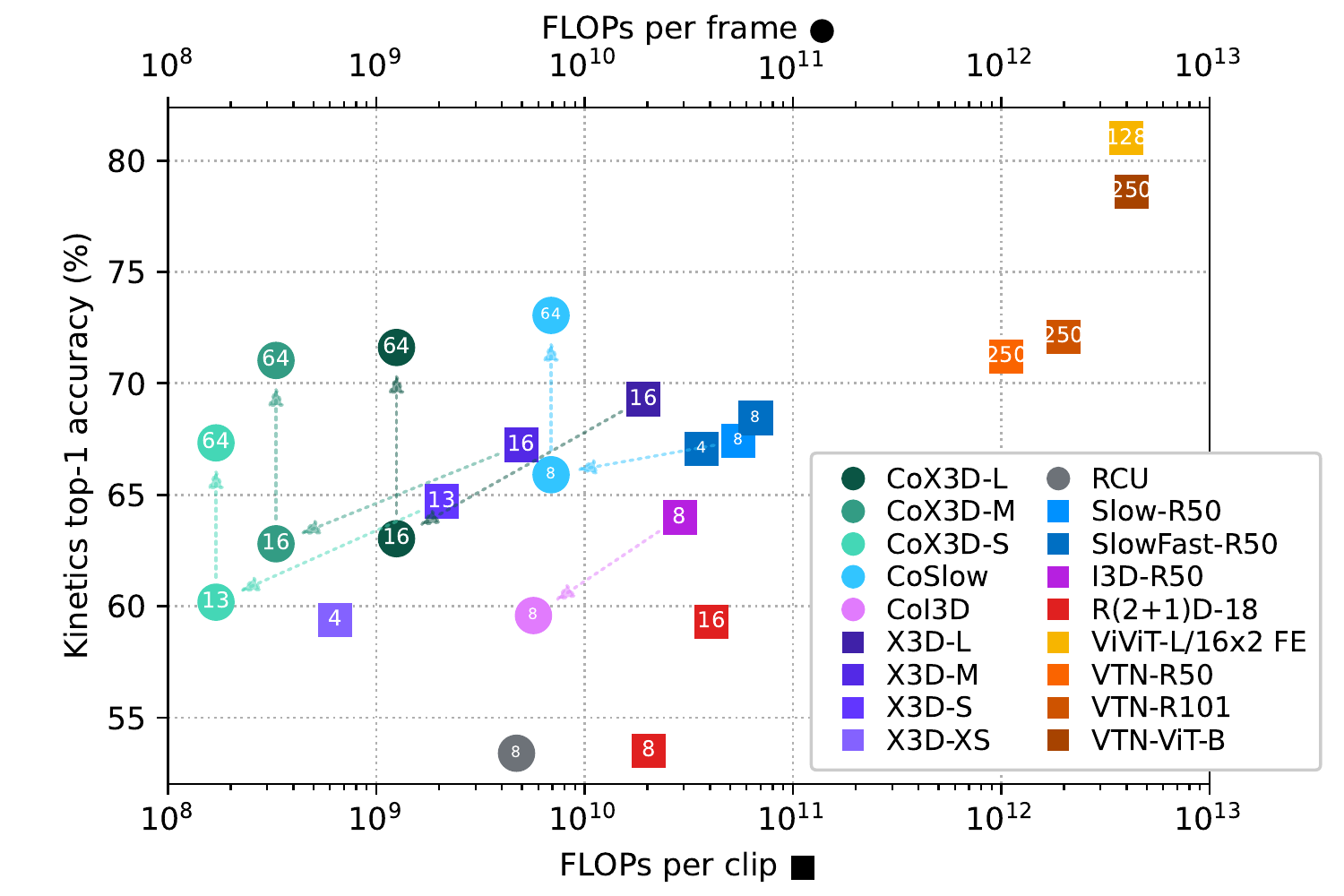}
    \caption{
        {Accuracy/complexity trade-off} for Continual 3D CNNs and recent state-of-the-art methods on Kinetics-400 using 1-clip/frame testing. 
        $\ssquare$~FLOPs per \textit{clip} are noted for regular networks, while $\sbullet$~FLOPs per \textit{frame} are shown for the Continual 3D CNNs. 
        Frames per clip / global average pool size is noted in the representative points.
        Diagonal and vertical arrows indicate a direct weight transfer from regular to Continual 3D CNN and an extension of the receptive field.
        The figure first appeared in paper~\ref{pap:co3d}~\cite{hedegaard2022co3d}.
    }
    \label{fig:co3d-test-acc-vs-flops}
\end{figure}

\begin{table}[t]
\begin{center}
\resizebox{\textwidth}{!}{
\begin{tabular}{llrrrrrrrr}
    \toprule
    &\multirow{2}{*}{\textbf{Model}} 
    &\textbf{Acc.}
    &\textbf{Par.}
    &\textbf{Mem.}  
    &\textbf{FLOPs}
    &\multicolumn{4}{c}{\textbf{Throughput} (preds/s)}
        \\ \cline{7-10}
        && (\%) & (M) & (MB) & (G) & CPU & TX2  & Xavier  &2080Ti
    \\
    \midrule
    \parbox[t]{1mm}{\multirow{9}{*}{\rotatebox[origin=c]{90}{Clip}}}                  
    & I3D-R50                           & 63.98     & 28.04    & 191.59    & 28.61      & 0.93   & 2.54     & 9.20      & 77.15  \\
    & R(2+1)D-18$_8$                    & 53.52     & 31.51    & 168.87    & 20.35      & 1.75   & 3.19     & 6.82      & 130.88 \\
    & R(2+1)D-18$_{16}$                 & 59.29     & 31.51    & 215.44    & 40.71      & 0.83   & 1.82     & 3.77      & 75.81  \\
    & Slow-8×8-R50                      & 67.42     & 32.45    & 266.04    & 54.87      & 0.38   & 1.34     & 4.31      & 61.92  \\
    & SlowFast-8×8-R50                  & 68.45     & 66.25    & 344.84    & 66.25      & 0.34   & 0.87     & 2.72      & 30.72  \\ 
    & SlowFast-4×16-R50                 & 67.06     & 34.48    & 260.51    & 36.46      & 0.55   & 1.33     & 3.43      & 41.28  \\
    & X3D-L                             & 69.29     & 6.15     & 240.66    & 19.17      & 0.25   & 0.19     & 4.78      & 36.37  \\
    & X3D-M                             & 67.24     & 3.79     & 126.29    & 4.97       & 0.83   & 1.47     & 17.47     & 116.07 \\
    & X3D-S                             & 64.71     & 3.79     & 61.29     & 2.06       & 2.23   & 2.68     & 42.02     & 276.45 \\
    & X3D-XS                            & 59.37     & 3.79     & 28.79     & 0.64       & 8.26   & 8.20     & 135.39    & 819.87 \\
    \midrule
    \parbox[t]{1mm}{\multirow{11}{*}{\rotatebox[origin=c]{90}{Frame}}} & RCU$_8$~\cite{singh2019recurrent}     & 53.40     & 12.80     & - & 4.71 & - & - & - & - \\
    & \textit{Co}I3D$_{8}$              & 59.58     & 28.04    & 235.87    & 5.68       & 3.00   & 2.41     & 14.88     & 125.59 \\
    & \textit{Co}I3D$_{64}$             & 56.86     & 28.04    & 236.08    & 5.68       & 3.15   & 2.41     & 14.89     & 126.32 \\
    & \textit{Co}Slow$_{8}$             & 65.90     & 32.45    & 175.98    & 6.90       & 2.80   & 1.60     & 6.18      & 113.77 \\
    & \textbf{\textit{Co}Slow$_{64}$}    & \textbf{73.05}  & \textbf{32.45}  & \textbf{176.41}  & \textbf{6.90}   & \textbf{2.92}   & \textbf{1.60}  & \textbf{6.19} & \textbf{102.00}  \\
    & \textit{Co}X3D-$\text{L}_{16}$    & 63.03     & 6.15     & 184.29    & 1.25       & 2.30   & 0.99     & 25.17     & 206.65 \\
    & \textbf{\textit{Co}X3D-$\text{L}_{64}$}    & \textbf{71.61}     & \textbf{6.15}    & \textbf{184.37}     & \textbf{1.25}      &  \textbf{2.30}   & \textbf{0.99}       & \textbf{27.56}       & \textbf{217.53}  \\
    & \textit{Co}X3D-$\text{M}_{16}$    & 62.80     & 3.79    & 68.88     & 0.33      & 7.57   & 7.26       & 88.79      & 844.73  \\
    & \textbf{\textit{Co}X3D-$\text{M}_{64}$}    & \textbf{71.03}     & \textbf{3.79}    & \textbf{68.96}    & \textbf{0.33}      & \textbf{7.51}   & \textbf{7.04}       & \textbf{86.42}      & \textbf{796.32}  \\
    & \textit{Co}X3D-$\text{S}_{13}$    & 60.18     & 3.79    & 41.91     & 0.17      & 13.16  & 11.06      & 219.64      & 939.72 \\
    & \textbf{\textit{Co}X3D-$\text{S}_{64}$}    & \textbf{67.33}     & \textbf{3.79}    & \textbf{41.99}     & \textbf{0.17}      & \textbf{13.19}  & \textbf{11.13}     & \textbf{213.65}      & \textbf{942.97} \\
    \bottomrule
\end{tabular}
}
\end{center}
\caption{
    Kinetics-400 benchmark results. The noted accuracy is the single clip or frame top-1 score using RGB as the only input-modality. 
    The performance was evaluated using publicly available pre-trained models without any further fine-tuning. 
    For throughput comparison, predictions per second denote frames per second for the \textit{Co} models and clips per second for the remaining models. Throughput results are the mean of 100 measurements. 
    Pareto-optimal models are marked with bold.
    Mem. is the maximum allocated memory during inference noted in megabytes.
    The table first appeared in paper~\ref{pap:co3d}~\cite{hedegaard2022co3d}.
}
\label{tab:benchmark-kinetics400}
\end{table}

The benchmark results are presented in \cref{tab:benchmark-kinetics400} and visually depicted for the accuracy/FLOPs trade-off in \cref{fig:co3d-test-acc-vs-flops}.
Here, clip-sizes are illustrated with subscripts and within the illustrated data points. 
Across all tested CINs, we observed impressive FLOPs reductions and increases in throughput. These were generally proportional to the clip size of models (\eg, 7.95$\times$ fewer FLOPs for \textit{Co}Slow compared to Slow, which had a clip size of 8, and 15.34$\times$ fewer for \textit{Co}X3D-L compared to X3D-L, which had a clip size of 16), but with lower relative throughput increases than FLOPs reductions due the to runtime overhead of reading and writing cached features. 
While the parameter counts for continual and regular 3D CNNs were identical, the maximum allocated GPU memory was lower for most CINs despite their use of feature-caching. 
This can be explained by the larger intermediary feature-maps handled by regular networks compared to CINs, which did not include the temporal dimension.
Finally, the test accuracies of CIN network versions were slightly reduced due to the model shift incurred by the missing end-padding with zeros. Since the computational cost of increasing the temporal receptive field of a CIN is negligible, the lost accuracy can be regained by increasing the receptive field for the average pooling layer residing as the penultimate network layer in all the considered networks. Networks with larger average pooling size of $64$ frames are denoted accordingly in \cref{tab:benchmark-kinetics400} and \cref{fig:co3d-test-acc-vs-flops}.
Additional details as well as ablation studies of network initialization and global average pooling sizes, as well as results on the Charades dataset~\cite{sigurdsson2016charades} are available in paper~\ref{pap:co3d}.

\section{Continual Spatio-temporal Graph Convolutional Networks}\label{sec:costgcn}
\subsection{Human actions as a skeleton sequence}
\begin{figure}[tb]
    \centering
    \includegraphics[width=0.75\linewidth]{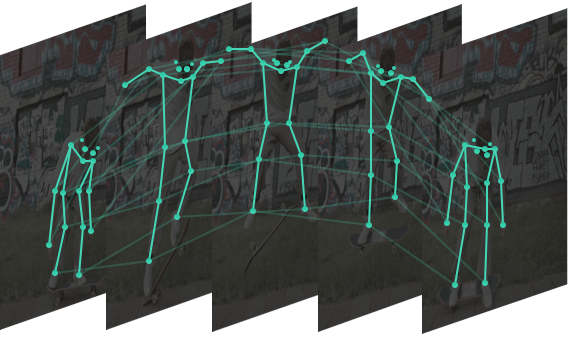}
    \caption{
        Sequence of human body poses modeled as a spatio-temporal graph.
    }
    \label{fig:skeletons}
\end{figure}

Instead of predicting human action categories directly from raw RGB video clips, another approach explicitly models human body poses for each frame in a sequence and utilizes those for prediction of action labels. 
Specifically, a body pose can be extracted with pose estimation tools such as OpenPose~\cite{cao2019openpose} and modeled as a skeleton graph $\mathcal{G} = (\mathcal{V} ,\mathcal{E})$ where body joint positions are nodes $\mathcal{V}$ and limbs are edges $\mathcal{E}$. 
To construct a spatio-temporal graph, joints are connected across time-steps in the skeleton sequence as well. This is illustrated in \cref{fig:skeletons}.

\subsection{Related works}
Early works on skeleton-based action recognition did not use the graph representation, but either made pseudo-images from rearranged body joint positions, which were subsequently processed by a CNN~\cite{ke2017new}, or trained RNN models on feature-vectors comprising concatenated joints~\cite{shahroudy2016ntu, zhang2017view}.
However, both of these representations discarded valuable information on the non-Euclidean skeleton structure.

Graph Convolutional Networks (GCNs), which are described in publication \ref{pap:chap4}, are a recently proposed class of neural networks for processing graph-structured data.
The Spatio-temporal Graph Convolutional Network (ST-GCN)~\cite{yan2018stgcn} was the first method to utilize GCNs for skeleton-based action recognition. 
Here, graph convolutions were used for spatial processing while regular convolutions aggregated the temporal information.
Multiple extensions of ST-GCN have been proposed, including 2s-AGCN~\cite{shi2019two}, which uses two input streams (\ie, one considering joint features and another considering bone features) and which enhances ST-GCN with a graph attention mechanism considering node similarity; 
DGNN~\cite{shi2019skeletondirected} modeled skeleton connections with a directed graph; S-TR~\cite{plizzari2021str} used self-attention layers instead of graph convolutions; and MS-G3D~\cite{liu2020disentangling} extracted long-range features with multi-scale graph convolutions.

The ST-GCN~\cite{yan2018stgcn} and multiple derived networks~\cite{shi2019two, plizzari2021str} employ spatial graph convolution (GC) and temporal convolution (TC) as separate network layers. In practice, GC varies among methods\footnote{See paper \ref{pap:costgcn} for exact formulations of graph convolution (GC) layers.}, and TC is implemented as a 2D convolution. 
A spatio-temporal graph convolution block which computes the feature $\mathbf{H}_l$ at layer $l$ takes the form:
\begin{equation}
    \mathbf{H}_l = \text{ReLU}\left(\text{Res}(\mathbf{H}_{l-1}) + \text{BN}(\text{TC}(\text{GC}(\mathbf{H}_{l-1})))\right),
    \label{eq:stgcn-spatio-temporal-block}
\end{equation}
where BN denotes Batch Normalization, and a residual connection (Res) performs a point-wise convolution when input and output channel dimensions vary. Multiple such blocks are stacked throughout the network, before a global average pooling layer and a prediction head aggregate the final prediction.

Various approaches have been explored to reduce the high computational complexity of ST-GCN-based methods. These include Neural Architecture Search~\cite{peng2020learning, heidari2021progressive}, selection of informative frames~\cite{negar2020tagcn}, and the use of channel shift operations in place of temporal convolutions~\cite{cheng2020skeleton, cheng2021extremely}.
However, both the ST-GCN and its derived methods suffer the same restriction as 3D CNNs during online processing: They require full skeletons sequences as their input and hence need a sliding window of $T$ cached input skeletons for stream processing.

\subsection{Continual Spatio-temporal Graph Convolutions}
During continual processing of an input stream, the network should produce temporal features for each time step $\mathbf{H}_{l}^{(t)}$ rather than for the whole sequence $\mathbf{H}_{l}$. 
This behavior can be achieved by reformulating the spatio-temporal graph convolution block (see \cref{eq:stgcn-spatio-temporal-block}) to conform with CIN requirements (\cref{sec:cin-def}).
The separation between the graph and temporal convolutions makes the transition easy: 
First, the regular temporal convolution is replaced with a continual temporal convolution (\textit{Co}TC), and second, the residual connection is delayed according to \cref{eq:cin-delay} based on the parameters of \textit{Co}TC. The continual spatio-temporal graph convolution block is given by
\begin{equation}
    \mathbf{H}_{l}^{(t)} = \text{ReLU}\left(\text{Delay}(\text{Res}(\mathbf{H}_{l-1}^{(t)})) + \text{BN}(Co\text{TC}(\text{GC}(\mathbf{H}_{l-1}^{(t)})))\right).
    \label{eq:costgcn-spatio-temporal-block}
\end{equation}
To construct the full \textit{Continual} ST-GCN variant, multiple such blocks are stacked, followed by a running global average pooling layer and a prediction head, which produces a prediction each time-step. 
The simplicity of the transformation from \cref{eq:stgcn-spatio-temporal-block} to \cref{eq:costgcn-spatio-temporal-block} exemplifies the utility of the CIN reformulation. 
While a considerable engineering effort is required to handle network delays and inputs shapes properly\footnote{The \textit{Continual Inference} library described in paper \ref{pap:colib} and \cref{sec:colib} alleviates the engineering effort substantially.}, the effect on the computational complexity of per-step prediction is dramatic:
Continual inference only requires the processing of a single frame with the network body and prediction head, \ie, $\mathcal{O}(Co\text{Net}) \approx \mathcal{O}(\text{Body}) + \mathcal{O}(\text{Head})$, whereas sliding-window processing requires the body to process the whole sequence of $T$ skeletons for each step, \ie, $\mathcal{O}(\text{Net}) \approx T \cdot \mathcal{O}(\text{Body}) + \mathcal{O}(\text{Head})$. 
For $\mathcal{O}(\text{Body}) \gg \mathcal{O}(\text{Head})$, which is the case for most neural networks, computational savings are thus proportional to $T$.

\subsection{Experiments}

\begin{table}[bt]
\begin{center}
\resizebox{\textwidth}{!}{
\begin{tabular}{lclllllll}
    \toprule
    \multirow{2}{*}{\textbf{Model}} 
    &\textbf{Frames/} 
    &\multicolumn{2}{c}{\textbf{Accuracy (\%)}}
    &\textbf{Par.}
    &\textbf{Max mem.}
    &\textbf{FLOPs/pred}
    &\multicolumn{2}{c}{\textbf{Throughput (preds/s)}}
        \\ 
            \cline{3-4} 
            \cline{8-9}
        & \textbf{pred} & X-Sub & X-View & (M) & (MB) & (G) & CPU & GPU
    \\
    \midrule
    ST-GCN                              & 300   & 86.0  & 93.4  & 3.14      & \phantom{0}45.3    & 16.73     & \phantom{0}2.3     & \phantom{0}180.8 \\
    
    ST-GCN$^*$                          & 300   & 86.3 \textcolor{lgreen}{\tiny($+0.3$)}  & 93.8 \textcolor{lgreen}{\tiny($+0.4$)}  & 3.14      & \phantom{0}72.6 \textcolor{lred}{\tiny($160\%$)}   & 36.90 \textcolor{lred}{\tiny\phantom{00}($\uparrow2.2\times$)}      & \phantom{0}1.1 \textcolor{lred}{\tiny\phantom{0}($\downarrow2.1\times$)}     & \phantom{00}90.4 \textcolor{lred}{\tiny\phantom{0}($\downarrow2.0\times$)} \\
    
    \textit{Co}ST-GCN                   & 4     & 85.3 \textcolor{lred}{\tiny($-0.7$)}  & 93.1 \textcolor{lred}{\tiny($-0.3$)}   & 3.14       & \phantom{0}36.0 \textcolor{lgreen}{\tiny\phantom{0}($79\%$)}     & \phantom{0}0.27 \textcolor{lgreen}{\tiny\phantom{0}($\downarrow63.2\times$)}      & 32.3 \textcolor{lgreen}{\tiny($\uparrow14.0\times$)}     & 2375.2 \textcolor{lgreen}{\tiny($\uparrow13.1\times$)} \\
    
    \textit{Co}ST-GCN$^*$               & 1     & 86.3 \textcolor{lgreen}{\tiny($+0.3$)} & 93.8 \textcolor{lgreen}{\tiny($+0.4$)}   & 3.14     & \phantom{0}36.1 \textcolor{lgreen}{\tiny\phantom{0}($80\%$)}    & \phantom{0}0.16 \textcolor{lgreen}{\tiny ($\downarrow107.7\times$)}     & 46.1 \textcolor{lgreen}{\tiny($\uparrow20.0\times$)}     & 4202.2 \textcolor{lgreen}{\tiny($\uparrow23.2\times$)} \\
    
    \midrule
    AGCN                                & 300   & 86.4  & 94.3  & 3.47   & \phantom{0}48.4  & 18.69     & \phantom{0}2.1   & \phantom{0}146.2 \\
    
    AGCN$^*$                            & 300   & 84.1 \textcolor{lred}{\tiny($-2.3$)}  & 92.6 \textcolor{lred}{\tiny($-1.7$)}  & 3.47        & \phantom{0}76.4 \textcolor{lred}{\tiny($158\%$)}   & 40.87
    \textcolor{lred}{\tiny\phantom{00}($\uparrow2.2\times$)}     & \phantom{0}1.0 \textcolor{lred}{\tiny\phantom{0}($\downarrow2.1\times$)}     & \phantom{00}71.2 \textcolor{lred}{\tiny\phantom{0}($\downarrow2.0\times$)} \\
    
    \textit{Co}AGCN                     & 4     & 86.0 \textcolor{lred}{\tiny($-0.4$)} & 93.9 \textcolor{lred}{\tiny($-0.4$)} & 3.47     & \phantom{0}37.3 \textcolor{lgreen}{\tiny\phantom{0}($77\%$)}       & \phantom{0}0.30 \textcolor{lgreen}{\tiny\phantom{0}($\downarrow63.2\times$)}     & 25.0 \textcolor{lgreen}{\tiny($\uparrow11.9\times$)}     & 2248.4 \textcolor{lgreen}{\tiny($\uparrow15.4\times$)} \\
    
    \textit{Co}AGCN$^*$                 & 1     & 84.1 \textcolor{lred}{\tiny($-2.3$)}  & 92.6 \textcolor{lred}{\tiny($-1.7$)}  & 3.47   & \phantom{0}37.4 \textcolor{lgreen}{\tiny\phantom{0}($77\%$)}       & \phantom{0}0.17 \textcolor{lgreen}{\tiny($\downarrow108.8\times$)}      & 30.4 \textcolor{lgreen}{\tiny($\uparrow14.5\times$)}     & 3817.2 \textcolor{lgreen}{\tiny($\uparrow26.1\times$)} \\
    
    \midrule
    S-TR                                & 300   & 86.8     & 93.8     & 3.09    & \phantom{0}74.2      & 16.14         & \phantom{0}1.7     & \phantom{0}156.3 \\
    
    S-TR$^*$                            & 300   & 86.3 \textcolor{lred}{\tiny($-0.5$)}      & 92.4 \textcolor{lred}{\tiny($-1.4$)}     & 3.09    & 111.5 \textcolor{lred}{\tiny($150\%$)}      & 35.65 \textcolor{lred}{\tiny\phantom{00}($\uparrow2.2\times$)}        & \phantom{0}0.8 \textcolor{lred}{\tiny\phantom{0}($\downarrow2.1\times$)}     & \phantom{00}85.1 \textcolor{lred}{\tiny\phantom{0}($\downarrow1.8\times$)} \\
    
    \textit{Co}S-TR                     & 4     & 86.5 \textcolor{lred}{\tiny($-0.3$)}     & 93.3 \textcolor{lred}{\tiny($-0.5$)}      & 3.09    & \phantom{0}35.9 \textcolor{lgreen}{\tiny\phantom{0}($48\%$)}      & \phantom{0}0.22 \textcolor{lgreen}{\tiny\phantom{0}($\downarrow63.2\times$)}         & 30.3 \textcolor{lgreen}{\tiny($\uparrow17.8\times$)}     & 2189.5 \textcolor{lgreen}{\tiny($\uparrow14.0\times$)} \\
    
    \textit{Co}S-TR$^*$                 & 1     & 86.3 \textcolor{lred}{\tiny($-0.3$)}     & 92.4 \textcolor{lred}{\tiny($-1.4$)}     & 3.09   & \phantom{0}36.1 \textcolor{lgreen}{\tiny\phantom{0}($49\%$)}       & \phantom{0}0.15 \textcolor{lgreen}{\tiny($\downarrow107.6\times$)}         & 43.8 \textcolor{lgreen}{\tiny($\uparrow25.8\times$)}     & 3775.3 \textcolor{lgreen}{\tiny($\uparrow24.2\times$)} \\
    
    \bottomrule
\end{tabular}
}
\end{center}
\caption{
    {NTU RGB+D 60 transfer accuracy and performance benchmarks}. Noted is the top-1 validation accuracy using joints as the only modality. 
    Max mem. is the maximum allocated memory on GPU during inference noted in megabytes.
    Max. mem, FLOPs, and throughput on CPU account for one new prediction with batch size one, while throughput on GPU uses the largest fitting power of two as batch size.
    Parentheses indicate the \textcolor{lgreen}{improvement} / \textcolor{lred}{deterioration} relative to the baseline models ST-GCN, AGCN, and S-TR.
    The table first appeared in paper \ref{pap:costgcn}.
}
\label{tab:skel-benchmark-speed}
\end{table}

In practice, the ST-GCN-based methods use long skeletal sequences as inputs.
To reduce the computational complexity, multiple layers in the network use a temporal stride of two. 
In the context of CINs, however, the temporal stride has the undesired effect of reducing the prediction rate (see \cref{sec:co-arch-considerations}). In addition to direct CIN reformulations of ST-GCN~\cite{yan2018stgcn}, AGCN~\cite{shi2019agcn}, and S-TR~\cite{plizzari2021str}, \ie the proposed \textit{Co}ST-GCN, \textit{Co}AGCN, and \textit{Co}S-TR models, variants with stride one and without temporal padding are also created. 
While the weights of regular networks and their CIN counterpart can be shared, the stride modification requires fine-tuning of the model. 
As described in \cref{sec:co-arch-considerations}, this can be done either with full-sequence training or with step-wise training. Both options were explored in paper \ref{pap:costgcn} and found to give similar results.

We tested each baseline model alongside continual padding and stride-free variants on the NTU RGB+D 60 dataset~\cite{shahroudy2016ntu}.
In addition to accuracy and parameter count, we benchmarked the maximum allocated GPU memory, FLOPs per prediction, as well as throughput on a MacBook Pro 16" 2019 2.6 GHz i7 processor and an NVIDIA RTX 2080Ti GPU. The results of this benchmark are presented in \cref{tab:skel-benchmark-speed}.
In each case, the conversion to CIN comes with the desired reduction in FLOPs and a throughput increase. While direct conversion of the baseline models results in a $63\times$ FLOPs reduction per prediction, the removal of zero-padding and reduction of temporal stride (marked with an asterisk `$*$' in \cref{tab:skel-benchmark-speed}) reduces the per-prediction complexity further, to approx. $108\times$ less FLOPs. 
While the on-hardware throughput increases do not reach multiple magnitudes of improvement due to the overhead of repeat memory access, they are increased by an average of $14.4\times$ for direct conversion and $22.3\times$ for stride-modified models (marked with `$*$'). 
Moreover, maximum allocated memory is reduced by approx. 22\% for \textit{Co}ST-GCN and \textit{Co}AGCN and by 51\% for \textit{Co}S-TR due to the smaller intermediary feature-maps, which exclude the temporal dimension. The accuracy is similar across all models, except for (\textit{Co})AGCN$^*$, which has a slightly lower accuracy due to a modification to the self-attention, which restricts it to satisfy causality by only considering the current skeleton rather than all skeletons in the sequence.

\begin{figure}[tb]
    \centering
    \includegraphics[width=0.8\linewidth]{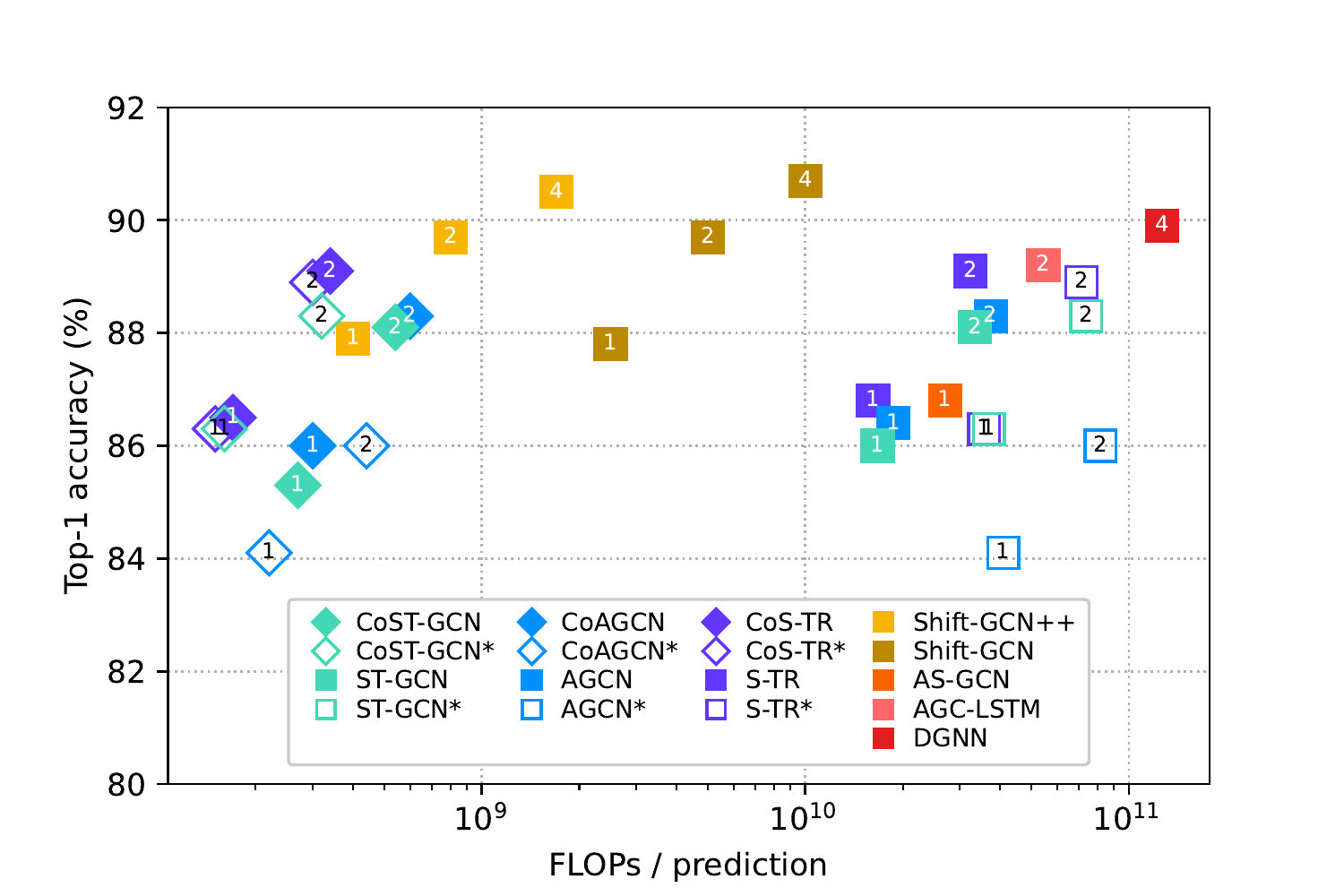}
    \caption{
        {Accuracy/complexity trade-off} on NTU RGB+D 60 X-Sub for $\sdiamond$ \textit{Continual} and $\ssquare$ prior methods during online inference.
        Numbers denote the number of data streams employed for each evaluation.
        $^*$Architecture modification with stride one and no padding.
        The figure first appeared in paper \ref{pap:costgcn}.
    }
    \label{fig:ntu60-test-acc-vs-flops}
\end{figure}

Paper \ref{pap:costgcn} also features extensive comparisons with prior works on the NTU RGB+D 60~\cite{shahroudy2016ntu}, NTU RGB+D 120~\cite{liu2019ntu120}, and Kinetics Skeleton 400~\cite{kay2017kinetics} benchmarks, where the stride-modified continual models provided the best accuracy/complexity trade-offs for online processing.
A visual depiction of the Accuracy/FLOPs trade-off for NTU RGB+D 60 is provided in \cref{fig:ntu60-test-acc-vs-flops}. 
For additional details, please see paper \ref{pap:costgcn}.

\section{Continual Transformers}\label{sec:cotrans}

\subsection{Related works}
Transformers~\cite{vaswani2017attention} are among the most influential DNN architecture innovations of the past decade. They are constructed around the core building-block of scaled dot-product attention (SDA), which performs global attention among a set of input tokens. 
The SDA uses three sets of $n$ tokens denoted query, key, and value, $\mQ, \mK, \mV \in \mathbb{R}^{n \times d}$. 
Given a single input set, $\mX \in \mathbb{R}^{n \times d'}$, query, key and value can be retrieved as projections with embedding weights $\mW_{\{q,j,v\}} \in \mathbb{R}^{d' \times d}$, \ie, $\mQ = \mX \mW_q$, $\mK = \mX \mW_k$, and $\mV = \mX \mW_v$. 
The token-wise similarity between key and query matrices is used to aggregate information from a value matrix as follows:
\noindent
\begin{align}
    \text{Att}(\mQ, \mK, \mV) &= \mD^{-1} \mA \mV
    \\
    \mA &= \text{exp}\left(\mQ \mK^{\top} / \sqrt{d} \right)
    \\
    \mD &= \text{diag}\left( \mA \mathbbm{1}_n ^\top \right),
        \label{eq:scaled-dot-product-attention}
\end{align}
where $\mathbbm{1}_n$ is a row-vector of $n$ ones.
Here, the outer product between $\mQ$ and $\mK$ gives rise to a quadratic computational complexity of $\mathcal{O}(n^2d)$.

Much work has been dedicated to improving the transformer efficiency. 
The Image Transformer~\cite{parmar2018image} and Vision Transformer~\cite{dosovitskiy2021vit} use fixed groups of $n_b$ tokens which attend to one another locally to reduce the complexity to $\mathcal{O}(n_b^2 d)$.
As opposed to a fixed grouping scheme, Reformer~\citep{kitaev2020reformer} learns groupings via locality-sensitive hashing to reduce the complexity to $\mathcal{O}(n d \log n)$.
Linformer~\citep{wang2020linformer}, Nyströmformer~\citep{xiong2021nystromformer} and Performer~\citep{choromanski2021rethinking} approximate the self-attention matrix with a $\mathcal{O}(nd)$ complexity.
However, none of the above-mentioned methods optimize for sequential attention, where one new token is considered each time-step during stream processing. During online processing, they are restricted to operate on tokens in a sliding-window, all of which are passed each time a new prediction is desired.
Paper \ref{pap:cotrans} presents the first formulation of redundancy-free sequential self-attention, which retains weight-compatibility with the SDA in \cref{eq:scaled-dot-product-attention}, but is optimized for online processing. Specifically, this work provides a bottom-up reformulation of Transformers as CINs.

\subsection{Continual Scaled Dot-product Attention}
The global interaction between tokens passed to the SDA operation is simultaneously its greatest benefit and the main hindrance to efficient computation.
In the continual inference setting, it poses an issue as well: How can we efficiently separate computations across time-steps when the output of the first frame we observed depends on the newest frame and vice versa? 
As stream processing needs to handle a potentially endless sequence, we have no choice but to restrict the considered sequence to at most $n$ tokens.
Furthermore, there are two behavioral options to consider when we process a time-step: Should the \textit{Continual} SDA output (a) $n$ updated tokens retroactively in accordance with a regular SDA used with sliding-window processing, or (b) a single updated token corresponding to a desired token position, \eg, the latest frame or a class token?
Paper \ref{pap:cotrans} proposes efficient formulations for both, namely the \textit{Continual Retroactive SDA} and the \textit{Continual Single-output SDA}, which have computational and memory complexities of $\mathcal{O}(nd)$ and produce identical results as \cref{eq:scaled-dot-product-attention}.

\subsubsection{Continual Retroactive SDA}

\begin{figure}[bt]
    \centering
    \includegraphics[width=0.7\linewidth]{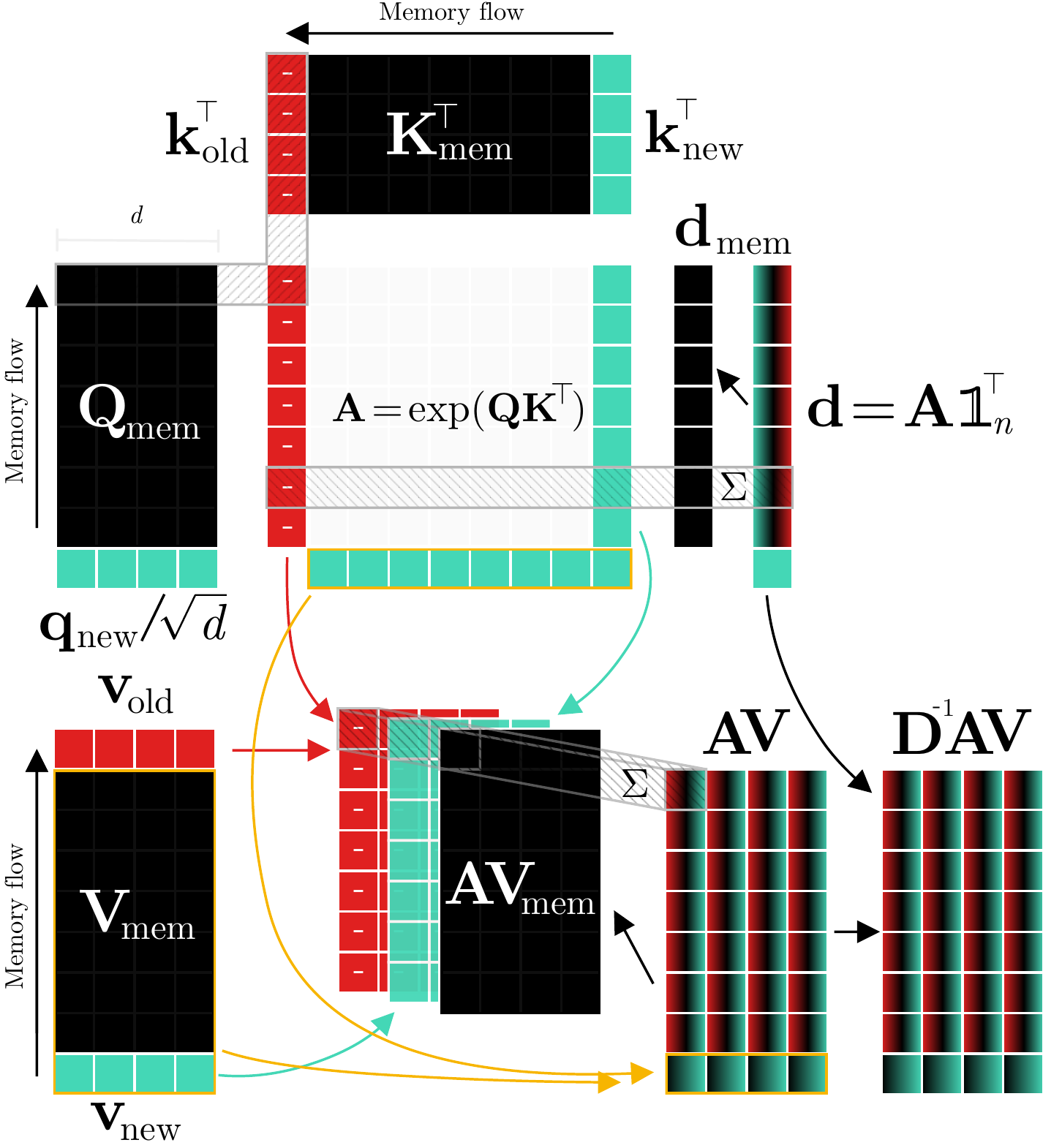}
    \caption{
        {Continual Retroactive Dot-Product Attention}. 
        The query ($\mQ$), key ($\mK$), and value ($\mV$) matrices are aggregated over time by caching the step vectors \textcolor{teal}{$\vqn$}, \textcolor{teal}{$\vkn$}, and \textcolor{teal}{$\vvn$} in each their FIFO queue (denoted by $\square_\text{mem}$). During each step, only the entries of $\mA$ associated with \textcolor{teal}{$\vqn$}, \textcolor{teal}{$\vkn$} and the oldest $\mK$ step, \textcolor{ruby}{$\vko$} are computed. 
        The diagonal entries of the row-normalization matrix $\mD$ as well as the $\mA\mV$ can be updated retroactively by subtracting features corresponding to \textcolor{ruby}{$\vko$} and adding features related to \textcolor{teal}{$\vkn$} to the cached outputs of the previous step, $\mD_\text{mem}$ and $\mA\mV_\text{mem}$, respectively.
        The figure first appeared in paper \ref{pap:cotrans}.
    }
    \label{fig:co-re-dot-prod-attention}
\end{figure}

Instead of caching $\mA$, which would inflict a memory complexity of $\mathcal{O}(n^2)$, the input sequences $\mQ$, $\mK$, and $\mV$ are treated as first-in-first-out queues.
Then, the idea is to add contributions from new tokens, $\vqn, \vkn, \vvn \in \mathbb{R}^{1 \times d}$ while subtracting contributions from old tokens, $\vko, \vvo \in \mathbb{R}^{1 \times d}$, which have slid out of scope.
First, we compute new entries of $\mA$, namely $\text{exp}\left( \mQ_\text{mem}\vkn^\top\right)$, as well old entries, $\text{exp}\left( \mQ_\text{mem}\vko^\top \right)$.
We can use these alongside cached entries of the softmax row-normalization vector, $\vd_{\text{mem}}$, to compute and updated row-normalization vector
\begin{equation}
    \vd^{(-n+1:-1)} 
        = \vd_\text{mem}^{(-n+2:0)} 
        - \text{exp}\left( \mQ_\text{mem}\vko^\top \right)
        + \text{exp}\left( \mQ_\text{mem}\vkn^\top\right).
    \label{eq:core-sdpa-d-update}
\end{equation}
The newest entry is computed from scratch:
\begin{equation}
    \vd^{(0)} = \text{exp}\left( \frac{\vqn}{\sqrt{d}}\left(\mK_\text{mem} \mathbin\Vert  \vkn \right)^\top \right) \mathbbm{1}_n^\top,
    \label{eq:core-sdpa-d-0}
\end{equation}
where $\mathbin\Vert$ denotes matrix concatenation.
Similarly, the prior $\mA\mV$ is cached as $\mA\mV_{\text{mem}}$ and updated based on the outer products of the newly computed entries of $\mA$ as well as $\vvn$ and $\vvo$:
\begin{equation}
    \mA\mV^{(-n+1:-1)} = 
        \mA\mV_\text{mem}^{(-n+2:0)} 
        - \text{exp}\left(\mQ_\text{mem} \vko^\top \right) \vvo  
        + \text{exp}\left(\mQ_\text{mem} \vkn^\top \right) \vvn.
    \label{eq:core-sdpa-av-update}
\end{equation}
The row corresponding to the newest value is computed from scratch:
\begin{equation}
    \mA\mV^{(0)} 
        = \text{exp}\left(
            \frac{\vqn}{\sqrt{d}} 
            \left(\mK_\text{mem} \mathbin\Vert  \vkn \right)^\top
        \right)
        \left(\mV_\text{mem} \mathbin\Vert  \vvn \right).
    \label{eq:core-sdpa-av-0}
\end{equation}
Finally, the attention output is attained by normalizing $\mA\mV$ with $\vd$:
\begin{equation}
    CoRe\text{Att}(\vqn, \vkn, \vvn) 
    = \vd^{-1} \odot \mA \mV,
    \label{eq:core-sdpa-dav}
\end{equation}
where $\odot$ denotes element-wise multiplication.
The above-described process is graphically illustrated in \cref{fig:co-re-dot-prod-attention}.

\subsubsection{Single-output SDA}
\begin{figure}[tb]
    \centering
    \includegraphics[width=0.7\linewidth]{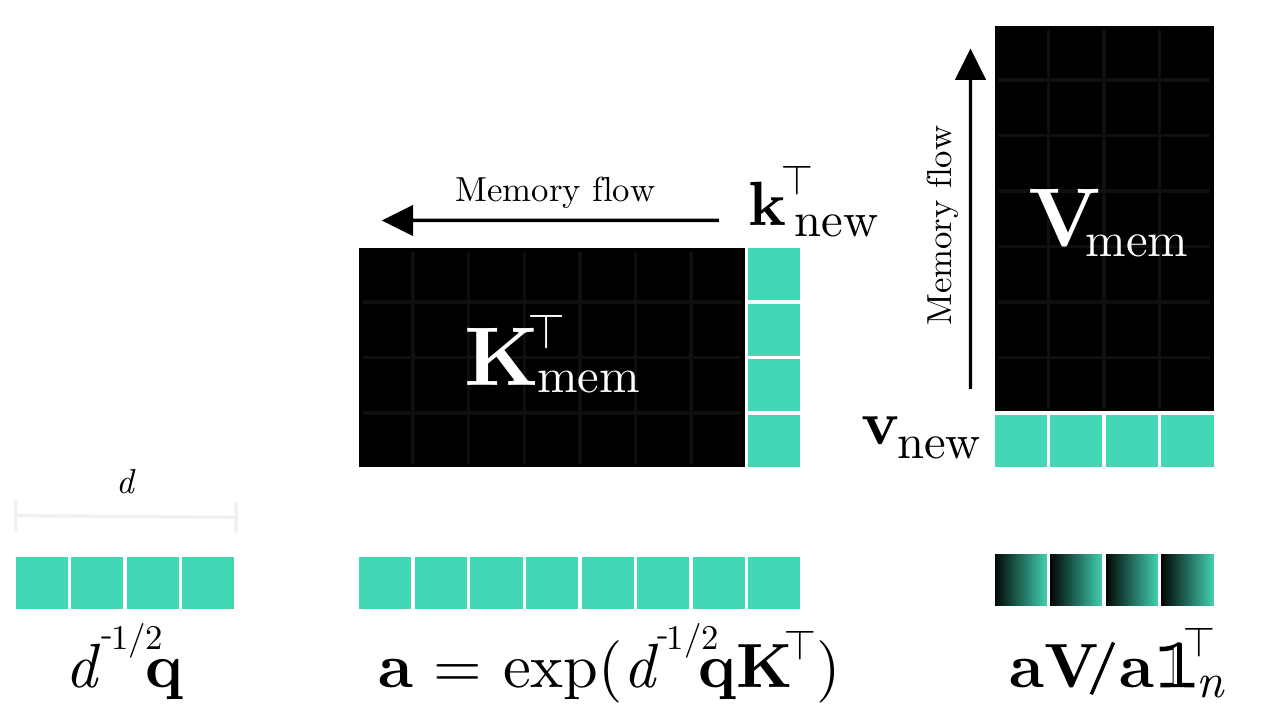}
    \caption{
        {Continual Single-Output Dot-Product Attention}. 
        The key ($\mK$) and value ($\mV$) matrices are aggregated over time by caching the step vectors $\vkn$ and $\vvn$ in a FIFO queue. During each step, only the attention output associated with $\vq$ is computed.
        The figure first appeared in paper \ref{pap:cotrans}.
    }
    \label{fig:cosi-dot-prod-attention}
\end{figure}

To construct an efficient single-output SDA for continual inference, $\mK$ and $\mV$ are treated as a first-in-first-out queues again, where the oldest token is discarded when a new token is received. 
Since only a single output is desired, the attention matrix $\mA$ is limited to a single row $\va$ corresponding to the desired step query $\vq$:
\begin{equation}
    \va = \text{exp}\left(
        \frac{\vq}{\sqrt{d}} 
        \left(\mK_\text{mem} \mathbin\Vert  \vkn \right)^{\top} 
    \right).
\end{equation}
The attention computation is straight-forward:
\begin{equation}
    CoSi\text{Att}(\vq, \vkn, \vvn) 
        = \va \left(\mV_\text{mem} \mathbin\Vert  \vvn \right) 
        / \va\mathbbm{1}_n^\top.
        \label{eq:leading-scaled-dot-product-attention-att}
\end{equation}
\cref{fig:cosi-dot-prod-attention} provides a visualization of the process.

\subsection{Continual Transformer Encoder}
The full \textit{Continual Transformer Encoder} architecture is structured in a near-identical manner to the non-continual version, with only minor variations required in the implementation of the \textit{Continual Multi-head Attention} and \textit{Continual Transformer Encoder} block.

\subsubsection{Continual Multi-head Attention}
The \textit{Continual Multi-head Attention} is computed by concatenating the outputs of $h$ continual attention heads for the current query, key, and value vectors projected by $\mW_Q^{i}, \mW_K^{i} \in \mathbbm{R}^{d \times d_K/h}$, $\mW_V^{i} \in \mathbbm{R}^{d \times d_V/h}$, and performing a per-token linear projection using an output weight, $\mW_O \in \mathbb{R}^{d_V \times d_O}$:
\begin{equation}
    Co\text{MHA}(\vq, \vk, \vv) 
    = \left(\Concat_{i=0}^{h-1} Co\text{Att}(
        \vq \mW_Q^{i}, \vk \mW_K^{i}, \vv \mW_V^{i}
    )\right) \mW_O.
\end{equation}
Here, \textit{Co}Att$(\cdot)$ can either be the retroactive or single-output SDA, \textit{CoRe}Att$(\cdot)$ or \textit{CoSi}Att$(\cdot)$, described in \cref{eq:core-sdpa-dav} and \cref{eq:leading-scaled-dot-product-attention-att}, respectively.

\subsubsection{Continual Transformer Encoder block}
The \textit{Continual Transformer Encoder block} first adds the results of the continual multi-head attention for the current input token $\vx$ residually to the input. 
\begin{equation}
    \vy = \text{LayerNorm} \left(\text{Sel}(\vx) + Co\text{MHA}(\vx, \vx, \vx) \right),
\end{equation}
Here, we use the notation of a selection function, $\text{Sel}(\cdot)$, to note that only a single (\eg, last) token should be selected when the Continual Single-output approach is used. 
The final output is then given by token-wise projection with a two-layer feed-forward network:
\begin{gather}
    \vz = \text{LayerNorm} \left(\vy + \text{FF}(\vy) \right) \\
    \text{FF}(\vx) = \text{ReLU}(\vx\mW_1 w_1)\mW_2 + w_2.
\end{gather}

\subsubsection{Recycling Positional Encoding}
Transformer Encoders do not have inherent positional bias. Hence, it is common to augment input tokens $\vx_i$ with positional encodings $\vp_i$.
However, for regular transformers, the index $i$ denotes a position in a sequence rather than a position in time. 
This poses an issue, since the last token at time $t$ will be next-to-last at time $t+1$, and thus in need of a different positional encoding.
Stated another way, each new input token step also changes the required positional encoding of prior steps. In effect, caching of prior values is not an option under the common absolute positional encoding.
As a solution, the position should be fixed in time rather than in sequence. After $T$ steps have passed, the positional encodings can then be recycled:
\begin{equation}
    \Tilde{\vx}_t = \vx_t + \vp_{\tau_t}, \label{eq:add-enc}
    \qquad
    \tau_t = (\tau_{t-1} + 1) \text{ mod } T.
\end{equation}
In the ablation studies conducted in paper~\ref{pap:cotrans}, the use of this \textit{Recycling Positional Encoding} was found to be crucial to attain good performance during continual inference.

\subsubsection{Architectural considerations}
\begin{figure}[tb]
    \centering
    \includegraphics[width=\linewidth]{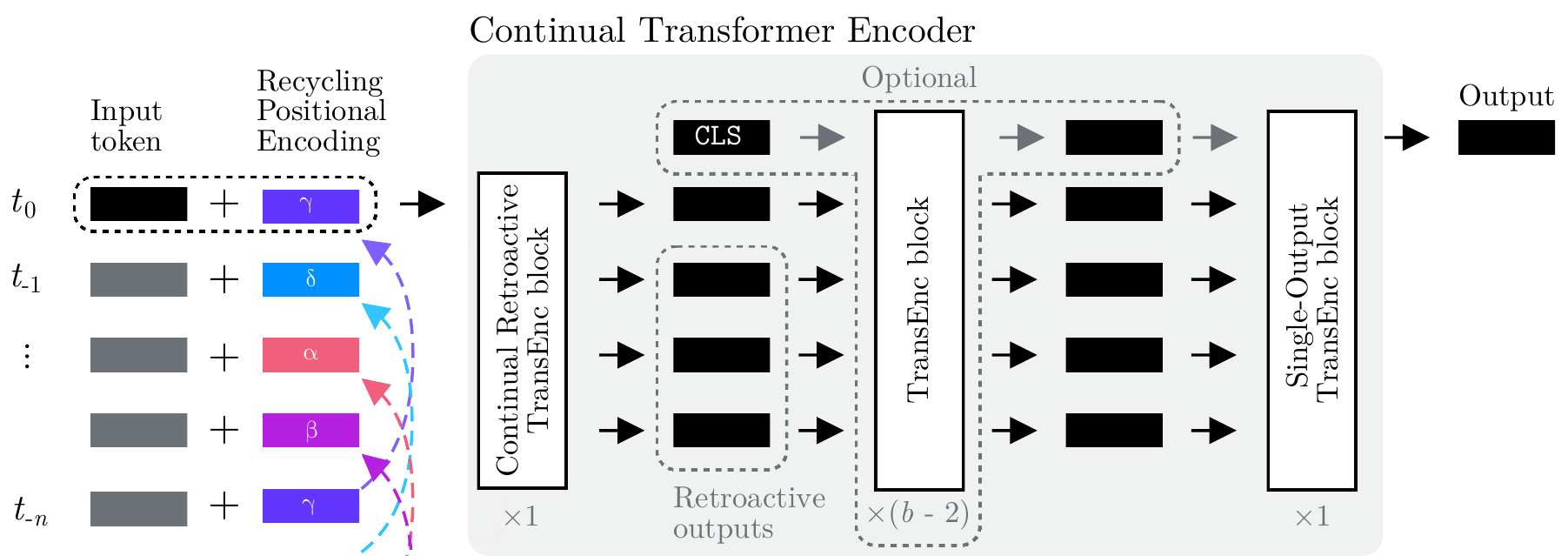}
    \caption{
        Multi-block Continual Transformer Encoder with Recycling Positional Encoding. For $b>2$ blocks, regular Transformer Encoder blocks can be added between an initial Continual Retroactive block and a final Single-Output block. A class-token may be used after the initial block.
        The figure first appeared in paper \ref{pap:cotrans}.
    }
    \label{fig:cotrans-three-layer-architecture}
\end{figure}

The full continual Transformer Encoder architecture with Recycling Positional Encoding and Continual Transformer Encoder blocks is illustrated in \cref{fig:cotrans-three-layer-architecture}.
If a \textit{single} transformer encoder block is desired, the Continual Single-output Transformer Encoder block is ideal for classification tasks and the Continual Retroactive Transformer Encoder block for sequence to sequence tasks. 
For \textit{two} blocks, a Continual Retroactive Transformer Encoder block should be employed first and a regular Single-output Transformer Encoder last.
For \textit{three blocks or more}, the same setup applies for the first and last block, but regular transformer encoders can be stacked in between.
Due to the one-to-many nature of Continual Retroactive Transformer Encoders, their efficiency gains quickly diminish as more blocks are added. Consequently, we recommend the use of one- or two block architectures.
While this may seem restrictive, concurrent work has shown that shallow Transformer architectures may perform as well as deeper networks if the parameter count is matched by scaling out the number of attention heads~\cite{brown2020wide}.

\subsection{Experiments}
The Continual Transformer Encoders are validated in the context of \textit{Online Action Detection}~\cite{geest2016online}, where the task is to predict per-frame human action categories without peeking into the future.
Here, recent works primarily utilize pre-extracted features from two CNN streams trained on RGB images and optical flow fields, respectively, and aggregate the temporal information via RNNs~\cite{gao2017red, xu2019temporal, eun2020learningtd} or Transformers~\cite{wang2021oadtr, xu2021long}.
Action anticipation is a related task, which may be learned in parallel, as it has been found to improve OAD as well.
In the experiments with Continual Transformer Encoders, a simplified OadTR~\cite{wang2021oadtr} architecture was employed, which did not include the action anticipation decoder and class token, and which used fewer Transformer Encoder blocks (denoted by b1 and b2 for one and two blocks). 
In the ablation studies conducted in paper \ref{pap:cotrans}, these modifications were found to have limited effect on predictive performance despite reducing computational complexity significantly.

\begin{table}[bt]
\begin{center}
\begin{tabular}{llcccc}
    \toprule
    \multirow{2}{*}{\textbf{Model}} & \multirow{2}{*}{\textbf{Feat.}}  & \textbf{THUMOS14} & \textbf{TVSeries}     & \textbf{FLOPs} \\
                                    &                                  & \textbf{mAP (\%)}          & \textbf{mcAP (\%)}              &\textbf{(M)} \\
    \midrule
    RED~\citep{gao2017red}           &\multirow{12}{*}{A.Net}& 45.3\phantom{\tiny{$\pm$0.0}}             & 79.2\phantom{\tiny{$\pm$0.0}}                & - \\
    TRN~\citep{xu2019temporal}       &                       & 47.2\phantom{\tiny{$\pm$0.0}}             & 83.7\phantom{\tiny{$\pm$0.0}}                & 1387.5 \\ 
    FATS~\citep{kim2021temporally}   &                       & 51.6\phantom{\tiny{$\pm$0.0}}             & 81.7\phantom{\tiny{$\pm$0.0}}                & - \\
    IDN~\citep{eun2020learningtd}    &                       & 50.0\phantom{\tiny{$\pm$0.0}}             & 84.7\phantom{\tiny{$\pm$0.0}}                & - \\
    TFN~\citep{eun2021temporal}      &                       & 55.7\phantom{\tiny{$\pm$0.0}}             & 85.0\phantom{\tiny{$\pm$0.0}}                & - \\
    LSTR~\citep{xu2021long}          &                       & \best{65.3}\phantom{\tiny{$\pm$0.0}}             & 88.1\phantom{\tiny{$\pm$0.0}}                & - \\
    OadTR~\citep{wang2021oadtr}      &                       & \nextbest{58.3}\phantom{\tiny{$\pm$0.0}}             & 85.4\phantom{\tiny{$\pm$0.0}}                & 2445.6 \\
    OadTR$^\dagger$                 &                       & 57.0\tiny{$\pm$0.5}             & \best{88.6}\tiny{$\pm$0.1}               & 2445.6 \\
    OadTR-b2$^\dagger$              &                       & 56.6\tiny{$\pm$0.3}             & \nextbest{88.3\tiny{$\pm$0.2} }                    & 1008.1 \\
    OadTR-b1$^\dagger$              &                       & 56.3\tiny{$\pm$0.2}             & 88.1\tiny{$\pm$0.1}                     & 605.5 \\
    \textit{Co}OadTR-b2 (ours)      &                       & 56.8\tiny{$\pm$0.4}             & 87.7\tiny{$\pm$0.6}                     & \nextbest{410.9} \\
    \textit{Co}OadTR-b1 (ours)      &                       & 56.1\tiny{$\pm$0.7}             & 87.6\tiny{$\pm$0.7}                    & \best{9.6} \\

    \midrule

    TRN~\citep{xu2019temporal}       &                       & 62.1\phantom{\tiny{$\pm$0.0}}             & 86.2\phantom{\tiny{$\pm$0.0}}                & 1462.0 \\
    FATS~\citep{kim2021temporally}   & \multirow{10}{*}{Kin.}& 59.0\phantom{\tiny{$\pm$0.0}}             & 84.6\phantom{\tiny{$\pm$0.0}}                & - \\
    IDN~\citep{eun2020learningtd}    &                       & 60.3\phantom{\tiny{$\pm$0.0}}             & 86.1\phantom{\tiny{$\pm$0.0}}                & - \\
    PKD~\citep{zhao2020privilegedkd} &                       & 64.5\phantom{\tiny{$\pm$0.0}}             & 86.4\phantom{\tiny{$\pm$0.0}}                & - \\
    {LSTR}~\citep{xu2021long}        &                       & \best{69.5}\phantom{\tiny{$\pm$0.0}}             & \best{89.1}\phantom{\tiny{$\pm$0.0}}                & - \\
    OadTR~\citep{wang2021oadtr}      &                       & \nextbest{65.2}\phantom{\tiny{$\pm$0.0}}             & 87.2\phantom{\tiny{$\pm$0.0}}                & 2513.5 \\
    {OadTR$^\dagger$}               &                       & 64.2\tiny{$\pm$0.3}             & \nextbest{88.6\tiny{$\pm$0.1}}              & {2513.5} \\ 
    OadTR-b2$^\dagger$              &                       & 64.5\tiny{$\pm$0.5}             & 88.3\tiny{$\pm$0.2}               & 1075.7 \\ 
    OadTR-b1$^\dagger$              &                       & 63.9\tiny{$\pm$0.5}             & 88.1\tiny{$\pm$0.1}               & 673.0 \\ 
    {\textit{Co}OadTR-b2 (ours)}    &                       & 64.4\tiny{$\pm$0.1}             & 87.6\tiny{$\pm$0.7}               & \nextbest{411.9} \\ 
    {\textit{Co}OadTR-b1 (ours)}    &                       & 64.2\tiny{$\pm$0.4}             & 87.7\tiny{$\pm$0.4}               & \best{10.6 }\\ %
    \bottomrule
\end{tabular}
\caption{
    {Online Action Detection} results.
    FLOPs per prediction are noted for inference on THUMOS14. 
    The \best{best} and \nextbest{next-best} metrics are highlighted.
    $^\dagger$ indicates use official source code or modifications there-off.
    The table first appeared in paper \ref{pap:cotrans}.
}\label{tab:oad-prior-works}
\end{center}
\end{table}

A comparison with prior works on the THUMOS14~\cite{idrees2017thumos} and TVSeries~\cite{geest2016online} datasets is presented  in \cref{tab:oad-prior-works}. 
Please see paper~\ref{pap:cotrans} for descriptions of experimental methodology.
Two sets of results are shown in \cref{tab:oad-prior-works} using features from a Temporal Segment Network~\cite{wang2018tsn} pre-trained on either ActivityNet~\cite{heilbron2015activitynet} (top) or Kinetics-400~\cite{kay2017kinetics} (bottom).
Compared to OadTR, the proposed \textit{Co}OadTR-b1 and \textit{Co}OadTR-b2 reduce FLOPs by 255$\times$ and 6.1$\times$, respectively. While minor concessions to the predictive performance are observed (at most 1\% point), these are expected given the lower parameter count of the b1 and b2 models. Comparing continual and non-continual simplified \mbox{OadTR-b\{1,2\}} implementations, \ie, near-identical architectures which either use sliding-window processing or continual inference processing, the \textit{Co}OadTR-b1 reduces FLOPs by approx. $63\times$ and the \textit{Co}OadTR-b2 by $2.5\times$. The discrepancy in efficiency gains is explained by the additional computations needed in the two-block architecture to aggregate retroactively computed tokens.

Additional results for audio classification on GTZAN~\cite{tzanetakis2002musical} as well as ablation studies on positional encoding, number of blocks, and removal of decoder and class token are available in paper \ref{pap:cotrans}.

\section{A Python library to support implementation}\label{sec:colib}
While the architectural reformulations presented in this chapter may be simple enough conceptually, a considerable engineering effort is required to handle network delays and inputs shapes properly to retain weight-compatibility between offline and online inference versions.
To aid this effort, paper~\ref{pap:colib} presents a Python library, \textit{Continual Inference} (with shorthand \texttt{co}), which augments the standard PyTorch \texttt{nn} modules with the ability to perform efficient continual inference.

The library is designed to adhere closely to the CIN definition in \cref{sec:cin-def} and to support the necessary architectural considerations outlined in \cref{sec:co-arch-considerations}.
Moreover, the library is implemented with strict PyTorch compatibility as outlined in implementation principles \ref{prin:identical} and \ref{prin:forward-modes}.

\begin{figure}[hb]
    \centering
    \includegraphics[width=\linewidth]{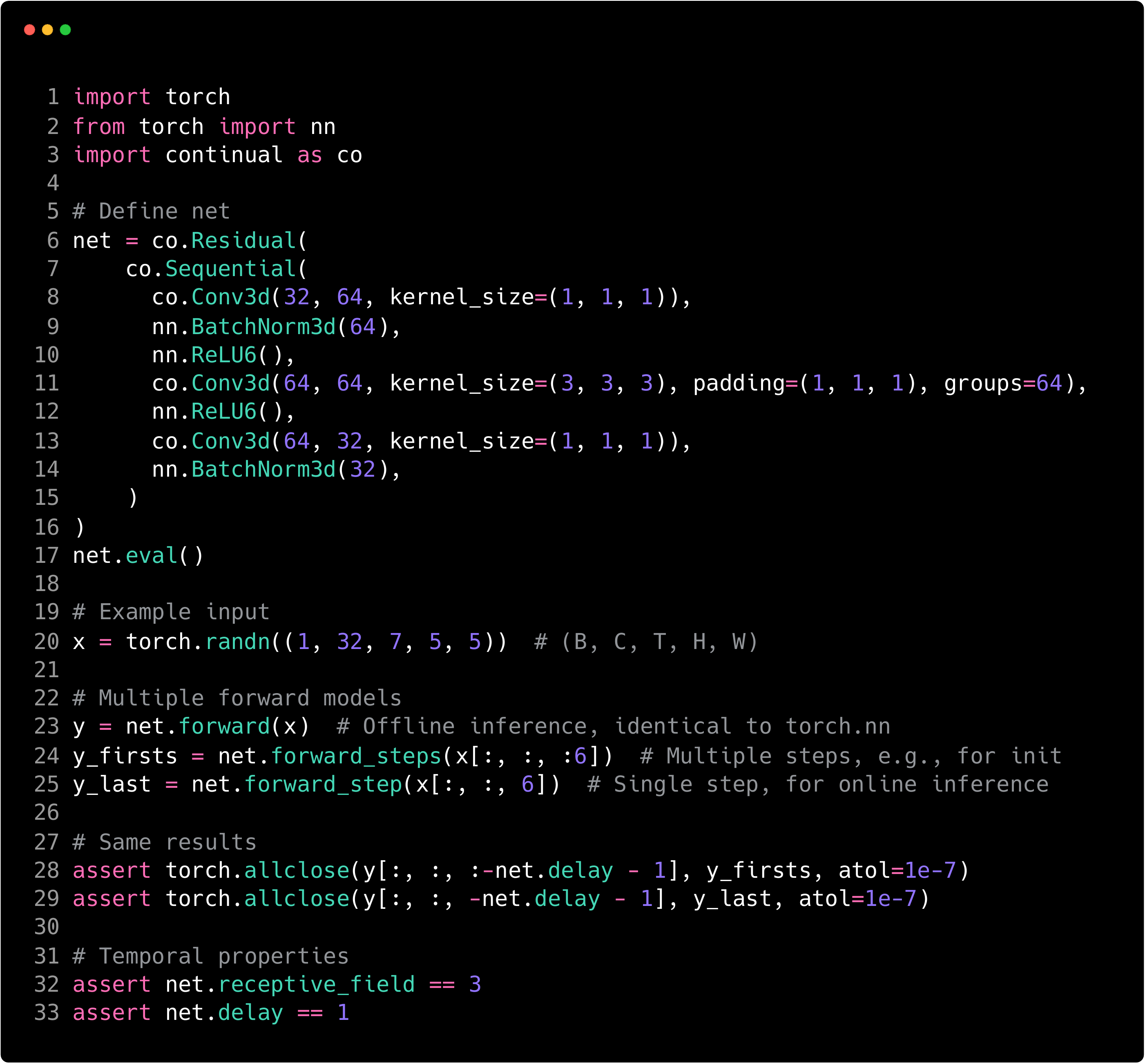}
    \caption{
        Depth-wise separable 3D convolution block, \texttt{net}, implemented with the \textit{Continual Inference} library. Note the mix of \texttt{nn} and \texttt{co} modules (lines 8-14). The library modules feature multiple forward modes for offline and online inference (lines 23-25), which produce the same results (lines 28-29). Moreover, modules are enhanced with temporal properties (lines 32-33).
    }
    \label{fig:colib-dsconv}
\end{figure}

\begin{principle}[Compatibility with PyTorch]
\texttt{co} modules with identical names to \texttt{nn} modules also have:
\begin{enumerate}
    \item identical \texttt{forward},
    \item identical model weights,
    \item identical or extended constructors,
    \item identical or extended supporting functions.
\end{enumerate}
\label{prin:identical}
\end{principle}
\begin{principle}[Call modes]
\texttt{co} modules provide three forward operations:
\begin{enumerate}
    \item \texttt{forward}: takes a (spatio-) temporal input and operates identically to the \texttt{forward} of an \texttt{nn} module,
    \item \texttt{forward\_step}: takes a single time-step as input without a time-dimension and produces an output corresponding to \texttt{forward}, had its input been shifted by one time-step, given identical prior inputs. 
    \item \texttt{forward\_steps}: takes multiple time-steps as input and produces outputs identical to applying \texttt{forward\_step} the number of times corresponding to the temporal size of the input.
\end{enumerate}
\label{prin:forward-modes}
\end{principle}
These principles are best understood with an example, as presented in \cref{fig:colib-dsconv}.
The \textit{Continual Inference} library features a comprehensive re-implementation of basic \texttt{nn} network modules as well as an enhanced set of composition modules such as \texttt{co.Residual}, which automatically matches the delay of the wrapped module. 
For more details on network modules and advanced uses, please see paper~\ref{pap:colib} and the library code at \url{https://github.com/lukashedegaard/continual-inference}.%

\section{Conclusion}
This chapter provided an introduction to Continual Inference Networks (\cref{sec:cin-def}), a class of neural networks tailored for redundancy-free online processing of never-ending (continual) data streams.
They tackle the issue of computational redundancy observed during sliding-window processing of spatio-temporal networks (\eg, 3D CNNs, Spatio-temporal GCNs, and Transformers).
Through the use of temporal re-distributions of computational operations and minor architectural modification, many prior DNNs, which were previously required to use sliding-window processing for online execution, can be reformulated to operate in a sequential manner to efficiently process online data streams.

After outlining the principles of CINs, an overview was given of \textit{Continual 3D CNNs} (\cref{sec:co3d}, paper~\ref{pap:co3d}), a general reformulation of the 3D CNN. Through conversions of multiple prior state-of-the-art 3D CNNs for human activity recognition, Continual 3D CNNs achieved up to $15\times$ FLOPs reduction per prediction and $9\times$ and $7\times$ throughput increase on CPU and GPU for networks with a clip-size of sixteen without sacrificing accuracy or retraining the networks.

\textit{Continual Spatio-temporal GCNs} (\cref{sec:costgcn}, paper~\ref{pap:costgcn}) are the augmentation of the ST-GCN network family, which brings efficient online processing to spatio-temporal graphs. Here, architectures which are weight-compatible with prior ST-GCN-based methods, achieve $63\times$ FLOPs reductions and up to $18\times$ and $15\times$ throughput increase on CPU and GPU, respectively, with identical accuracy on clips of 300 frames. \textit{Co}ST-GCN-based networks with minor architectural modifications can be fine-tuned with only few training epochs to achieve similar accuracy, FLOPs reductions of $108\times$, and throughput increases up to $26\times$ and $24\times$ on CPU and GPU.

\textit{Continual Transformer Encoders} (\cref{sec:cotrans}, paper~\ref{pap:cotrans}) provided a bottom-up reformulation of the Transformer and its scaled dot-product attention, which reduced the per-step computational complexity from $\mathcal{O}(n^2 d)$ to $\mathcal{O}(nd)$, where $n$ is the sequence receptive field and $d$ is the token dimension. This yielded reductions in FLOPs per prediction of $63\times$ and $2.5\times$ for one- and two-block transformers, respectively, on token sequences of length 64. Through a systematic ablation of a prior state-of-the-art method, OadTR, a one-block \textit{Co}OadTR architecture achieved a relative FLOPs reduction of $255\times$ with only $0.7\%$-point lower average predictive performance on online action detection benchmarks.

Finally, the \textit{Continual Inference} library (\cref{sec:colib}, paper~\ref{pap:colib}) provides a comprehensive re-implementation of basic PyTorch DNN components which were augmented with the ability to process temporal data efficiently in a sequential manner. 
In addition, the library features an augmented composition interface, which automatically handles delays and aggregates temporal properties to guarantee compliance with the principles of Continual Inference Networks.

\chapter{Compressing and accelerating derived networks with Structured Pruning Adapters}\label{chap:spa}




\section{Introduction}

Since the inception of deep neural networks (DNNs), there has been an unfailing trend of steadily increasing model size and amount of training data in pursuit of improved predictive performance (see \cref{fig:params-through-time}).
This has reached a point where the largest models require training on the order of one million GPU hours with NVIDIA A100 GPUs~\cite{scao2022bloom}.
However, the computational resources needed to train such DNNs are beyond the capacity of most public institutions and universities. While efforts such as BLOOM~\cite{scao2022bloom} showcase the possibility of collective training, the large private research labs such as OpenAI and DeepMind currently lead the race. Still, even the largest commercial players will eventually be confronted with severely diminishing returns for the money spent on model training. 

Faced with economic and environmental realities, a possible scenario for the 2030s is a gradual slow-down in the trend of progressively larger models and longer intervals between the release of new state-of-the-art DNNs trained on large-scale datasets. 
The increasingly common term of a \textit{foundation model} itself indicates something solid and slow or unchanging. After all, who wants a shaky foundation?
With a smaller diversity in state-of-the-art base-models, their adaptation to use-case specific deployments will likely become the primary focus of most deep learning practitioners.
In such a scenario, the practical focus is to achieve the best trade-offs between predictive, computational, and storage efficiency. This issue leads to the primary research question of this chapter:
\begin{itemize}
    \item[RQ2]\emph{How can we make parameter-efficient adaptations to pre-trained models while improving their computational efficiency on derived tasks?}
\end{itemize}

\section{Related works}

\subsection{Feature-extraction and fine-tuning}
The adaptation of pre-trained models to novel tasks, \ie, transfer learning, can take on many forms.
\textit{Feature-extraction} methods fix the model parameters of the first model layers, which act as feature extractors. Then either end-layers or a linear weighted combination of intermediary representations can be trained~\cite{peters2018deep} on new tasks.
In contrast, \textit{fine-tuning} resumes training of the whole network and only re-initializes the required prediction layer.
In computer vision, fine-tuning generally performs better than feature-extraction, at least for transfers from ImageNet-1k to other tasks~\cite{kornblith2019cvpr}. However, the usability of fine-tuning is reduced when tasks deviate from the source task~\cite{he2019rethinking}.
A similar trend can be observed in natural language processing~\cite{peters2019tune}.
Multiple variants and extensions of fine-tuning have been explored. These include \textit{discriminative fine-tuning}~\cite{howard2018universal}, where later layers, which poses the least general knowledge~\cite{yosinski2014how}, are trained with higher learning rate than early layers; \textit{gradual fine-tuning}~\cite{howard2018universal}, where the last layer is trained initially and earlier layers are gradually added to the set of trainable layers; and \textit{chain-thaw}, which alternates the fine-tuning of individual layers while other layers fixed~\cite{felbo2017using}.

\subsection{Adapters}
An alternative to training the original model parameters are \textit{adapters}~\cite{rebuffi2017learning}. These can be used for transfer learning by adding and training a small set of new model parameters throughout the source network while keeping prior weights fixed. While both adapters and feature-extraction methods leave original model weights fixed, adapters modify the intermediary network representations and feed them back to the network.
Adapters are particularly suitable when multiple networks are derived from a single source model. Since they learn tasks with orders of magnitude fewer parameters than fine-tuning, they are also significantly easier to share if users already possess the original source weights. This has important uses in both federated learning, where the transfer of gradients between devices in a network is the main training bottleneck~\cite{elvebakken2023adaptive}, as well as for adaptation of foundation models and sharing of adapted weights. For instance, a fine-tuned BLOOM model requires 330GB of storage space, which may make the storage of many fine-tuned model variants untenable, whereas multiple adapters can be stored at low cost.

Adapters for CNNs were proposed with point-wise convolutions in series~\cite{rebuffi2017learning} or parallel~\cite{rebuffi2018efficient} with the residual connections.
Bottle-neck projection layers in parallel with dense source layers~\cite{zhu2021counter, hu2022lora} or distributed in series with existing layer~\cite{houlsby2019parameter, pfeiffer2021adapterfusion, mahabadi2021compacter} have been explored in the adaptation of transformers.

While adapters learn only a faction of parameters compared to the source model size, the computational complexity of adapted networks remains as high or higher than the source model. If derived tasks, which are often simpler than source tasks, can be learned by training only a few parameters, should it not then also be possible to use fewer active parameters at inference?

\subsection{Fine-pruning}
\cref{chap:related} outlined common approaches to making model inference more efficient, one of which is pruning (\cref{sec:pruning}).
While most works prune networks with the goal of producing a light-weight network to perform the {same} task as the unpruned network, pruning can also be used to learn {new} tasks. 
In fact, appropriate selection of network weights without further training is enough to adapt to new tasks~\cite{mallya2018piggyback, ramanujan2020whats}, even if it does not achieve optimal predictive performance.
Better target task performance can be achieved with a combination of pruning and fine-tuning, \ie, \textit{fine-pruning}~\cite{sanh2020movement}, which reduces the parameter count of the trained model while modifying network parameters. 
However, the resulting model does not necessarily possess better inference characteristics as such depends on the pruning approach used. 
As described in \cref{sec:pruning}, accelerating inference on GPUs requires a structured pruning method. 
Structured fine-pruning has been explored with both channel-pruning~\cite{yeom2021pruning} and block-pruning~\cite{lagunas2021block}. 
While these methods successfully learn derived networks with both fewer parameters and improved inference efficiency compared with the source network, it is possible to further improve the parameter:accuracy:FLOPs trade-offs through the use of adapters alongside structured pruning.

\section{Structured Pruning Adapters}
\textit{Structured Pruning Adapters}, which were proposed in paper~\ref{pap:spa}, learn a set of $L$ structurally sparse weight matrices $\{\mW_t\}_{t \in 1..L}$ for the target task based on the adaptation $a(\cdot)$ of dense source matrices $\{\mW_s\}_{s \in 1..L}$, each using a small set of target weights, $\Delta\mW_t$, where $\mW_t, \mW_s \in \mathbb{R}^{n \times m}$ and $\left|\Delta\mW_t\right| \ll \left|\mW_s\right|$. Moreover, a structured binary pruning mask $\mM_t \in \{0,1\}^{n \times m}$ is used.
Target weights are derived by fusing source weights with a parallel adapter and applying the binary mask with element-wise multiplication ($\odot$):
\begin{equation}
    \mW_t = (\mW_s + a(\Delta\mW_t)) \odot \mM_t.
    \label{eq:spa-basic-parallel}
\end{equation}
In particular, channel-based pruning, where full rows and columns of the matrix $\mM_t$ are zero-valued, is expressed as:
\begin{equation}
    \mW_t = (\mW_s + a(\Delta\mW_t)) \odot \vm_\text{row} \vm_\text{col}^\top.
    \label{eq:spa-channel-parallel}
\end{equation}

\subsection{Structured Pruning Low-rank Adapter}
\begin{figure}[tb]
    \centering
    \includegraphics[width=0.7\linewidth]{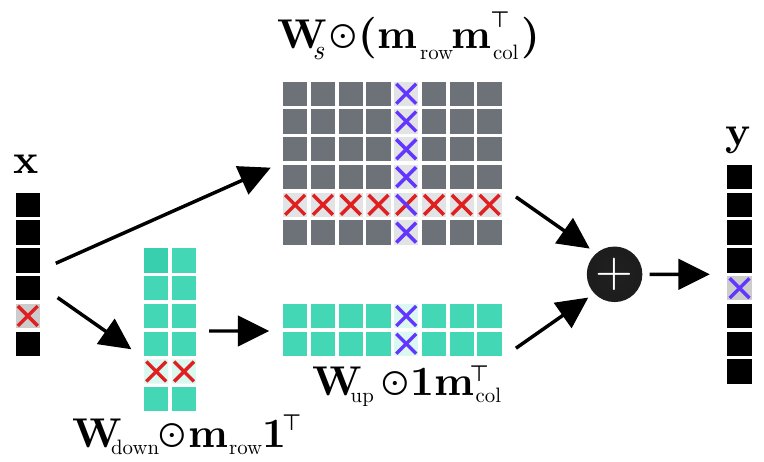}
    \caption{Structured Pruning Low-rank Adapter (SPLoRA). The in/out channel pruning mask affects the adapter as well as source weights.}
    \label{fig:splora}
\end{figure}

A simple yet powerful fusible adapter realization is the Low-rank Adapter~\cite{hu2022lora}
\begin{equation}
    \mW_t = \mW_s + \mW_\text{down}\mW_\text{up},
    \label{eq:spa-lora}
\end{equation}
which uses a bottle-neck projection, $\mW_\text{down}\mW_\text{up}$, where $\mW_\text{down} \in \sR^{n \times r}$, $\mW_\text{up} \in \sR^{r \times m}$, and $r \ll n,m$, to adapt source weights.
Combining \cref{eq:spa-channel-parallel} and \cref{eq:spa-lora} by setting $a(\Delta\mW_t)) = \mW_\text{down}\mW_\text{up}$, we can define the \textit{Structured Pruning Low-rank Adapter (SPLoRA)}: 
\begin{equation}
    \mW_t = \mW_s \odot \vm_\text{row} \vm_\text{col}^\top
        + (\mW_\text{down} \odot \vm_\text{row} \mathbbm{1}^\top ) (\mW_\text{up} \odot \mathbbm{1} \vm_\text{col}^\top ).
    \label{eq:splora}
\end{equation}
Here, row and column pruning masks, $\vm_\text{row}$ and $\vm_\text{col}$, explicitly prune both $\mW_s$ and the adapter weights, $\mW_\text{down}$ and $\mW_\text{up}$.
This is illustrated in \cref{fig:splora}.
For additional details including a derivation of \cref{eq:splora} please see paper~\ref{pap:spa}.

\section{Experiments}
Paper~\ref{pap:spa} proposed SPLoRA as an alternative to fine-tuning under channel-based pruning.
Both learning methods have identical run-time characteristics considering that fused SPLoRA weight contain similar structure and parameter count as fine-pruned weights. 
Accordingly, the experiments are centered around comparisons of parameter count and predictive accuracy. 
Specifically, a ResNet-50 network pre-trained on ImageNet-1k is adapted to three image classification tasks, CIFAR-10~\cite{krizhevsky09learning}, Oxford Flowers 102~\cite{nilsback2008automated}, and Cats and Dogs~\cite{elson2007asirra}, across four pruning criteria based on weight magnitude~\cite{li2017pruning}, gradient~\cite{sun17meprop}, a Taylor expansion approximating changes to the cost function~\cite{molchanov2017pruning}, and layer-wise relevance propagation (LRP)~\cite{yeom2021pruning}. 
Details on experimental protocol and parameter sensitivity are available in paper~\ref{pap:spa}.

\begin{figure}[tb]
    \centering
    \begin{subfigure}{0.45\linewidth}
        \centering
        \includegraphics[width=\linewidth]{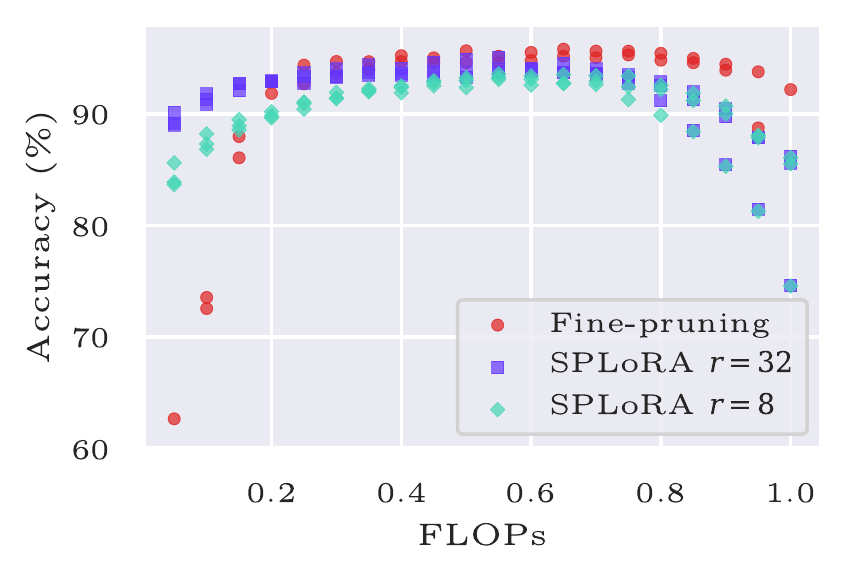}
        \caption{Accuracy versus FLOPs.}
        \label{fig:spa-oxf-results-acc-vs-flops}
    \end{subfigure}
    \hfill
    \begin{subfigure}{0.45\linewidth}
        \centering
        \includegraphics[width=\linewidth]{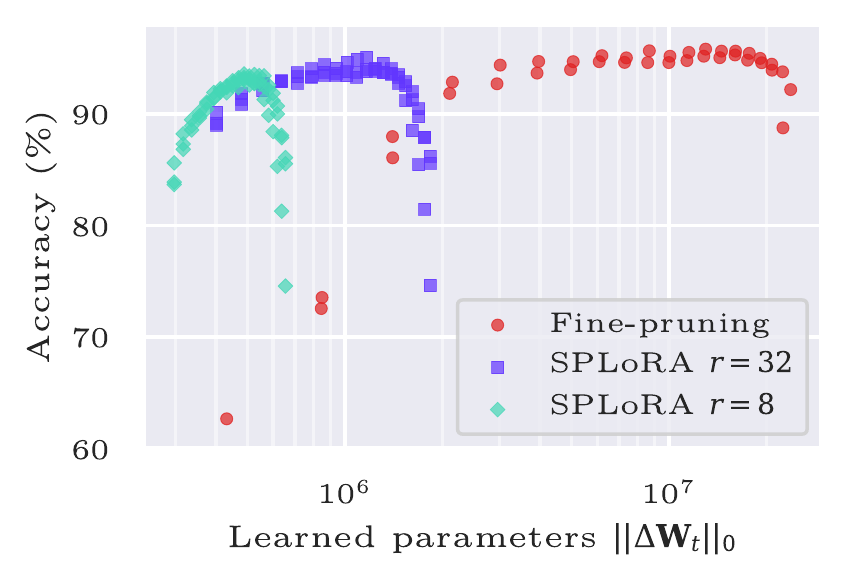}
        \caption{Accuracy versus learned parameters.}
        \label{fig:spa-oxf-results-acc-vs-params}
    \end{subfigure}
    \begin{subfigure}{0.45\linewidth}
        \centering
        \includegraphics[width=\linewidth]{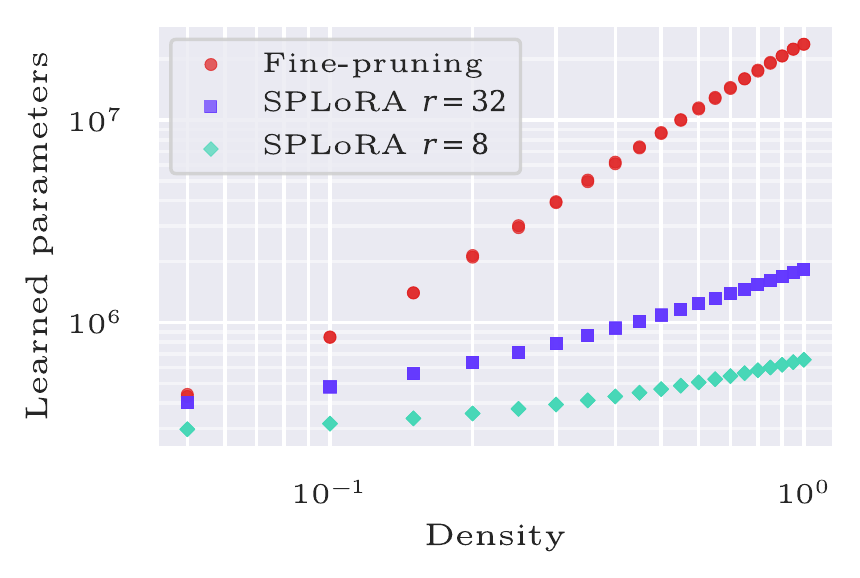}
        \caption{Learned parameter count versus model density fraction.}
        \label{fig:spa-oxf-results-params-vs-ratio}
    \end{subfigure}
    
    \caption{Trade-offs between accuracy (\%), learned parameter count ($\lVert \mW_t \rVert_0$) and floating-point operations (FLOPs) on Oxford Flowers 102 using fine-pruning (\textcolor{ruby}{$\sbullet$}) and SPLoRA with ranks 32 (\textcolor{ioite}{$\ssquare$}) and 8 (\textcolor{amazonite}{$\sdiamond$}) for gradient-based~\cite{sun17meprop} channel pruning.
    }
    \label{fig:spa-oxf-results}
\end{figure}

\cref{tab:spa-resnet50} present experimental results of the average accuracy across datasets shown for each combination of learning and pruning method at 30\% and 10\% weight density, \ie, at respectively 70\% and 90\% pruned weights.
Here, SPLoRA-$r8$, \ie, SPLoRA with rank 8 bottle-neck, requires $19.38\times$ fewer parameters on average at 30\% density with only $1.60\%$ lower accuracy. At 10\% pruning, the model requires $4.65\times$ fewer parameters while increasing accuracy by $5.61\%$ on average.
Likewise, the SPLoRA-$r32$ configuration requires $6.65\times$ fewer parameters with only $0.54\%$ lower accuracy on average at 30\% density. At 10\% it uses $2.05\times$ fewer parameters while improving accuracy by $8.15\%$.
As observed in \cref{fig:spa-oxf-results-params-vs-ratio} as well as \cref{tab:spa-resnet50}, the relative parameter efficiency of SPLoRA adapters becomes lower when more aggressive pruning is used.
However, SPLoRA retains accuracy more effectively than fine-tuning as the network is pruned (\cref{fig:spa-oxf-results-acc-vs-flops}) and generally exhibits significantly more desirable trade-offs between predictive accuracy and learned parameter count (\cref{fig:spa-oxf-results-acc-vs-params}).

\begin{table}[tbh]
\begin{center}
\resizebox{\textwidth}{!}{
\begin{tabular}{llrlll}
\toprule 
{\textbf{Pruning}}
    &\multirow{1}{*}{\textbf{Learning}}
    &\multirow{1}{*}{\textbf{Density}}
    &\multicolumn{1}{c}{\textbf{$\Delta$Params}}
    &\multicolumn{1}{c}{\textbf{FLOPs}}
    &\multicolumn{1}{c}{\textbf{Avg. accuracy}}
    \\
{\textbf{method}}
    & {\textbf{method}}
    &
    & \multicolumn{1}{c}{(K)} 
    & \multicolumn{1}{c}{(M)} 
    & \multicolumn{1}{c}{(\%)}
    \\

\midrule 

& \multirow{1}{*}{Fine-tuning}  & 100\% & $23{,}520.8	$\tiny$\pm 0.0$	  & $1{,}304.7	$\tiny$\pm 0.0$	& $\mathbf{96.20}	$\tiny$\pm 0.05$\vspace{3pt}\\
\multirow{1}{*}{Unpruned}           & \multirow{1}{*}{LoRA-$r32$}   & 100\% & $1{,}644.5 	$\tiny$\pm 0.0$ \textcolor{lgreen}{ \ ($\downarrow14.3\times$)}	& $1{,}304.7	$\tiny$\pm 0.0$	& $90.83	$\tiny$\pm 2.16$ \textcolor{lred}{ \ ($-5.37$)} \vspace{3pt}\\
                                    & \multirow{1}{*}{LoRA-$r8$}    & 100\% & \phantom{$0{,}$}$\mathbf{466.3}	$\tiny$\pm 0.0$ \textcolor{lgreen}{ \ ($\downarrow50.4\times$)}	& $1{,}304.7	$\tiny$\pm 0.0$	& $91.72	$\tiny$\pm 2.15$ \textcolor{lred}{ \ ($-4.48$)}\\
\midrule 
                                    & \multirow{2}{*}{Fine-pruning} & 30\% & $4{,}427.1                                         $\tiny$\pm 72.5$  & \phantom{$0{,}$}$785.4   $\tiny$\pm 5.4$  & $\mathbf{96.22}                               $\tiny$\pm 0.77$  \\
                                    &                               & 10\% & \phantom{$0{,}$}$599.1                                         $\tiny$\pm 20.1$  & \phantom{$0{,}$}$352.1   $\tiny$\pm 3.9$  & $85.16                               $\tiny$\pm 1.08$  \vspace{3pt}\\
\multirow{2}{*}{Weight}           
                                    & \multirow{2}{*}{SPLoRA-$r32$} & 30\% & \phantom{$0{,}$}$618.3   $\tiny$\pm 2.5$ \textcolor{lgreen}{ \ ($\downarrow 7.2\times$)}  & \phantom{$0{,}$}$778.6   $\tiny$\pm 2.9$  & $96.20  $\tiny$\pm 0.31$ \textcolor{gray}{ \ ($-0.02$)}\\
                                    &                               & 10\% & \phantom{$0{,}$}$294.8   $\tiny$\pm 0.4$ \textcolor{lgreen}{ \ ($\downarrow 2.0\times$)}  & \phantom{$0{,}$}${335.4}   $\tiny$\pm 7.3$  & $\mathbf{92.88}   $\tiny$\pm 0.82$ \textcolor{lgreen}{ \ ($+7.72$)}\vspace{3pt}\\
                                    & \multirow{2}{*}{SPLoRA-$r8$}  & 30\% & \phantom{$0{,}$}$\mathbf{210.0}  $\tiny$\pm 1.9$ \textcolor{lgreen}{ \ ($\downarrow 21.1\times$)}  & \phantom{$0{,}$}$773.4   $\tiny$\pm 2.1$  & $95.15  $\tiny$\pm 0.17$ \textcolor{lred}{ \ ($-1.07$)} \\
                                    &                               & 10\% & \phantom{$0{,}$}$\mathbf{128.9}   $\tiny$\pm 0.1$ \textcolor{lgreen}{ \ ($\downarrow 4.6\times$)}  & \phantom{$0{,}$}$338.9   $\tiny$\pm 7.0$  & $91.27   $\tiny$\pm 0.65$ \textcolor{lgreen}{ \ ($+6.11$)}\\
\midrule
                                    & \multirow{2}{*}{Fine-pruning} & 30\% & $3{,}719.8                                         $\tiny$\pm 59.2$  & \phantom{$0{,}$}$571.9   $\tiny$\pm 6.5$  & $\mathbf{96.13}                               $\tiny$\pm 0.28$ \\
                                    &                               & 10\% & \phantom{$0{,}$}$615.7                                          $\tiny$\pm 4.3$  & \phantom{$0{,}$}$244.7   $\tiny$\pm 3.3$  & $83.58                               $\tiny$\pm 0.69$ \vspace{3pt}\\
\multirow{2}{*}{Gradient}          
                                    & \multirow{2}{*}{SPLoRA-$r32$} & 30\% & \phantom{$0{,}$}$601.0   $\tiny$\pm 0.4$ \textcolor{lgreen}{ \ ($\downarrow 6.2\times$)}  & \phantom{$0{,}$}$564.6   $\tiny$\pm 1.5$  & $95.55  $\tiny$\pm 0.20$ \textcolor{gray}{ \ ($-0.58$)}\\
                                    &                               & 10\% & \phantom{$0{,}$}$293.1   $\tiny$\pm 0.4$ \textcolor{lgreen}{ \ ($\downarrow 2.1\times$)}  & \phantom{$0{,}$}${244.0}   $\tiny$\pm 1.9$  & $\mathbf{94.18}   $\tiny$\pm 0.33$ \textcolor{lgreen}{ \ ($+10.60$)}\vspace{3pt}\\
                                    & \multirow{2}{*}{SPLoRA-$r8$}  & 30\% & \phantom{$0{,}$}$\mathbf{205.2}  $\tiny$\pm 0.2$ \textcolor{lgreen}{ \ ($\downarrow 18.1\times$)}  & \phantom{$0{,}$}$565.2   $\tiny$\pm 4.1$  & $94.61  $\tiny$\pm 0.22$ \textcolor{lred}{ \ ($-1.52$)}\\
                                    &                               & 10\% & \phantom{$0{,}$}$\mathbf{128.3}   $\tiny$\pm 0.0$ \textcolor{lgreen}{ \ ($\downarrow 4.8\times$)}  & \phantom{$0{,}$}$245.4   $\tiny$\pm 4.0$  & $91.97  $\tiny$\pm 0.40$ \textcolor{lgreen}{ \ ($+8.39$)} \\
\midrule
                                    & \multirow{2}{*}{Fine-pruning} & 30\% & $3{,}392.8                                         $\tiny$\pm 81.1$  & \phantom{$0{,}$}$559.9   $\tiny$\pm 0.7$  & $\mathbf{95.61}                               $\tiny$\pm 0.25$ \\
                                    &                               & 10\% & \phantom{$0{,}$}$576.8                                          $\tiny$\pm 9.9$  & \phantom{$0{,}$}$236.9   $\tiny$\pm 3.3$  & $83.01                               $\tiny$\pm 1.66$ \vspace{3pt}\\
\multirow{2}{*}{Taylor}            
                                    & \multirow{2}{*}{SPLoRA-$r32$} & 30\% & \phantom{$0{,}$}$599.7   $\tiny$\pm 0.9$ \textcolor{lgreen}{ \ ($\downarrow 5.7\times$)}  & \phantom{$0{,}$}$555.5   $\tiny$\pm 6.7$  & $95.37  $\tiny$\pm 0.25$ \textcolor{gray}{ \ ($-0.24$)}\\
                                    &                               & 10\% & \phantom{$0{,}$}$292.6   $\tiny$\pm 0.5$ \textcolor{lgreen}{ \ ($\downarrow 2.0\times$)}  & \phantom{$0{,}$}${242.0}   $\tiny$\pm 1.5$  & $\mathbf{93.93}   $\tiny$\pm 0.04$ \textcolor{lgreen}{ \ ($+10.92$)}\vspace{3pt}\\
                                    & \multirow{2}{*}{SPLoRA-$r8$}  & 30\% & \phantom{$0{,}$}$\mathbf{205.3}  $\tiny$\pm 0.1$ \textcolor{lgreen}{ \ ($\downarrow 16.5\times$)}  & \phantom{$0{,}$}$566.2  $\tiny$\pm 10.9$  & $94.46  $\tiny$\pm 0.29$ \textcolor{lred}{ \ ($-1.15$)}\\
                                    &                               & 10\% & \phantom{$0{,}$}$\mathbf{128.4}   $\tiny$\pm 0.1$ \textcolor{lgreen}{ \ ($\downarrow 4.5\times$)}  & \phantom{$0{,}$}$243.2   $\tiny$\pm 9.8$  & $91.60   $\tiny$\pm 0.35$ \textcolor{lgreen}{ \ ($+8.59$)}\\
\midrule
                                    & \multirow{2}{*}{Fine-pruning} & 30\% & $4{,}428.1                                         $\tiny$\pm 20.6$  & \phantom{$0{,}$}$719.9   $\tiny$\pm 0.7$  & $\mathbf{96.85}                               $\tiny$\pm 0.10$ \\
                                    &                               & 10\% & \phantom{$0{,}$}$608.4                                          $\tiny$\pm 6.3$  & \phantom{$0{,}$}$301.6   $\tiny$\pm 2.3$  & $90.54                               $\tiny$\pm 1.40$ \vspace{3pt}\\
\multirow{2}{*}{LRP}           
                                    & \multirow{2}{*}{SPLoRA-$r32$} & 30\% & \phantom{$0{,}$}$592.6   $\tiny$\pm 0.9$ \textcolor{lgreen}{ \ ($\downarrow 7.5\times$)}  & \phantom{$0{,}$}$585.5   $\tiny$\pm 6.7$  & $95.52  $\tiny$\pm 0.21$ \textcolor{lred}{ \ ($-1.33$)}\\
                                    &                               & 10\% & \phantom{$0{,}$}$290.9   $\tiny$\pm 0.2$ \textcolor{lgreen}{ \ ($\downarrow 2.1\times$)}  & \phantom{$0{,}$}${270.4}   $\tiny$\pm 6.4$  & $\mathbf{93.90}   $\tiny$\pm 0.36$ \textcolor{lgreen}{ \ ($+3.36$)} \vspace{3pt}\\
                                    & \multirow{2}{*}{SPLoRA-$r8$}  & 30\% & \phantom{$0{,}$}$\mathbf{203.3}  $\tiny$\pm 0.5$ \textcolor{lgreen}{ \ ($\downarrow 21.8\times$)}  & \phantom{$0{,}$}$591.1  $\tiny$\pm 12.4$  & $94.20  $\tiny$\pm 0.19$ \textcolor{lred}{ \ ($-2.65$)} \\
                                    &                               & 10\% & \phantom{$0{,}$}$\mathbf{128.0}   $\tiny$\pm 0.1$ \textcolor{lgreen}{ \ ($\downarrow 4.7\times$)}  & \phantom{$0{,}$}$281.6   $\tiny$\pm 1.7$  & $91.17  $\tiny$\pm 0.77$ \textcolor{gray}{ \ ($-0.64$)}  \\
\bottomrule 
\end{tabular}
}
\end{center}
\caption{Channel-based transfer-pruning from ResNet-50 pre-trained on ImageNet to CIFAR-10, Oxford Flowers 102, and Cats \& Dogs using pruning methods based on Weight~\cite{li2017pruning}, Gradient~\cite{sun17meprop}, Taylor~\cite{molchanov2017pruning}, and LRP~\cite{yeom2021pruning}. 
Please note that SPLoRA and LoRA are identical at 100\% density. 
$\Delta$Params and floating-point operations (FLOPs) are shown for CIFAR-10. Mean $\pm$ standard deviation over three runs is shown for each metric. The average accuracy over datasets is noted as the average of means and standard deviations within a dataset.
Changes relative to closest fine-pruning density with same pruning method are specified in parentheses. The best metric for each combination of pruning-method and density is highlighted with bold.
}
\label{tab:spa-resnet50}
\end{table}

\section{Conclusion}
This chapter presented the work on Structured Pruning Adapters (paper~\ref{pap:spa}). 
In particular, the proposed channel-based method, SPLoRA, was benchmarked against fine-tuning with channel-pruning with a battery of pruning criteria and on multiple transfer learning tasks in the image domain. 
SPLoRA systematically exhibited superior parameter efficiency, requiring $19\times$ fewer parameters on average at $30\%$ remaining weight with only $1.60\%$ lower accuracy. 
In the low-density regimen at 10\% remaining weights, SPLoRA retained accuracy better than fine-pruning, achieving $8.15\%$ point higher accuracy on average while requiring $2.05\times$ fewer learned parameters.



\chapter{Conclusions and future work}\label{chap:conclusion}
Deep neural networks (DNNs) play an ever more important role in many technological solutions we use everyday. Unlocking phones with facial scans, analyzing photo libraries to suggest compilations, and ensuring our safety while driving by monitoring the road gradually become taken for granted.
However, the size of DNNs and energy expenditure associated with their deployment increases at an accelerating pace.
While much work in the field has focused primarily on applying DNNs on novel problems or improving predictive performance on prior ones, this dissertation represents an effort to make DNNs more computationally and memory efficient. Hopefully, these improvements will translate to more power and hardware efficient deployment of DNNs.

Specifically, \cref{chap:cin} describes efforts to accelerate the inference of spatio-temporal DNNs on time-series processing tasks during online deployment. 
These include the development of a new framework of neural network architectures, that operate efficiently on continual streams of data. The theory and experimental validation of \textit{Continual Inference Networks} (CINs), as they were dubbed, was developed over publications \ref{pap:co3d} to \ref{pap:colib}, each of which tackled the translation of state-of-the-art architectures for different offline inference tasks to the context of online stream processing.

\textit{Continual 3D Convolutional Neural Networks} (\cref{sec:co3d}, paper~\ref{pap:co3d}) translate 3D CNNs, which in prior formulations were hamstrung to operate exclusively of full spatio-temporal clips, to a sequential formulation, where each time-step is processed incrementally. Continual 3D CNNs reduce the floating-point operations (FLOPs) by up to $15\times$ and increase throughput up to $9\times$ on CPU and $7\times$ on GPU during online inference compared to ordinary 3D CNNs applied with a sliding-window approach. Importantly, the reformulation is weight-compatible with prior architectures.

\textit{Continual Spatio-temporal Graph Convolutional Networks} (\cref{sec:costgcn}, paper~\ref{pap:costgcn}) are the corresponding translation of ST-GCN-based works for classification of spatio-temporal graphs. Since the CIN formulation generally offers improvements in computational complexity in proportion to the corresponding temporal receptive field during sliding-window processing with non-CINs, Continual ST-GCN formulations achieve up to $108\times$ FLOPs reductions as well as a $26\times$ and $24\times$ throughput increase on CPU and GPU for networks with the commonly used clip size of 300 time-steps.

\textit{Continual Transformer Encoders} (\cref{sec:cotrans}, paper~\ref{pap:cotrans}), the CIN reformulation of the Transformer Encoder, likewise offer significant benefits during online processing. 
By utilizing novel computational schemes for the scaled dot-product attention, a one-block Continual Transformer Encoder can reduce computational complexity by $63\times$ for a sequence length of $64$. A two-block version reduces complexity by $2.5\times$.

To support the implementation of CINs, the python library \textit{Continual Inference} was implemented (\cref{sec:colib}, paper~\ref{pap:colib}). This augments the standard components of PyTorch with the ability to perform the forward operation efficiently in a progressive manner, time-step by time-step.

Though CINs have left their infancy, there are still many avenues of future work. 
One such avenue is the exploration of CINs for uses other than classification, \eg, for efficient object detection and tracking or video-based anomaly detection.
Moreover, CINs in their current form assume that data in the input stream is uniformly sampled in time. However, data streams with unevenly spaced samples can easily occur on embedded devices, where the limited computational power may occasionally be diverted to other tasks than data acquisition and processing. The development of CINs that operate gracefully in this context would be of great value to such deployments.
The current publications and above-described future works on CINs assume hand-crafted translation from ordinary networks to efficient CINs. 
However, it may be possible to automatically identify similar architectural modifications through automated analysis of the computational graph for consecutive time-steps.
Finally, a great deal of work on community outreach is needed to make an impact with CINs, including the creation of tutorials, blogposts, and collaborations with other tools in the machine learning ecosystem.

\cref{chap:spa} and paper~\ref{pap:spa} presented an orthogonal work on \textit{Structured Pruning Adapters} (SPAs). 
Motivated by scenarios where a large source model is adapted to many down-stream tasks, SPAs were proposed as an alternative to fine-tuning with pruning. SPAs utilize adapters to learn downstream tasks with an order of magnitude fewer parameters, provided the source model weights are also available. Through the combination of channel-based pruning and fusible parallel adapters, the proposed channel-SPAs required $19\times$ fewer learned parameters at a modest $1.60\%$ lower accuracy at moderate pruning of $30\%$ remaining weights. Under aggressive pruning at $10\%$ remaining weights, the channel-SPAs retained an $8.15\%$-point higher accuracy than fine-tuning while requiring $2.05\times$ fewer learned parameters.

The work described in \cref{chap:spa} has just scratched the surface of viable methods in the intersection between structured pruning and low-parameter adapter networks. Specifically, SPAs for block-pruning and N:M pruning are interesting directions of future research. Moreover, an explanation of why the channel-SPAs systematically retained their predictive performance better than fine-tuning in the highly pruned regimen is yet to be found. An experimentally validated explanation of this phenomenon could be important in furthering the pruning research field and might the enable discovery of even more efficient pruning methods.

\backmatter

\cleardoublepage
\bibliographystyle{plainnat} 
\bibliography{bibliography}

\mainmatter 
\part{Publications}
\pagenumbering{arabic}
\setsecnumdepth{subsubsection}

\appendix
\renewcommand{\appendixname}{Publication}
\renewcommand{\thechapter}{\arabic{chapter}}

\label{part:publications}








\chapter{Continual 3D Convolutional Neural Networks for Real-time Processing of Videos}
\label{pap:co3d}
\mycbox{amazonite}{white}{}

\noindent\textbf{Authors}: \underline{Lukas Hedegaard} and Alexandros Iosifidis.\\

\noindent\textbf{Publication}: Computer Vision – ECCV 2022. ECCV 2022. Lecture Notes in Computer Science, vol 13664. Springer, Cham. https://doi.org/10.1007/978-3-031-19772-7\_22. \\

\noindent\textbf{Link}: \url{https://link.springer.com/chapter/10.1007/978-3-031-19772-7_22}

\chapter{Continual Spatio-Temporal Graph Convolutional Networks}
\label{pap:costgcn}
\mycbox{ioite}{white}{}

\noindent\textbf{Authors}: \underline{Lukas Hedegaard}, Negar Heidari, and Alexandros Iosifidis.\\

\noindent\textbf{Publication}: Pattern Recognition, Volume 140, 2023, 109528, ISSN 0031-3203.\\

\noindent\textbf{Link}: \url{https://www.sciencedirect.com/science/article/pii/S0031320323002285}

\chapter{Continual Transformers: Redundancy-Free Attention for Online Inference}
\label{pap:cotrans}
\mycbox{amazonite}{white}{}

\noindent\textbf{Authors}: \underline{Lukas Hedegaard}, Arian Bakhtiarnia, and Alexandros Iosifidis.\\

\noindent\textbf{Publication}: International Conference on Learning Representations, 2023.\\

\noindent\textbf{Link}: \url{https://openreview.net/pdf?id=PolHquob8M7}

\chapter{Continual Inference: A Library for Efficient Online Inference with Deep Neural Networks in PyTorch}
\label{pap:colib}
\mycbox{ioite}{white}{}

\noindent\textbf{Authors}: \underline{Lukas Hedegaard}, and Alexandros Iosifidis.\\

\noindent\textbf{Publication}: Computer Vision – ECCV 2022 Workshops. Lecture Notes in Computer Science, vol 13803. Springer. Isbn: 978-3-031-25066-8. \\

\noindent\textbf{Link}: \url{https://link.springer.com/chapter/10.1007/978-3-031-25082-8_2}

\chapter{Human Activity Recognition}
\label{pap:chap14}
\mycbox{amazonite}{white}{}

\noindent\textbf{Authors}: \underline{Lukas Hedegaard}, Negar Heidari, and Alexandros Iosifidis.\\

\noindent\textbf{Publication}: ``Deep Learning for Robot Perception and Cognition'', A. Iosifidis and A. Tefas, Eds., 1st ed., Academic Press, Jan. 2022, ch. 14, isbn: 9780323857871. \\

\noindent\textbf{Link}: \url{https://www.sciencedirect.com/science/article/pii/B9780323857871000191}

\chapter{Graph Neural Networks}
\label{pap:chap4}
\mycbox{ioite}{white}{}

\noindent\textbf{Authors}: Negar Heidari, \underline{Lukas Hedegaard}, and Alexandros Iosifidis.\\

\noindent\textbf{Publication}: ``Deep Learning for Robot Perception and Cognition'', A. Iosifidis and A. Tefas, Eds., 1st ed., Academic Press, Jan. 2022, ch. 4, isbn: 9780323857871.\\

\noindent\textbf{Contribution}: Primary author of section 4.4 ``Graph attention network (GAT)'' and reviewer of the remaining chapter contents. \\

\noindent\textbf{Link}: \url{https://www.sciencedirect.com/science/article/pii/B9780323857871000099}

\chapter{Structured Pruning Adapters}
\label{pap:spa}
\mycbox{amazonite}{white}{}

\noindent\textbf{Authors}: \underline{Lukas Hedegaard}, Omar Ali Sheikh-Omar and Alexandros Iosifidis.\\

\noindent\textbf{Publication}: ArXiv preprint arXiv:2211.10155, 2022. \\

\noindent\textbf{Link}: \url{https://arxiv.org/abs/2211.10155}

\end{document}